\documentclass{article}

\usepackage{microtype}
\usepackage{graphicx}
\usepackage{subcaption}
\usepackage{booktabs} % for professional tables

\usepackage{hyperref}

\usepackage[preprint]{icml2026}

\usepackage{amsmath}
\usepackage{amssymb}
\usepackage{mathtools}
\usepackage{amsthm}

\usepackage{multirow}
\usepackage{multicol}

\usepackage{lipsum}

\usepackage[capitalize,noabbrev]{cleveref}
\usepackage{enumitem}

\theoremstyle{plain}

\theoremstyle{definition}

\theoremstyle{remark}

\usepackage[textsize=tiny]{todonotes}

\usepackage[table]{xcolor}
\definecolor{cyan3}{RGB}{130,170,168}
\definecolor{cyan1}{RGB}{142,171,168}
\definecolor{blue1}{RGB}{160,180,195}
\definecolor{blue2}{RGB}{145,165,180}
\definecolor{blue3}{RGB}{135,155,170}
\definecolor{green3}{RGB}{130,150,130}
\definecolor{red2}{RGB}{205,150,140}
\renewcommand{\upuparrows}{%
  \mathrel{\scalebox{0.8}[1]{$\uparrow$}\mkern0mu\scalebox{0.8}[1]{$\uparrow$}}%
}
\renewcommand{\downdownarrows}{%
  \mathrel{\scalebox{0.8}[1]{$\downarrow$}\mkern0mu\scalebox{0.8}[1]{$\downarrow$}}%
}
\newcommand{\upupuparrows}{%
  \mathrel{\scalebox{0.7}[1]{$\uparrow$}\mkern0mu\scalebox{0.7}[1]{$\uparrow$}\mkern0mu\scalebox{0.7}[1]{$\uparrow$}}%
}

\usepackage{tcolorbox}
\tcbuselibrary{skins, raster, breakable}
\tcbset{
  mybox1/.style={
    colback=cyan1!10!white, %gray!10!white,            % 内容背景为浅青色和白色混合
    colframe=cyan3!90!green, %gray!60!white,           % 边框为深青色和黑色混合
    coltitle=white,                   % 标题文字颜色为白色
    fonttitle=\bfseries,              % 标题字体加粗
    colbacktitle=cyan3!90!green, %gray!60!white,        % 标题背景为蓝色和青色的渐变
    title filled,                     % 标题区域填充颜色
    coltitle=white,                   % 标题文字颜色为白色
    boxrule=2pt,                      % 边框宽度
    rounded corners=all,              % 圆角
    toptitle=3pt,
    bottomtitle=3pt,
    left=10pt,            % 左边内边距
    right=10pt,           % 右边内边距
    top=5pt,             % 上方内边距
    bottom=5pt,           % 下方内边距
  },
  mybox2/.style={
    colback=blue1!10!white, %gray!10!white,            % 内容背景为浅青色和白色混合
    colframe=blue3!90!blue, %gray!60!white,           % 边框为深青色和黑色混合
    coltitle=white,                   % 标题文字颜色为白色
    fonttitle=\bfseries,              % 标题字体加粗
    colbacktitle=blue3!90!blue, %gray!60!white,        % 标题背景为蓝色和青色的渐变
    title filled,                     % 标题区域填充颜色
    coltitle=white,                   % 标题文字颜色为白色
    boxrule=2pt,                      % 边框宽度
    rounded corners=all,              % 圆角
    toptitle=3pt,
    bottomtitle=3pt,
    left=10pt,            % 左边内边距
    right=10pt,           % 右边内边距
    top=5pt,             % 上方内边距
    bottom=5pt,           % 下方内边距
  }
}
\newtcolorbox[auto counter, number within=section]{template1}[2][]{
  title={#2}, % 自动生成编号和标题
  mybox1,
  #1
}
\newtcolorbox[auto counter, number within=section]{template2}[2][]{
  title={#2}, % 自动生成编号和标题
  mybox2,
  #1
}

\newtcolorbox[auto counter, number within=section]{btemplate1}[2][]{
  title={#2}, % 自动生成编号和标题
  mybox1,
  breakable,
  #1
}

\newtcolorbox[auto counter, number within=section]{promptbox}[2][]{
  title={#2}, % 自动生成编号和标题
  mybox2,
  breakable,
  #1
}

\usepackage{tabularx}

\definecolor{morandi-azure}{RGB}{147,169,185}
\definecolor{morandi-slate}{RGB}{139,151,158}
\newtcolorbox[auto counter, number within=section]{rubricbox}[1]{
    enhanced,
    breakable,
    colback=morandi-azure!10,       % 需要 xcolor 包支持 (或用 blue!3)
    colframe=morandi-slate,% 边框：深蓝
    coltitle=white,        % 标题文字：白
    fonttitle=\bfseries,   % 标题加粗
    boxrule=2pt,
    title={Rubric \thetcbcounter:~#1}, % 自动标题 "Rubric D.1: ..."
    label={rubric:\thetcbcounter},      % 自动 Label
    colbacktitle=morandi-slate, %gray!60!white,        % 标题背景为蓝色和青色的渐变
    title filled,                     % 标题区域填充颜色
    coltitle=white,                   % 标题文字颜色为白色
    boxrule=2pt,                      % 边框宽度
    rounded corners=all,              % 圆角
    toptitle=3pt,
    bottomtitle=3pt,
    left=10pt,            % 左边内边距
    right=10pt,           % 右边内边距
    top=5pt,             % 上方内边距
    bottom=5pt,           % 下方内边距
}

\icmltitlerunning{Can Large Language Models Simulate Human Cognition Beyond Behavioral Imitation?}

\begin{document}

\twocolumn[
  \icmltitle{Can Large Language Models Simulate Human Cognition\\ Beyond Behavioral Imitation?}

  % It is OKAY to include author information, even for blind submissions: the
  % style file will automatically remove it for you unless you've provided
  % the [accepted] option to the icml2026 package.

  % List of affiliations: The first argument should be a (short) identifier you
  % will use later to specify author affiliations Academic affiliations
  % should list Department, University, City, Region, Country Industry
  % affiliations should list Company, City, Region, Country

  % You can specify symbols, otherwise they are numbered in order. Ideally, you
  % should not use this facility. Affiliations will be numbered in order of
  % appearance and this is the preferred way.
  \icmlsetsymbol{equal}{*}

  \begin{icmlauthorlist}
    \icmlauthor{Yuxuan Gu}{yyy}
    \icmlauthor{Lunjun Liu}{yyy}
    \icmlauthor{Xiaocheng Feng}{yyy}
    \icmlauthor{Kun Zhu}{yyy}
    \icmlauthor{Weihong Zhong}{yyy}
    \icmlauthor{Lei Huang}{yyy}
    \icmlauthor{Bing Qin}{yyy}
    %\icmlauthor{}{sch}
    %\icmlauthor{Firstname8 Lastname8}{sch}
    %\icmlauthor{Firstname8 Lastname8}{yyy,comp}
    %\icmlauthor{}{sch}
    %\icmlauthor{}{sch}
  \end{icmlauthorlist}

  \icmlaffiliation{yyy}{Harbin Institute of Technology}
  %\icmlaffiliation{comp}{Company Name, Location, Country}
  %\icmlaffiliation{sch}{School of ZZZ, Institute of WWW, Location, Country}

  \icmlcorrespondingauthor{Xiaocheng Feng}{xcfeng@ir.hit.edu.cn}
  %\icmlcorrespondingauthor{Firstname2 Lastname2}{first2.last2@www.uk}

  % You may provide any keywords that you find helpful for describing your
  % paper; these are used to populate the "keywords" metadata in the PDF but
  % will not be shown in the document
  \icmlkeywords{Machine Learning, ICML}

  \vskip 0.3in
]
\SetLabelAlign{parcenter}{\parbox[t]{\labelwidth}{\centering#1}}
% this must go after the closing bracket ] following \twocolumn[ ...

% This command actually creates the footnote in the first column listing the
% affiliations and the copyright notice. The command takes one argument, which
% is text to display at the start of the footnote. The \icmlEqualContribution
% command is standard text for equal contribution. Remove it (just {}) if you
% do not need this facility.

% Use ONE of the following lines. DO NOT remove the command.
% If you have no special notice, KEEP empty braces:
\printAffiliationsAndNotice{}  % no special notice (required even if empty)
% Or, if applicable, use the standard equal contribution text:
% \printAffiliationsAndNotice{\icmlEqualContribution}

\begin{abstract}
An essential problem in artificial intelligence is whether LLMs can simulate human cognition or merely imitate surface-level behaviors, while existing datasets suffer from either synthetic reasoning traces or population-level aggregation, failing to capture authentic individual cognitive patterns.
We introduce a benchmark grounded in the longitudinal research trajectories of $217$ researchers across diverse domains of artificial intelligence, where each author's scientific publications serve as an externalized representation of their cognitive processes.
To distinguish whether LLMs transfer cognitive patterns or merely imitate behaviors, our benchmark deliberately employs a cross-domain, temporal-shift generalization setting.
A multi-dimensional cognitive alignment metric is further proposed to assess individual-level cognitive consistency. 
Through systematic evaluation of state-of-the-art LLMs and various enhancement techniques, we provide a first-stage empirical study on the questions: (1) How well do current LLMs simulate human cognition? and (2) How far can existing techniques enhance these capabilities?
%Our findings reveal critical gaps between behavioral mimicry and genuine cognitive alignment, offering insights for developing more human-like artificial intelligence systems.
\end{abstract}

\section{Introduction}

\begin{figure*}[t]
\centering
\includegraphics[width=0.95\textwidth]{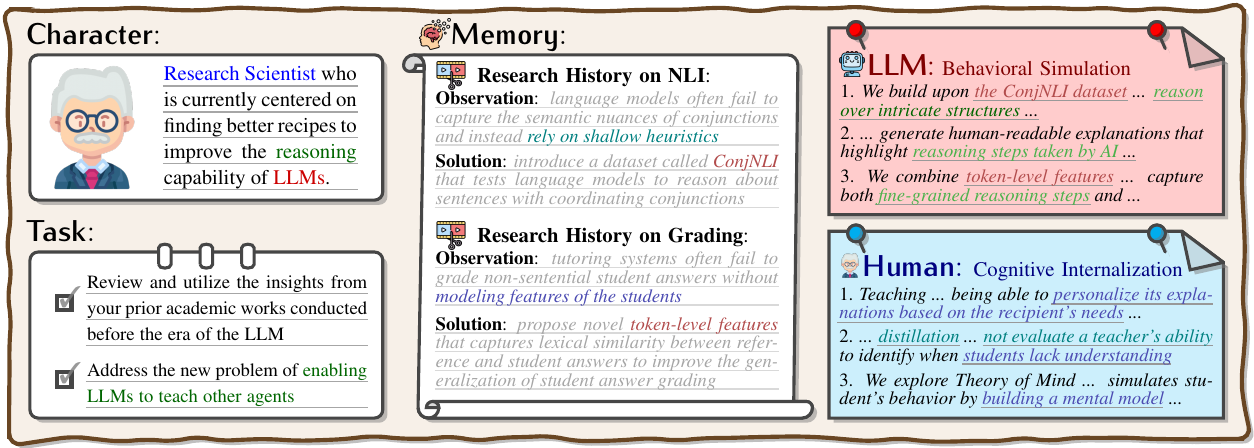}
%\vspace{-1pt}
\caption{Motivating case of human simulation.
The memory panel presents two representative cases of the author's prior research with reasoning traits: (1) an explicit awareness of \textcolor{teal}{shallow heuristic modeling} in LLMs and a deliberate effort to avoid such shortcuts, and (2) a problem-solving style that starts from \textcolor{blue!40!gray}{modeling the research objectives}. \textcolor{red!50!black}{Upper-right Panel}: LLMs can correctly interpret the \textcolor{green!40!black}{task objective} but seem to rely on \textcolor{red!40!gray}{directly imitating specific methods} from the research memory. \textcolor{blue!50!black}{Lower-right Panel}: In contrast, the authentic author shares \textcolor{blue!40!gray}{a consistent abstraction and reasoning trajectory} across problem scenarios. This comparison illustrates a central question motivating our study: whether LLMs can transcend behavioral imitation to internalize and reproduce human-like cognitive processes.
}
\label{fig:motivation}
\vspace{-3pt}
\end{figure*}

Large language models (LLMs) have demonstrated remarkable advances \cite{achiam2023gpt, dubey2024llama, yang2025qwen3, team2025kimi, liu2025deepseek} in a wide range of applications, 
%multi-turn dialogue generation \cite{yi2024survey}, 
such as code completion \cite{jiang2024survey}, complex task solving \cite{plaat2024reasoning}, and science discovery \cite{zhou-etal-2025-hypothesis}.
%including complex reasoning, tool-augmented problem solving, and multi-step planning. 
These emerging capabilities have reactivated interest in a fundamental question of artificial intelligence: \textit{to what extent can machines simulate human intelligence}? \cite{turing2004intelligent, mccarthy2006proposal} %turing2007computing
Recent progress \cite{chen2024from} in role-playing, persona alignment, and individualized generation suggests that LLMs can simulate human behaviors \cite{shao-etal-2023-character,wang-etal-2024-rolellm,ye-etal-2025-cpo,wang2025coser} and personality traits \cite{jiang2023evaluating,serapio2023personality,jiang-etal-2024-personallm,dai2025profile}.
%gao-etal-2025-tailorrpa %lee-etal-2025-llms
%\citep{jiang2023personallm, shuster2022blenderbot, zhao2023persona}. 
%However, it remains unclear whether these behaviors reflect genuine cognitive alignment or merely surface-level imitation.
On the other hand, concerns have also been raised that such behaviors may reflect merely surface-level imitation rather than genuine cognitive alignment \cite{qin-etal-2025-r}, as discussed in the motivating illustration of \Cref{fig:motivation}.
%provides a motivating illustration for this discussion.
%such as perception \cite{minsky2007steps}, problem solving \cite{simon1971human}, and reasoning \cite{newell1972human}.
%the apparent human-like behavior of LLMs may reflect behavioral imitation rather than genuine cognitive simulation.
%However, such efforts focus primarily on behavioral similarity \cite{qin-etal-2025-r}, rather than on capturing cognitive processes \cite{anderson1995cognitive}, such as perception \cite{minsky2007steps}, problem solving \cite{simon1971human}, and 
%abstraction \cite{mccarthy2006proposal}, induction \cite{minsky2007steps}
%reasoning \cite{newell1972human}, which characterize the human mind.

%Importantly, such examples are illustrative rather than conclusive, and it remains unclear whether they reflect systematic differences or isolated phenomena.

%
% Despite their practical success, existing approaches exhibit fundamental limitations. Most models can mimic surface-level stylistic signals—such as tone, preferences, or Big-Five personality profiles—but they do not reproduce how humans structure problems, identify knowledge gaps, or formulate solution strategies. For instance, while an LLM may imitate the writing style of a renowned scientist, it rarely replicates their characteristic reasoning patterns, such as how they decompose a research problem or justify the introduction of a new method. This distinction between behavioral features and cognitive modes of thought is crucial: achieving the former does not imply progress toward the latter, and the two have often been conflated in prior work.

%lack of comprehensive evaluation and large scale benchmark
To date, there is still a lack of large-scale, reliable benchmarks grounded in authentic individual cognitive data for evaluating the extent to which LLMs simulate human cognition.
%Although role-playing datasets \cite{yin-etal-2025-charactercraft,xiang-etal-2025-rmtbench,qin-etal-2025-r} approximate human reasoning traces using LLMs, such synthetic annotations remain substantially distant from genuine reasoning processes.
Role-playing datasets \cite{yin-etal-2025-charactercraft, xiang-etal-2025-rmtbench, qin-etal-2025-r, xie2025human, wang2026humanllm} use LLMs to synthesize reasoning traces, which are externalized artifacts of cognitive processes \cite{newell1972human}, such approximations remain ontologically distinct from genuine human cognition.
Meanwhile, human-annotated reasoning benchmarks \cite{cobbe2021gsm8k,dalvi-etal-2021-explaining,jiang2025rexthinkergroundedobjectreferring} typically mix reasoning traces across different individuals, collapsing human cognitive patterns into population-level statistics \cite{sternberg1997thinking}, 
thereby breaking the consistency of the underlying human cognition.
%and making them unsuitable for evaluating cognitive simulation.
%This limitation stems not from a lack of interest, but from several fundamental challenges in obtaining and curating human cognitive data at scale.
These limitations stem from the following fundamental challenges in obtaining human cognitive data:
%Understanding and evaluating the alignment of human cognition remains a challenge for several reasons.
(1) \textit{Human cognition is inherently implicit}, where only its
behavioral artifacts, such as actions, are observable proxies \cite{ericsson1984protocol};
(2) \textit{Human cognition is heuristic and unstructured}, making it difficult to obtain a formalized representation \cite{gilovich2002heuristics};
(3) \textit{Externalized cognition is extremely hard to accumulate}, as humans rarely provide their internal thinking step-by-step in natural settings \cite{wurgaft2025scaling}, where collecting large amounts of consistent reasoning traces from the same person is even more difficult.
In this work, we take a further step toward modeling human cognition by identifying scientific publications \cite{zhou-etal-2025-hypothesis,xu2025probing} of a specific researcher as a scalable source. 
We observe that these publications, especially their introduction sections, naturally externalize the cognitive process by which an author solves scientific problems \cite{newell1972human}. 
Although such scientific documents are not raw cognitive streams \cite{van1994think} but textualized manifestations of human reasoning shaped by peer discussion, revision, and disciplinary conventions, they nevertheless provide a realistic approximation of an individual's cognitive processes,
revealing stable and consistent cognitive patterns that persist across different research domains and over time,
and thus form a suitable foundation for studying human cognitive alignment in LLMs.
%that reveal consistent cognitive patterns across research domains and over time,
%as they consistently address problems across different research domains and time periods, 
%making them a suitable foundation for studying and evaluating human cognitive alignment in LLM.

% In this work, we introduce a new perspective: scientific writing—particularly the Introduction section of research papers—provides a naturally structured and scalable externalization of human cognitive processes. Drawing on the Implicit Cognitive Cycle framework \citep{hofstadter1979godel, newell1990unified}, we observe that authors consistently engage in four key cognitive actions when crafting an Introduction:
% (1) Problem Representation,
% (2) Gap Analysis,
% (3) Strategy Selection, and
% (4) Solution Planning.
% These components form a coherent cognitive scaffold through which researchers articulate how they understand a problem, why the problem matters, and how they propose to address it. As such, scientific Introductions offer a unique opportunity to study human thinking in a structured yet naturalistic setting.

In detail, we construct a benchmark grounded in the longitudinal research trajectories of real researchers, building a personalized publication corpus for each of $217$ famous researchers, treating each author as a distinct cognitive subject. For each author, we organize their publications into topic-based clusters and adopt a cross-domain generalization setting, in which papers from earlier or established research topics serve as the source context for retrieval or learning, while papers from the author's most recent and previously unseen topic are reserved for testing. This design challenges LLMs to reason about new research problems in a novel domain while remaining cognitively consistent with the author's prior work. By enforcing topic disjointness between source and test domains, our benchmark deliberately limits the effectiveness of surface-level behavioral imitation, such as mimicking writing style or topic-specific methods. Instead, it compels LLMs to capture and transfer stable, individual-level cognitive patterns across domains, such as the personalized philosophy of problem identification and motivation formation. 
To faithfully evaluate cognitive consistency, we further introduce a multi-level evaluation framework that includes dedicated cognitive alignment metrics to assess whether the generated reasoning trace preserves consistent cognitive patterns with respect to the target author's memory, alongside conventional lexical and semantic alignment metrics \cite{zheng2023judging}.
As a result, we provide a rigorous benchmark to better distinguish genuine cognitive simulation from shallow behavioral mimicry.

Building on the framework introduced above, we systematically probe the extent to which LLMs can simulate human cognition. Our analysis is organized around two central questions: \textbf{(1) How well do current LLMs simulate human cognition?} and \textbf{(2) How far can existing techniques enhance these capabilities?} We examine a diverse set of state-of-the-art LLMs, including both open-source and API-based models, and analyze their reasoning patterns under carefully controlled domain-shift conditions. In addition, we evaluate a broad range of enhancement strategies, from theory-guided prompting techniques to supervised learning-based methods.
Overall, our contributions are:
\begin{itemize}[itemsep=-2pt, topsep=-2pt] %[itemsep=-3pt, topsep=-2pt]
    \item We revisit the long-standing problem of human simulation in artificial intelligence through the lens of modern LLMs, providing a systematic study at scale.
    %that provide the first systematic study at scale.
    %and provide the first systematic empirical study that distinguishes behavioral imitation from cognitive internalization at scale.
    \item We present a benchmark that combines authentic individual cognition histories with explicit domain shifts, facilitating controlled evaluation at the cognitive level.
    %We introduce a new benchmark that integrates authentic individual cognition histories with explicit domain shifts, enabling controlled cognitive evaluation.
    %for controlled evaluation of cognitive simulation.
    %, enabling controlled evaluation of high-dimensional cognitive simulation beyond surface-level behavioral alignment.
    \item We conduct a comprehensive analysis on state-of-the-art LLMs and representative enhancement techniques.
    %—including prompting and supervised learning—and provide fine-grained diagnostic insights into where current approaches succeed, where they fail, and why existing methods predominantly reinforce statistical fitting rather than transferable cognitive abstraction.
\end{itemize}

\section{Cognitive Benchmark Framework}
\label{sec:data_construction}
\begin{figure*}[t]
\centering
\includegraphics[width=0.90\linewidth]{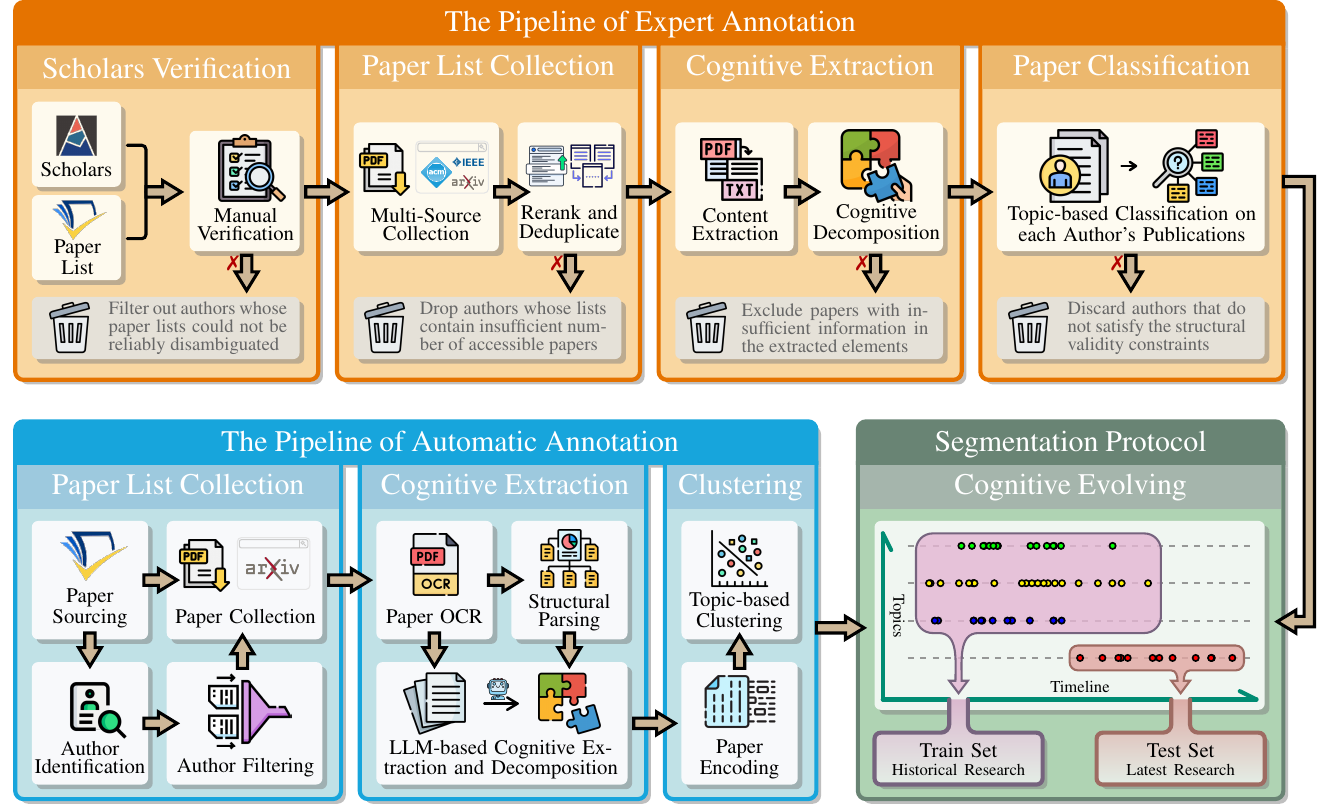}
%\captionsetup{skip=3pt}
\caption{
Overview of the construction pipeline for our benchmark. 
It consists of two separate lines: \textcolor{orange!90!black}{Expert Annotation}, which involves fine-grained verification and systematic quality filtering at each stage to ensure data reliability, and \textcolor{cyan!90!black}{Automatic Annotation}, which utilizes OCR and LLM-based extraction for expandable data collection.
The \textcolor{teal!90!green}{Segmentation Protocol} shows a splitting strategy based on the average publication dates of research topics, enabling cognitive evolution from the test set to the train set.
%Overview construction pipeline of our benchmark. The pipeline consists of two separate lines: \textcolor{orange!90!black}{Expert Annotation} that involves manual verification and quality control at each stage, and \textcolor{cyan!90!black}{Automatic Annotation} leveraging OCR and LLM-based extraction for scalable data collection. Both pipelines include systematic filtering steps to ensure data quality. The \textcolor{teal!90!green}{Segmentation Protocol} illustrates a chronological splitting strategy based on research topics, creating temporally ordered train and test sets to evaluate models across emerging research directions.
}
\vspace{-1pt}
\label{fig:pipeline}
\end{figure*}

\subsection{Overview}
\noindent\textbf{Principles} \;
Two core principles guide the construction of our benchmark:
One is to collect meta-level reasoning trajectories. Rather than merely collecting all reasoning components, annotators are instructed to focus more on the meta-level ones that may reflect human cognition. Besides, we deliberately source research papers from highly influential scholars, as their writings may contain more conceptual and principled thoughts.
The other is to prevent cheating through behavioral mimicry, where LLMs are not allowed to achieve high scores by superficially memorizing writing styles and tool usage. We explicitly partition the train and test sets with domain and temporal shifts to build imitation gap, with multi-dimensional cognitive evaluation metrics designed in addition to traditional lexical and semantic ones.
\noindent\textbf{Data Source} \; We identify prominent researchers in the field of artificial intelligence, filtering by the h-index and AMiner rankings\footnote{\url{https://www.aminer.cn/ranks/home}} to guarantee their influence and recognition. For each selected scholar, we collect papers from arXiv\footnote{\url{https://arxiv.org/}} and Semantic Scholar\footnote{\url{https://www.semanticscholar.org/}} in which they appear as the first author. 
Since it typically represents the primary intellectual contributions rather than collaborative efforts, this setting can make the reasoning trajectories of individual researchers more consistent. We employ $10$ senior Ph.D. students with specialized expertise in AI research for human annotation.

% We identify $217$ prominent researchers in the field of artificial intelligence, filtering by the h-index and AMiner rankings\footnote{\url{https://www.aminer.cn/ranks/home}} to guarantee their influence and recognition. For each selected scholar, we collect papers ($3422$ in total) from arXiv\footnote{\url{https://arxiv.org/}} and Semantic Scholar\footnote{\url{https://www.semanticscholar.org/}} in which they appear as the first author. 
% Since it typically represents the primary intellectual contributions rather than collaborative efforts, this setting can make the reasoning trajectories of individual researchers more consistent. Among them, papers from $100$ scholars are annotated by $10$ senior Ph.D. students with specialized expertise in AI research.

\begin{figure}[t]
\centering
\includegraphics[width=0.9\linewidth]{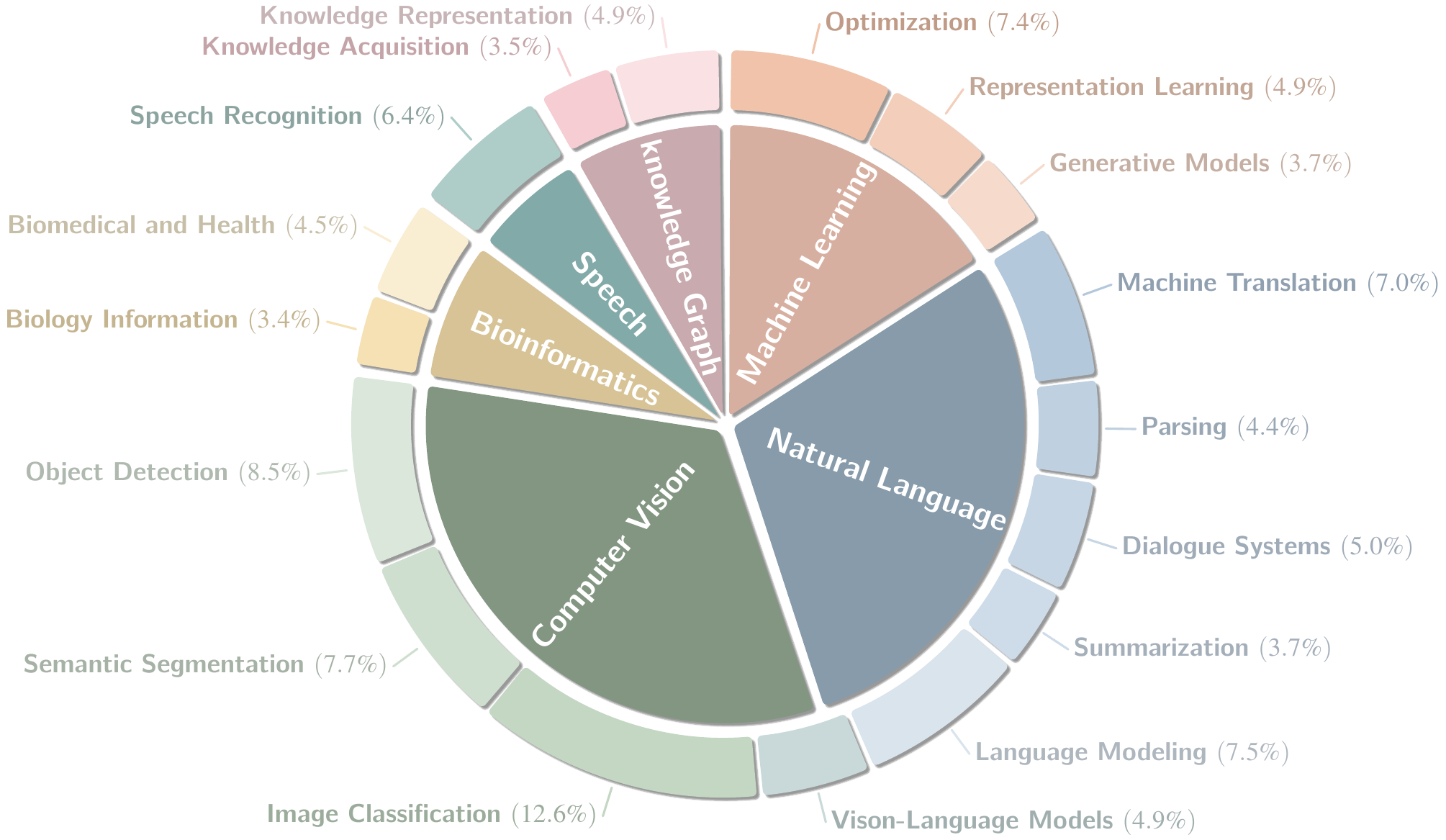}
\caption{Research topics in our benchmark.}
\vspace{-8pt}
\label{fig:piechart}
\end{figure}

\noindent\textbf{Topic Diversity} \; As shown in \Cref{fig:piechart}, our benchmark provides comprehensive coverage across AI subfields, organized into four major categories: Natural Language Processing, Computer Vision, Machine Learning, and related domains including Bioinformatics, Speech, and Knowledge Graphs. The distribution spans both foundational research areas, such as Optimization ($7.4\%$) and Representation Learning ($4.9\%$), and emerging frontiers including Vision-Language Models ($4.9\%$) and Language Modeling ($7.5\%$). This breadth ensures our evaluation captures diverse cognitive patterns across different research paradigms. 
%(需要补充：主题分布的设计考虑，为什么选择这样的比例分配)

% \begin{figure}[t]
% \centering
% \includegraphics[width=1\linewidth]{pics/timeline.pdf}
% \caption{Difference in Average Publication Year.}
% \label{fig:timeline}
% \end{figure}

% \begin{figure}[t]
% \centering
% \includegraphics[width=1\linewidth]{pics/train_test_scatter.pdf}
% \caption{Distribution of Research Topic.}
% \label{fig:topic_scatter}
% \end{figure}

% \begin{figure}[t]
% \centering
% \includegraphics[width=1\linewidth]{pics/variance_distribution.pdf}
% \caption{Difference in Research Topic.}
% \label{fig:topic_distance}
% \end{figure}

%\section{Cognitive Benchmark Construction}
%%%%%%%%%@llj\section{Data Construction}

\subsection{Dataset Collection and Annotation}
%To construct a reliable benchmark for cognitive simulation, we introduce a dual-stream data construction framework that balances high-fidelity evaluation with scalable training. 

%As illustrated in Figure \ref{fig:pipeline}, our framework consists of two parallel pathways: (1) an \textit{Expert-Annotated Stream} designed to establish a rigorous upper bound for cognitive alignment, and (2) an \textit{Automated Scalable Stream} to verify method generalization on large-scale data. 

As illustrated in \Cref{fig:pipeline}, we introduce a dual-stream data construction framework that balances high-fidelity cognitive alignment (\textcolor{orange!90!black}{Expert Annotation}) with scalable generalization (\textcolor{cyan!90!black}{Automatic Annotation}).
Both streams employ a \textcolor{teal!90!green}{Segmentation Protocol} to maximize domain divergence between train and test sets while maintaining consistent cognitive patterns.

\noindent\textbf{The Pipeline of Expert Annotation}\;
This stream focuses on constructing a high-fidelity dataset through strict manual verification and filtering. 
(1) \textbf{Scholars Verification}: We first anchor our selection on the AMiner rankings and cross-reference candidates with verified profiles on Semantic Scholar. Any author associated with more than five unverifiable papers is immediately discarded to prevent identity noise of author name ambiguity.
(2) \textbf{Paper List Collection}: Then we retrieve PDFs from multiple sources, such as arXiv, and deduplicate different versions of the same paper. Authors with insufficient valid publications are dropped to ensure adequate cognitive data for each individual.
(3) \textbf{Cognitive Extraction}: Annotators manually extract four critical cognitive elements of scientific exploration from the Introduction section: \textit{Background}, \textit{Motivation}, \textit{Approach}, and \textit{Solution}. These four elements correspond to the components of the human problem-solving process \cite{newell1972human}: problem representation, goal identification, operator selection, and execution \& evaluation, respectively. In addition, elements are required to be extracted directly from the source text to prevent subjective hallucination. We also apply a content-length filter to exclude papers with insufficient information.
(4) \textbf{Paper Classification}: Finally, annotators assign research topics to papers and group them as topic-based collections for each author. To ensure a meaningful assessment of cognitive evolution, we will discard authors that do not satisfy the structural validity constraints: each author should have publications across at least two distinct topics, with a minimum of three papers per collection.

\noindent\textbf{The Pipeline of Automatic Annotation}\;
% \subsection{Stream B: The Automated Scalable Pipeline}
This stream focuses on the scalability of the dataset with an automatic extraction and resolution process.
%is designed to validate the scalability of our approach using a metadata-driven pipeline.
(1) \textbf{Paper List Collection}: First, we employ an API-based paper-to-author collection strategy, where we source all computer science papers from 2022 to 2024, identify their first authors, and apply filtering to remove ambiguous entries. 
This produces a set of authors and their corresponding first-authored publications, which also enables efficient scaling to construct large-scale author datasets across other research domains.
(2) \textbf{Cognitive Extraction}: Then, we jointly use the \texttt{Grobid} \cite{10.1007/978-3-642-04346-8_62} and \texttt{MinerU} \cite{niu2025mineru25decoupledvisionlanguagemodel} tools to recognize text from PDFs, ensuring robustness across different formats. 
Structural parsing is further applied to identify the introduction section, where \texttt{GPT-5-nano} segments the section and extracts the four cognitive elements.
%\texttt{Grobid}\cite{GROBID} and \texttt{MinerU} \cite{niu2025mineru25decoupledvisionlanguagemodel}
(3) \textbf{Clustering}: Finally, we use \texttt{GPT-5-nano} to summarize papers into research topics based on their introductions, obtaining the corresponding encodings with \texttt{text-embedding-3-large}.
K-means clustering with multiple centroid numbers is employed for each author to construct the most suitable topic-based paper collections that satisfy structural validity constraints, similar to those in expert annotation.

% \paragraph{Reverse Screening \& First-Author Constraint.}
% We employ an API-driven "Paper-to-Author" sourcing strategy, targeting active CS researchers from 2022 to 2024. Crucially, to mitigate stylistic noise from collaborators, we enforce a \textbf{First-Author Only} constraint. This ensures that the extracted writing style reflects the target researcher's pure cognitive signature rather than a mixture of co-authors.

% \paragraph{Hybrid Structural Parsing.}
% We utilize a two-stage parsing mechanism to extract Introduction sectionfrom PDFs:
% \begin{itemize}
%     \item \textbf{Primary Parsing:}  is used for high-efficiency extraction on standard academic layouts.
%     \item \textbf{High-Precision Fallback:} For complex layouts where Grobid fails, we utilize , a visual-based parsing tool, to ensure high recall rates for long-tail data.
% \end{itemize}
% Subsequently, \texttt{GPT-4o-mini} is prompted to segment the text into the standard four cognitive components (Background, Motivation, Approach, Solution).

% \paragraph{TopicGPT-based Clustering.}
% To mimic human clustering, we adopt a TopicGPT paradigm:
% (1) \textbf{Topic Generation:} LLMs generate concise topic labels for each paper based on title and abstract.
% (2) \textbf{Embedding \& Clustering:} Topic labels are encoded using \texttt{text-embedding-3-large} and grouped via K-Means. We align the clustering hyperparameters (min cluster size $\ge$ 3, min clusters $\ge$ 2) with the expert pipeline to ensure structural consistency.

\begin{figure}[t]
\centering
\captionsetup{skip=3pt}
\includegraphics[width=0.9\linewidth]{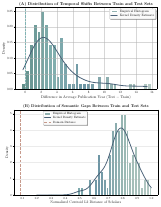}
\caption{Characterization of temporal and semantic variance between train and test sets across scholars. 
(A) Distribution of temporal shifts, quantified as the difference in average publication year (test minus train). Almost all authors' test sets are more recent than their train sets, with the distribution peaking at about $2$ years. 
(B) Distribution of semantic gaps, measured by the normalized centroid L2 distance between each author's train and test set embeddings. The dashed line indicates the baseline distance between the global centroids of all train and test sets. It reveals substantial within-author semantic distance concentrated between $0.7$ and $0.9$.}
\vspace{-8pt}
\label{fig:timeline}
\end{figure}

\noindent\textbf{Segmentation Protocol}\; We adopt a chronological strategy as the segmentation protocol based on the evolution of research topics. 
Paper collections of each author are sorted by average publication date, with the most recent cluster designated as the test set, representing the author's latest research direction.
Other earlier topics, capturing the author's historical research, form the train set that serves for model training, retrieval augmentation, and the construction of author profiles including memory and cognitive patterns.
As illustrated in \Cref{fig:timeline}, this segmentation ensures a cross-domain, temporal-shift generalization setting.

% \subsection{Unified Topic-Evolving Split Protocol}

\subsection{Evaluation Framework}
\begin{figure}[t]
\centering
\includegraphics[width=0.95\linewidth]{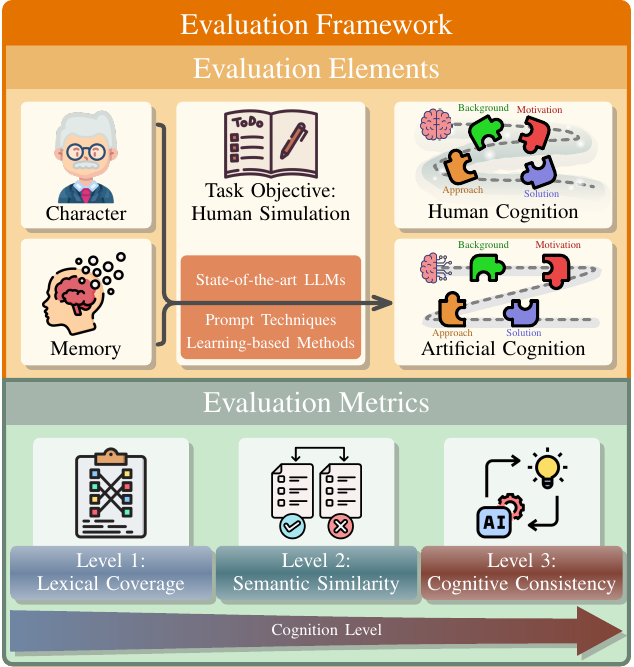}
\caption{Overview of our evaluation framework.}
\label{fig:evaluation}
\end{figure}

To systematically assess the fidelity of LLM-based human simulation, we propose a multi-level evaluation framework as illustrated in \Cref{fig:evaluation}. Given a scholar's memory of historical research, we instruct state-of-the-art LLMs, augmented with various prompting techniques and learning-based methods, to generate cognitive processes that solve problems in the same manner as the scholar. We evaluate the consistency between the artificial and authentic human cognitive processes across three hierarchical levels of increasing cognitive depth.
(1) \textbf{Lexical Coverage}: We use \texttt{ROUGE} \cite{lin-2004-rouge} scores to measure the recall of key terms and phrases from the reference cognitive process and \texttt{BertScore} \cite{zhang2019bertscore} to identify optimal token-level matches. 
(2) \textbf{Semantic Similarity}: We utilize the LLM-as-a-judge framework, where \texttt{GPT-5-nano} is prompted to score the alignment degree between generated cognitive processes and references.
(3) \textbf{Cognitive Consistency}: Since it is a universal feature of human cognition that tends to maintain long-term, stable patterns across scenarios \cite{messick1984nature}, we propose five evaluation dimensions corresponding to different levels of cognitive consistency.
Based on Witkin's field dependence-independence theory \cite{witkin1977field}, which demonstrates the perceptual and structural consistency of an individual’s cognition, we design the dimensions of foundational premise consistency and logic mapping consistency.
%, such as reliance on a frame of reference and a tendency toward cognitive restructuring, 
Besides, Kirton's adaption-innovation theory \cite{kirton1976adaptors} posits that individuals exhibit stable strategic preferences in problem-solving, where we introduce the consistency dimensions of strategic preference, keyword style, and taboo avoidance.
In addition to comparisons with human cognition, the consistency of these cognitive dimensions also needs to be examined in relation to scholars' memories, where their details are as follows: (a) \textit{Foundational Premise}: whether the generation inherits the author's consistent perspective and problem framing, e.g., persistently identifying research questions from data noise;
(b) \textit{Strategic Preference}: decision-making tendencies during technical trade-offs, e.g., favoring generalizable theoretical frameworks over practical applications;
(c) \textit{Logic Mapping}: whether the reasoning inherits the author's logical patterns, e.g., consistently employing symbolic formalization when confronting complex problems;
(d) \textit{Keyword Style}: lexical preferences and expression habits, e.g., usually using ``feature manifold'' rather than ``embedding space'';
and (e) \textit{Taboo Avoidance}: whether approaches the author has historically criticized are appropriately avoided, e.g., a researcher who critiques assumption-free brute-force fitting should not propose collecting data for direct training without theoretical grounding.
For both semantic and cognitive evaluation, we use \texttt{GPT-5-nano} to annotate responses on a $5$-point scale \cite{likert1932technique}. A score of $5$ denotes high consistency between the generated and human cognition or the author's cognitive patterns on the given dimension, $3$ indicates moderate correspondence, and $1$ means complete irrelevance.

\subsection{Dataset Statistics}
\label{sec:stats}

% 更新成真实数据了
\begin{table}[t]
  \centering
  \small % 稍微缩小字体以适应版面
  \setlength{\tabcolsep}{5pt} % 调整列间距
  \caption{Statistics of our benchmark.}
  \label{tab:stats_summary_total}
  \begin{tabular}{l|cc|c}
    \toprule
    \textbf{Statistic} & \textbf{Expert} & \textbf{Automatic} & \textbf{Total} \\
    \midrule
    \# Scholars & 100 & 117 & 217 \\
    \# Total Papers & 1,875 & 1,547 & 3,422 \\
    Papers per Author & 18.8 & 13.2 & 15.8 \\
    Clusters per Author & 2.3 & 2.3 & 2.3 \\
    \midrule
    \textit{Tokens per Component} & & & \\
    \quad Introduction & 614.9 & 637.2 & 625.0 \\
    \quad Background & 178.9 & 99.0 & 142.8 \\
    \quad Motivation & 88.2 & 55.1 & 73.2 \\
    \quad Approach & 107.2 & 66.2 & 88.7 \\
    \quad Solution & 88.6 & 66.7 & 78.7 \\
    \bottomrule
  \end{tabular}
\end{table}

As illustrated in \Cref{tab:stats_summary_total}, compared to automatic extraction, expert annotations capture richer cognitive signals with longer component lengths, with the most notable being Background ($178.9$ vs. $99.0$ tokens), which reflects the authors' philosophy of problem formulation. 
Such granularity variance motivates our experimental design: we focus our primary experiments and evaluations on the expert-annotated data to ensure reliable conclusions, where the automatic one demonstrates the potential for large-scale weak supervision in future work. More details of the two annotation methods are shown in Appendix \ref{appsec:stats}.

\section{Experiments}
\begin{table*}[t]
  \centering
  \captionsetup{skip=3pt}
  \caption{Main results of simulating human cognition across different LLMs. 
  We evaluate five scenarios: (1) \textbf{\textsc{Base Model}} represents the original output without any cognition history conditioning. (2) \textbf{\textsc{Other}} uses cognition histories randomly sampled from other authors, serving as an interference control group. (3) \textbf{\textsc{Self}} incorporates the target author's cognition history from the train set, representing our standard cross-domain setting. (4) \textbf{\textsc{SelfAkin}} leverages cognition histories from the same author's test set, examining in-domain behavioral leakage. (5) \textbf{\textsc{Leaky}} provides oracle access to the author's authentic cognition trajectory, serving as an upper bound. 
  %Differences across these settings allow us to probe whether performance gains arise from genuine cognitive abstraction or from increasingly strong forms of behavioral leakage.
  We report \textbf{ROUGE-1/2/L} and \textbf{BertScore} for lexical coverage, \textbf{LLM-as-a-Judge} for semantic similarity, and five cognitive consistency dimensions: Foundational \textbf{P}remise, \textbf{S}trategic Preference, \textbf{L}ogic Mapping, \textbf{K}eyword Style, and \textbf{T}aboo Avoidance. See Appendix \ref{apptab:exp} for more LLMs.}
  \begin{tabular}{l|cc|c|lllll}
    \toprule
    \midrule
    \multirow{2}{*}{\textbf{Method}} & \multicolumn{2}{c|}{\textbf{Lexical}} & \multirow{2}{*}{\textbf{Semantic}} & \multicolumn{5}{c}{\textbf{Cognitive}}\\
    &\textbf{ROUGE-1/2/L} & \textbf{BertScore} &  & \textbf{\;\;P.} & \textbf{\;\;S.} & \textbf{\;\;L.} & \textbf{\;\;K.} & \textbf{\;\;T.}\\
    \bottomrule
    
    \midrule
    \textsc{LLaMA-3.1} & $22.12/3.91/12.02$ &	$81.39$	& $3.22$ & $3.03$ & $3.15$ & $3.00$ &  $3.05$ & $3.20$\\
    %\rowcolor{blue!20!white}
    \textsc{- \small{Other}} & $23.11/3.94/12.70$ & $81.63$ & $2.94$ & $2.74\!\,\color{red!50!black}{\downarrow}$ & $2.95\!\,\color{red!50!black}{\downarrow}$ & $2.73\!\,\color{red!50!black}{\downarrow}$ &  $2.89\!\,\color{red!50!black}{\downarrow}$ & $3.09\!\,\color{red!50!black}{\downarrow}$\\
    \rowcolor{cyan!40!green!10!white}\textsc{- \small{Self}} & $24.10/4.45/13.10$ & $81.89$ & $3.13$ & $3.02$ & $3.23\!\,\color{green!50!black}{\uparrow}$ & $2.98$ &  $3.06$ & $3.32\!\,\color{green!50!black}{\uparrow}$\\
    \rowcolor{red!40!orange!10!white}\textsc{- \small{SelfAkin}} & $25.65/5.93/14.12$ & $82.36$ & $3.44$ & $3.55\!\,\color{green!50!black}{\upuparrows}$ & $3.77\!\,\color{green!50!black}{\upuparrows}$ & $3.41\!\,\color{green!50!black}{\upuparrows}$ & $3.34\!\,\color{green!50!black}{\upuparrows}$ & $3.82\!\,\color{green!50!black}{\upuparrows}$\\
    \textsc{- \small{Leaky}} & $26.70/6.79/14.75$ & $82.66$ & $3.63$ & $3.77\!\,\color{green!50!black}{\upupuparrows}$ & $3.86\!\,\color{green!50!black}{\upupuparrows}$ & $3.55\!\,\color{green!50!black}{\upupuparrows}$ & $3.45\!\,\color{green!50!black}{\upupuparrows}$ & $3.92\!\,\color{green!50!black}{\upupuparrows}$\\
    \midrule
    \midrule
    \textsc{Qwen-3} & $23.08/3.25/11.00$ & $81.21$ & $3.41$ & $3.51$ & $3.66$ & $3.44$ & $3.36$ & $3.62$\\
    %\rowcolor{blue!20!white}
    \textsc{- \small{Other}} & $21.95/2.89/10.66$ & $81.30$ & $3.19$ & $3.28\!\,\color{red!50!black}{\downarrow}$ &	$3.53\!\,\color{red!50!black}{\downarrow}$ & $3.26\!\,\color{red!50!black}{\downarrow}$ & $3.15\!\,\color{red!50!black}{\downdownarrows}$ & $3.60\!\,\color{red!50!black}{\downarrow}$ \\
    \rowcolor{cyan!40!green!10!white}\textsc{- \small{Self}} & $22.14/3.18/10.73$ & $81.44$ & $3.33$ & $3.51$ & $3.72\!\,\color{green!50!black}{\uparrow}$ & $3.45$ &  $3.27\!\,\color{red!50!black}{\downarrow}$ & $3.83\!\,\color{green!50!black}{\uparrow}$ \\
    \rowcolor{red!40!orange!10!white}\textsc{- \small{SelfAkin}} & $23.21/3.81/11.13$ & $81.76$ & $3.59$ & $4.07\!\,\color{green!50!black}{\upuparrows}$ & $4.26\!\,\color{green!50!black}{\upuparrows}$ & $3.96\!\,\color{green!50!black}{\upuparrows}$ & $3.68\!\,\color{green!50!black}{\upuparrows}$ & $4.17\!\,\color{green!50!black}{\upuparrows}$\\
    \textsc{- \small{Leaky}} & $24.06/4.42/11.43$ & $82.04$ & $3.83$ & $4.29\!\,\color{green!50!black}{\upupuparrows}$ & $4.38\!\,\color{green!50!black}{\upupuparrows}$ & $4.08\!\,\color{green!50!black}{\upupuparrows}$ &  $3.80\!\,\color{green!50!black}{\upupuparrows}$	& $4.31\!\,\color{green!50!black}{\upupuparrows}$ \\
    \midrule
    \midrule
    \textsc{Gemini-2.5} & $18.55/3.48/\phantom{0}9.05$ & $81.85$ & $3.48$ & $3.81$ & $3.91$ & $3.84$ & $3.46$ & $4.04$\\
    \textsc{- \small{Other}} & $19.33/3.36/\phantom{0}9.41$ & $81.65$ & $3.27$ & $3.54\!\,\color{red!50!black}{\downdownarrows}$ & $3.76\!\,\color{red!50!black}{\downarrow}$ & $3.64\!\,\color{red!50!black}{\downdownarrows}$ & $3.31\!\,\color{red!50!black}{\downdownarrows}$ & $3.92\!\,\color{red!50!black}{\downarrow}$\\
    \rowcolor{cyan!40!green!10!white}\textsc{- \small{Self}} & $19.77/3.63/\phantom{0}9.59$ & $81.92$ & $3.42$ & $3.78\!\,\color{red!50!black}{\downarrow}$ & $3.91$ & $3.81\!\,\color{red!50!black}{\downarrow}$ & $3.42\!\,\color{red!50!black}{\downarrow}$ & $4.13\!\,\color{green!50!black}{\uparrow}$\\
    \rowcolor{red!40!orange!10!white}\textsc{- \small{SelfAkin}} & $20.51/4.02/\phantom{0}9.91$ & $82.11$ & $3.59$ & $4.10\!\,\color{green!50!black}{\uparrow}$ & $4.16\!\,\color{green!50!black}{\uparrow}$ & $4.02\!\,\color{green!50!black}{\uparrow}$ & $3.56\!\,\color{green!50!black}{\uparrow}$ & $4.28\!\,\color{green!50!black}{\uparrow}$\\
    \textsc{- \small{Leaky}} & $21.72/4.64/10.43$ & $82.46$ & $3.86$ & $4.27\!\,\color{green!50!black}{\upuparrows}$ & $4.28\!\,\color{green!50!black}{\upuparrows}$ & $4.15\!\,\color{green!50!black}{\upuparrows}$ & $3.73\!\,\color{green!50!black}{\upuparrows}$ & $4.43\!\,\color{green!50!black}{\upuparrows}$\\
    \midrule
    \midrule
    \textsc{GPT-5.1} & $\phantom{0}9.52/1.65/\phantom{0}4.73$ & $80.65$ & $3.64$ & $4.16$ & $4.39$ & $4.12$ & $3.81$ & $4.40$\\
    \textsc{- \small{Other}} & $\phantom{0}9.32/1.61/\phantom{0}4.69$ & $80.61$ & $3.49$ & $4.08\!\,\color{red!50!black}{\downarrow}$ & $4.33\!\,\color{red!50!black}{\downarrow}$ & $4.00\!\,\color{red!50!black}{\downarrow}$ & $3.77\!\,\color{red!50!black}{\downarrow}$ & $4.32\!\,\color{red!50!black}{\downarrow}$\\
    \rowcolor{cyan!40!green!10!white}\textsc{- \small{Self}} & $\phantom{0}9.35/1.70/\phantom{0}4.74$ & $80.71$ & $3.62$ & $4.26\!\,\color{green!50!black}{\uparrow}$ & $4.45\!\,\color{green!50!black}{\uparrow}$ & $4.14$ & $3.84\!\,\color{green!50!black}{\uparrow}$ & $4.42$\\
    \rowcolor{red!40!orange!10!white}\textsc{- \small{SelfAkin}} & $\phantom{0}9.81/1.88/\phantom{0}4.89$ & $80.86$ & $3.72$ & $4.51\!\,\color{green!50!black}{\upuparrows}$ & $4.63\!\,\color{green!50!black}{\upuparrows}$ & $4.34\!\,\color{green!50!black}{\upuparrows}$ & $3.99\!\,\color{green!50!black}{\upuparrows}$ & $4.52\!\,\color{green!50!black}{\upuparrows}$\\
    \textsc{- \small{Leaky}} & $10.08/2.11/\phantom{0}5.05$ & $81.15$ & $3.99$ & $4.64\!\,\color{green!50!black}{\upupuparrows}$ & $4.69\!\,\color{green!50!black}{\upuparrows}$ & $4.48\!\,\color{green!50!black}{\upupuparrows}$ & $4.08\!\,\color{green!50!black}{\upupuparrows}$ & $4.61\!\,\color{green!50!black}{\upupuparrows}$\\
    \midrule
    \bottomrule
  \end{tabular}
  \label{tab:0}
  \vspace{-4pt}
\end{table*}

\subsection{Overview}
%Our Objective
%Experimental Setting
%This section presents comprehensive analyses structured around core questions on the extent to which LLMs can simulate human cognition and how far such capabilities can be enhanced by existing techniques.
This section presents comprehensive analyses structured around two core research questions regarding the extent to which LLMs can simulate human cognition and how far such capabilities can be enhanced by existing techniques. For the former question, we design a controlled evaluation scenario with systematic ablation variants across multiple LLMs, as illustrated in \Cref{tab:0} and \S\ref{q1}. Specifically, we examine how varying levels of information availability and cognitive leakage affect the fidelity of human cognitive simulation.  For the latter one, we conduct extensive experiments evaluating various enhancement techniques across different model architectures in \Cref{tab:1} and \S\ref{q2}.

We experiment on a wide range of state-of-the-art LLMs, including \textsc{GPT-5.1} \cite{openai_gpt5_1}, \textsc{Gemini-2.5} \cite{comanici2025gemini}, \textsc{Qwen-3-8B} \cite{yang2025qwen3}, and \textsc{LLaMA-3.1-8B} \cite{dubey2024llama}, with additional LLMs reported in the appendix. 
Beyond base models, we evaluate both prompting-based and learning-based enhancement techniques. Prompting strategies leverage Big Five personality traits \cite{jiang-etal-2024-personallm,dai2025profile,Zhu2025CanLI} and explicit cognitive modeling guidance \cite{qin-etal-2025-r,Zhan2025TestTimeMatchingDP}, while learning-based approaches \cite{Tang2024StepBackPD,dai2025profile} either train separate parameterizations for individual scholars or jointly optimize a unified model conditioned on structured scholar profiles.

\subsection{Results on Cognitive Simulation in LLMs}
\label{q1}

\begin{template1}{How Well Do LLMs Simulate Human Cognition?}
\texttt{Question 1:}\vspace{1pt}\\
\hspace*{5pt}\textit{Can LLMs Achieve Behavioral Simulation?}
\tcblower
\texttt{Answer:} \textcolor{green3!60!green}{Yes}.\vspace{1pt}\\
\textit{Most state-of-the-art LLMs are capable of achieving behavioral simulation to a considerable extent.}
\end{template1}

Our results provide consistent evidence that modern LLMs are capable of behavioral simulation, i.e., reproducing observable, individual-specific decisions and actions. 
%First, the systematic performance gap between \textsc{Self} and \textsc{Other} demonstrates that models can reliably distinguish a target scholar's historical research from randomly sampled authors. 
Across all evaluated backbones, conditioning on the target scholar's research history (\textsc{Self}) yields clear improvements over randomly sampled authors (\textsc{Other}), with increases of approximately $0.13$ to $0.19$ in semantic scores and $0.07$ to $0.28$ in cognitive facets. 
%For example, LLaMA3.1-8B improves from 2.94 to 3.13 in semantic score (+0.19), while Gemini2.5-Pro and GPT-5.1 exhibit similar gains (e.g., GPT-5.1 semantic: 3.49→3.62, +0.13). 
Such gaps (\textsc{Self} – \textsc{Other}) indicate that LLMs are sensitive to author identity and can condition on personalized behavioral cues rather than producing generic outputs.
Given in-distribution, recent scholar histories (\textsc{SelfAkin}), LLMs show substantially larger gains than base outputs, improving cognitive facets by $0.10$ to $0.62$. 
%semantic scores by $0.08$ to $0.22$ and cognitive facets by $0.10$ to $0.62$. 
%It suggests that when strong distributional overlap enables behavior imitation, LLMs can effectively amplify alignment with the target scholar's observable patterns.
%Across backbones, \textsc{SelfAkin} improves semantic scores by approximately +0.10 to +0.22 and cognitive facets by +0.30 to +0.62 compared to the base setting. For instance, Qwen3-8 B's cognitive P. score increases from 3.51 to 4.07 (+0.56), while LLaMA3.1-8B exhibits gains of +0.52 on P. (3.03→3.55) and +0.62 on T. (3.20→3.82). Similar trends are observed for Gemini2.5-Pro and GPT-5.1, suggesting that when strong distributional overlap enables behavior imitation, LLMs can effectively amplify alignment with the target scholar's observable patterns.
% rather than deeper cognitive internalization. %\textcolor{red}{X}\% of cases are rated as behavior simulation, compared to only \textcolor{red}{Y}\% exhibiting evidence of process-level reasoning alignment.
%Human evaluation in \Cref{tab:human} also illustrates that annotators predominantly characterize LLM outputs as reflecting behavior-level imitation with at least $88\%$ of cases rated as behavior simulation
%Human evaluation in \Cref{tab:human} shows that annotators predominantly characterize LLM outputs as behavior-level imitation, with at least $88\%$ of cases rated as B or C.
Human evaluation in \Cref{tab:human} shows that LLM outputs, whether referencing historical or recent in-distribution research (\textsc{Self} and \textsc{SelfAkin}), are predominantly characterized by annotators as behavior-level imitation, with at least $89\%$ of cases rated as \textbf{B} or \textbf{C}.
%, including at least $31\%$ in \textsc{Self} reflecting attempts to model cognitive processes but remaining at the behavioral level.
%, while \textcolor{red}{Y}\% cases reflect attempts to model cognitive processes but fail and remain at the level of behavioral imitation. 
%; see Appendix~\placeholder{A}). Taken together, automatic metrics and human judgments converge on an affirmative answer to Q1: current LLMs can achieve behavioral simulation, though this capability primarily manifests at the level of observable behavior rather than internal cognitive processes.
\begin{table}[t]
  \centering
  \captionsetup{skip=4pt}
  \caption{Human Evaluation. Three senior Ph.D. students evaluate 100 papers per method (one randomly sampled per scholar). Annotation scales: \textbf{A} is no relation to author's cognition; \textbf{B} is simple imitation of historical research; \textbf{C} is attempted but failed cognitive simulation, remaining behavioral; \textbf{D} is successful thinking pattern transfer. Inter-annotator agreement (Fleiss' kappa): $0.43$.}
  \begin{tabular}{l|cccc}
    \toprule
    \midrule
    \textbf{Method} & \textbf{A} & \textbf{B} & \textbf{C} & \textbf{D}\\
    \bottomrule
    
    \midrule
    %\multicolumn{5}{l}{\small\textsc{Qwen-3-235B}} \\
    \small{\textsc{Qwen-3-235B}} \textsc{\;\,- \tiny{Self}} & $11\%$ & $58\%$ & $31\%$ & $\phantom{0}0\%$\\
    \qquad \qquad \quad \textsc{- \tiny{SelfAkin}} & $\phantom{0}3\%$ & $35\%$ & $60\%$ & $\phantom{0}1\%$\\
    \midrule
    \midrule
    %\multicolumn{5}{l}{\small\textsc{GPT-5.1}} \\
    \small{\textsc{GPT-5.1}}\textsc{\;\,- \tiny{Self}} & $ \phantom{0}9\%$ & $51\%$ & $38\%$ & $\phantom{0}2\%$\\
    \qquad\;\,\textsc{- \tiny{SelfAkin}} &  $\phantom{0}3\%$ & $26\%$ & $67\%$ & $\phantom{0}4\%$\\
    \midrule
    \bottomrule
  \end{tabular}
  \label{tab:human}
  \vspace{-5pt}
\end{table}

\begin{template1}{How Well Do LLMs Simulate Human Cognition?}
\texttt{Question 2:}\vspace{1pt}\\
\hspace*{5pt}\textit{Can LLMs Achieve Cognitive Internalization?}
\tcblower
\texttt{Answer:} \textcolor{red2!60!red}{Not yet}.\vspace{1pt}\\
\textit{Current state-of-the-art LLMs remain distant from achieving reliable cognitive internalization.}
%\textit{Current state-of-the-art LLMs still exhibit a substantial gap in achieving cognitive internalization.}
\end{template1}

% Despite their success in behavioral simulation, our results reveal clear limitations in cognitive internalization. If models had truly internalized abstract cognitive mechanisms, performance under SELF should approach that of LEAKY, where the true cognition process is directly exposed.

% However, we consistently observe a substantial LEAKY–SELF gap across all models and cognitive dimensions. Even for the strongest systems, oracle access to true cognition trajectories yields markedly higher scores than those achievable under SELF. This gap indicates that models struggle to transfer or abstract latent cognitive structures from historical observations alone.

% Moreover, the gap does not diminish significantly with model scale or capability, suggesting that the limitation is not merely a matter of capacity. Instead, it reflects a deeper challenge: current LLMs largely rely on surface-level correlations rather than forming transferable, internal cognitive representations.

% Therefore, while LLMs can reproduce what humans do, they still fail to internalize how humans think, leading to a negative answer to Q2.
%Despite simulating observable behavior, 
Our results indicate that current LLMs largely fail at cognitive internalization, i.e., abstracting and transferring a scholar's latent cognitive processes across domains. 
%A first piece of evidence comes from the substantial gap between \textsc{SelfAkin} and \textsc{Self}. 
%While \textsc{SelfAkin} provides the model with in-distribution, same-domain scholar histories, \textsc{Self} requires the model to extract cross-domain commonalities from the same scholar's prior work. 
%Across all evaluated backbones, this gap 
In detail, the discrepancy between providing the latest in-domain research (\textsc{SelfAkin}) and the out-of-domain research histories (\textsc{Self}) is consistently large, with cognitive facets under \textsc{SelfAkin} outperforming \textsc{Self} by $0.33$ on average, suggesting that LLMs struggle to identify and internalize higher-level cognitive invariants shared across a scholar's work in different research areas.
%indicating that LLMs struggle to generalize cognitive traits beyond domain-specific behavioral similarity. This suggests that LLMs fail to identify and internalize higher-level cognitive invariants shared across a scholar's work in different research areas.
%Besides, even when in-domain behavioral signals are available, a pronounced gap remains between \textsc{Leaky} and \textsc{SelfAkin}. 
% Besides, after revealing the scholar's actual cognition trajectories (\textsc{Leaky}), LLMs achieve substantial additional gains over merely referencing recent in-domain work \textsc{SelfAkin}, with cognitive scores increasing by $0.25$ to $0.55$ across base models. This gap indicates that the reasoning processes implicitly modeled by LLMs remain far from the scholar's actual cognitive processes, even under favorable conditions. 
Moreover, revealing the scholar's cognition trajectories (\textsc{Leaky}) yields substantial gains over referencing recent in-domain work (\textsc{SelfAkin}),
%, with cognitive scores increasing by $0.25$ to $0.55$ across base models,
which further indicates that current LLMs' attempts to imitate human cognition remain far from the level of authentic human reasoning.
%In other words, what models learn from behaviorally similar contexts does not approximate the latent reasoning mechanisms that generate those behaviors.
Human evaluation further supports these findings,
%. Annotators observe that, 
while LLMs usually ($51\%$ of the cases on average belong to \textbf{C} or \textbf{D}) attempt to capture underlying cognitive processes, the vast majority ($95.6\%$) fail and consequently collapse to behavior-level imitation. Instances of successful cognitive internalization are exceedingly scarce.
Notably, when referencing out-of-distribution historical research, LLMs exhibit a \textit{lazy} tendency that bypasses the effortful extraction of meta-level human cognition and instead takes shortcuts by simply mimicking specific technical approaches or experimental setups from the references or even disregarding them entirely, which is evidenced by \textsc{Self} having substantially higher \textbf{A} or  \textbf{B} ratings ($31\%$ on average) than \textsc{SelfAkin}.
%This is evidenced by the substantially higher A+B ratings in out-of-distribution settings, suggesting that models take cognitive shortcuts when faced with less familiar reference patterns.

%instead take shortcuts by simply mimicking specific technical approaches or experimental setups from the references, or disregarding them entirely, as shown by the predominance of A and B ratings.

%When references are out-of-distribution, models exhibit a lazy tendency, bypassing the effortful extraction of meta-level cognition and instead defaulting to superficial behavioral imitation or even ignoring the references entirely. This laziness is reflected in the substantially higher

%Human evaluation further supports these findings. Annotators observe that, while LLMs occasionally attempt to model underlying cognitive processes, the vast majority of such attempts fail and ultimately revert to behavior-level imitation. Instances of successful cognitive internalization are exceedingly scarce (\textcolor{red}{Z}%).

%Taken together, these results lead to a negative answer to Q2: current LLMs do not reliably achieve cognitive internalization. Even when provided with rich historical signals or strong in-distribution cues, models primarily rely on behavioral similarity rather than abstracting and transferring latent cognitive processes.

\subsection{Results on Enhancing Cognitive Simulation}
\label{q2}
\begin{table*}[t]
  \centering
  \captionsetup{skip=3pt}
  \caption{Evaluation of enhancement techniques for human cognitive simulation. 
  We assess the effectiveness of various augmentation strategies on two open-source LLMs: \textsc{LLaMA-3.1-8B} and \textsc{Qwen-3-8B}. 
  The base model output and the \textsc{Self} condition are baselines. 
  We evaluate two categories of enhancement techniques: (1) Prompting techniques, including \textsc{PersonalityTrait} that incorporates personality traits as psychological priors, and \textsc{CognitiveGuide} which explicitly instructs LLMs with cognitive theories; (2) Learning-based methods, including \textsc{TrainIndividual} which trains independent parameters for each scholar, and \textsc{TrainUnified} that extracts profiles for scholars as their identifiers to train a unified model. The evaluation metrics are shared with those used in \Cref{tab:0}.}
  \begin{tabular}{l|cc|c|lllll}
    \toprule
    \midrule
    \multirow{2}{*}{\textbf{Method}} & \multicolumn{2}{c|}{\textbf{Lexical}} & \multirow{2}{*}{\textbf{Semantic}} & \multicolumn{5}{c}{\textbf{Cognitive}}\\
    &\textbf{ROUGE-1/2/L} & \textbf{BertScore} &  & \textbf{\;\;P.} & \textbf{\;\;S.} & \textbf{\;\;L.} & \textbf{\;\;K.} & \textbf{\;\;T.}\\
    \bottomrule

    \midrule
    \textsc{LLaMA-3.1} & $22.12/3.91/12.02$ &	$81.39$	& $3.22$ & $3.03$ & $3.15$ & $3.00$ & $3.05$ & $3.20$\\
    \textsc{- \small{Self}} & $24.10/4.45/13.10$ & $81.89$ & $3.13$ & $3.02$ & $3.23$ & $2.98$ & $3.06$ & $3.32$\\
    \midrule
    \rowcolor{cyan!40!green!10!white}\textsc{- \small{PersonalityTrait}} & $21.74/3.82/11.77$ & $81.54$ & $3.21$ & $3.10\!\,\color{green!50!black}{\uparrow}$ & $3.35\!\,\color{green!50!black}{\uparrow}$ & $3.07\!\,\color{green!50!black}{\uparrow}$ & $3.16\!\,\color{green!50!black}{\uparrow}$ & $3.36\!\,\color{green!50!black}{\uparrow}$\\
    \rowcolor{cyan!40!green!10!white}\textsc{- \small{CognitiveGuide}} & $19.36/3.48/10.45$ & $81.64$ & $3.24$ & $3.23\!\,\color{green!50!black}{\uparrow}$ & $3.47\!\,\color{green!50!black}{\uparrow}$ & $3.24\!\,\color{green!50!black}{\uparrow}$ & $3.21\!\,\color{green!50!black}{\uparrow}$ & $3.50\!\,\color{green!50!black}{\uparrow}$ \\
    \midrule
    \rowcolor{red!40!orange!10!white}\textsc{- \small{TrainIndividual}} & $22.49/3.94/12.19$ & $81.42$ & $3.21$ & $3.05$ & $3.31\!\,\color{green!50!black}{\uparrow}$ & $3.00$ & $3.15\!\,\color{green!50!black}{\uparrow}$ & $3.28\!\,\color{green!50!black}{\uparrow}$\\
    \rowcolor{red!40!orange!10!white}\textsc{- \small{TrainUnified}} & $28.04/4.34/14.09$ & $83.35$ & $3.03$	& $2.81\!\,\color{red!50!black}{\downarrow}$ & $3.00\!\,\color{red!50!black}{\downarrow}$ & $2.72\!\,\color{red!50!black}{\downarrow}$ & $2.86\!\,\color{red!50!black}{\downarrow}$ & $3.50\!\,\color{green!50!black}{\uparrow}$\\
    \midrule
    \midrule
    \textsc{Qwen-3} & $23.08/3.25/11.00$ & $81.21$ & $3.41$ & $3.51$ & $3.66$ & $3.44$ & $3.36$ & $3.62$\\
    \textsc{- \small{Self}} & $22.14/3.18/10.73$ & $81.44$ & $3.33$ & $3.51$ & $3.72$ & $3.45$ & $3.27$ & $3.83$ \\
    \midrule
    \rowcolor{cyan!40!green!10!white}\textsc{- \small{PersonalityTrait}} & $21.60/3.03/10.34$ & $81.36$ & $3.42$ & $3.58\!\,\color{green!50!black}{\uparrow}$ & $3.81\!\,\color{green!50!black}{\uparrow}$ & $3.66\!\,\color{green!50!black}{\uparrow}$ & $3.36$ & $3.77$ \\
    \rowcolor{cyan!40!green!10!white}\textsc{- \small{CognitiveGuide}} &  $20.86/2.98/10.03$ & $81.25$ & $3.64$ & $4.01\!\,\color{green!50!black}{\upuparrows}$ & $4.18\!\,\color{green!50!black}{\upuparrows}$ & $3.97\!\,\color{green!50!black}{\upuparrows}$ & $3.50\!\,\color{green!50!black}{\uparrow}$ & $4.21\!\,\color{green!50!black}{\upuparrows}$\\
    \midrule
    \rowcolor{red!40!orange!10!white}\textsc{- \small{TrainIndividual}} & $23.40/3.23/11.17$	& $81.18$ & $3.43$ & $3.50$ & $3.71$ & $3.49\!\,\color{green!50!black}{\uparrow}$ & $3.37$ & $3.69$\\
    \rowcolor{red!40!orange!10!white}\textsc{- \small{TrainUnified}} &  $27.16/4.12/13.78$ & $83.48$ & $2.99$ & $2.71\!\,\color{red!50!black}{\downdownarrows}$ & $2.89\!\,\color{red!50!black}{\downdownarrows}$ & $2.59\!\,\color{red!50!black}{\downdownarrows}$ & $2.82\!\,\color{red!50!black}{\downdownarrows}$ & $3.34\!\,\color{red!50!black}{\downarrow}$\\
    \midrule
    \bottomrule
  \end{tabular}
  \label{tab:1}
  \vspace{-5pt}
\end{table*}

%0.214036642	0.031922782	0.101014528	0.813930373	3.660972405	
%4.081	4.237	4.099	3.601	4.21

\begin{template2}{How Far Can Current Techniques Enhance the Cognitive Simulation Capabilities of LLMs?}
\texttt{Question 3:}\vspace{1pt}\\
\hspace*{5pt}\textit{How Effective Are Prompting Techniques?}
\tcblower
\texttt{Answer:} \textcolor{red!60!green}{To some extent but with limitations}.\vspace{1pt}\\
\textit{Prompting techniques guided by cognitive theories can provide some gains, but their effectiveness is highly constrained by the base cognitive capacities.}
\end{template2}

%Current prompting techniques yield marginal benefits for cognitive simulation. 
%For example, both PersonalityTrait and CognitiveGuide prompting tend to trade off a small amount of semantic alignment for gains on selected cognitive dimensions (e.g., P., S., or L.), suggesting that theory-informed prompts can steer models toward cognitively relevant signals rather than purely surface-level semantics.
%Guided by cognitive theories and psychological priors, \textsc{PersonalityTrait} and \textsc{CognitiveGuide}, LLMs exhibit some improvements on cognitive facets, often at the cost of slight lexical reduction. 
%Compared to the larger improvement achieved by \textsc{SelfAkin} and \textsc{Leaky}, prompting techniques can steer LLMs to reason from a cognitive perspective, but remain fundamentally constrained by the LLMs' limited  cognitive capacities.

Guided by cognitive theories and psychological priors, \textsc{PersonalityTrait} and \textsc{CognitiveGuide} yield moderate but consistent improvements on several cognitive facets, while leading to slight degradations in lexical metrics.
Compared to the larger gains achieved by \textsc{SelfAkin} and the oracle \textsc{Leaky} setting, prompting-based techniques can indeed steer LLMs toward cognitively aligned reasoning patterns, but their effectiveness remains fundamentally bounded by the models’ intrinsic cognitive capacities, as evidenced by the significant increases observed on \textsc{Qwen-3} than on \textsc{LLaMA-3.1} under \textsc{CognitiveGuide}.
In particular, prompting appears to reshape the expression of reasoning, rather than reliably inducing deeper cognitive abstraction.

\begin{template2}{How Far Can Current Techniques Enhance the Cognitive Simulation Capabilities of LLMs?}
\texttt{Question 4:}\vspace{1pt}\\
\hspace*{5pt}\textit{How Effective Are Learning-based Methods?}
\tcblower
\texttt{Answer:} \textcolor{red2!60!red}{To a minor extent}.\vspace{1pt}\\
\textit{%Learning-based methods mainly reinforce statistical fitting and fail to induce higher-level cognitive abstraction.
Learning-based methods focus on statistical fitting rather than higher-level cognitive abstraction.}
%\textit{Despite additional training, learning-based methods tend to reinforce behavioral imitation and fail to induce higher-level cognitive abstraction.}
\end{template2}

% Learning-based approaches, including fine-tuning under SELFAKIN and related settings, produce larger absolute gains than prompting. Notably, SELFAKIN substantially outperforms SELF, demonstrating that models can strongly benefit from in-distribution cognition leakage.

% However, this improvement comes with an important caveat. The SELFAKIN–SELF gap is consistently large, indicating that models are heavily relying on behavioral similarity and distributional overlap, rather than abstracting higher-level cognitive principles. Even after fine-tuning, the performance gap to the LEAKY upper bound remains significant.

% These results suggest that current learning-based methods still operate primarily in a behavior-capturing regime, rather than enabling true cognitive internalization and transfer.

% Looking forward, this limitation points to promising future directions. Methods such as cognitive meta-learning, cognition-guided reinforcement learning, or training paradigms explicitly designed to model and generalize latent cognitive processes may be necessary to move beyond behavior imitation toward genuine cognitive understanding.

Existing supervised learning methods are largely ineffective at cognitive internalization. For \textsc{TrainIndividual}, LLMs tend to capture surface-level statistical regularities as shortcuts rather than abstract cognitive patterns that generalize across domains. Performance gains are predominantly observed at the lexical level, whereas cognitive dimensions demonstrate negligible improvement or even degradation.
%LLMs cannot automatically learn to simulate cognitive processes across domains, exhibiting minimal improvements or even slight degradations at the cognition level ($-0.15$ to $0.06$), while effects are concentrated at the lexical level.
%In particular, improvements after training are minimal at the cognitive level (Level 3), with most cognitive facets remaining nearly unchanged or even slightly degraded compared to the \textsc{Self} baseline. 
%Instead, the observed gains are concentrated at the lexical level (Level 1), suggesting that supervised learning primarily captures surface-level statistical regularities rather than higher-order cognitive mechanisms. 
%This indicates that existing supervision signals do not provide sufficient pressure for models to abstract transferable cognitive processes from heterogeneous, cross-domain data.
The situation becomes even more severe for \textsc{TrainUnified}, 
%where profiles extracted from each scholar are used as identifiers to train shared parameters. 
where LLMs achieve substantial improvements in lexical metrics ($1.7$ to $3.4$ in ROUGE), which is in line with common observations in downstream task training, but suffer a systematic collapse in cognitive-level performance.
%Across both LLaMA-3.1 and Qwen-3, cognitive scores under \textsc{TrainUnified} decrease markedly across nearly all facets, despite strong gains in ROUGE and BERTScore. 
%This suggests that averaging individual cognitive traits in massive training will erase their specific patterns.
This suggests that averaging individual cognitive traits in massive training will erase their specific patterns, which is inherent to current large-scale pretraining paradigms.

%, reducing cognitive simulation to statistical fitting.

%Taken together, these results indicate that current learning-based approaches tend to reinforce behavior-level regularities while undermining cognitive fidelity. Whether training individual models or a unified model, existing supervision paradigms fail to preserve—let alone internalize—human-specific cognitive processes. This highlights a fundamental limitation of current learning methods and suggests that advancing cognitive simulation may require new paradigms, such as cognition-aware supervision, cognitive meta-learning, or reinforcement learning objectives explicitly aligned with latent reasoning processes.

\section{Related Work}
Role-playing agents \cite{chen2024from} have emerged as a promising direction for creating personalized conversational AI systems.
Early studies focus on constructing role-specific dialogue datasets and establishing foundational frameworks for character simulation \cite{shao-etal-2023-character,wang-etal-2024-rolellm,yu-etal-2024-neeko,zhou-etal-2024-characterglm}, ranging from trainable character models to domain-specific implementations such as anime characters \cite{li2023chatharuhi} and social support agents \cite{tu2023characterchat}.
Recent advances turn to sophisticated personality modeling and dynamic character optimization \cite{gao-etal-2025-tailorrpa,ye-etal-2025-cpo,wang2025coser,yin-etal-2025-charactercraft}, including explicitly modeling the Big Five personality dimensions through dynamic profile optimization \cite{jiang-etal-2024-personallm,dai2025profile}.

Existing human cognition benchmarks have explored various dimensions, including theory of mind capabilities \cite{chen-etal-2024-tombench,Tong2026CogToMAC} and emotional intelligence \cite{paech2023eq,sabour-etal-2024-emobench}.
Recent efforts attempt to benchmark human cognitive simulation capabilities by evaluating how well LLMs can simulate diverse human personas under predefined psychological characteristics \cite{xie2025human} or demographic profiles \cite{wang2026humanllm}. 
They rely on LLMs to synthesize human cognitive processes with manually defined personality traits. In contrast, our work collects and extracts authentic human cognitive processes from real human data, providing a more grounded foundation for evaluating LLMs' cognitive modeling capabilities.

\section{Conclusion}
Despite strong performance in behavioral alignment, current state-of-the-art LLMs remain limited in achieving cognitive internalization, with existing prompting techniques and learning-based methods largely reinforcing statistical fitting rather than inducing higher-level cognitive abstraction.
%Our experiments systematically distinguish behavioral simulation from cognitive internalization and show that contemporary LLMs primarily succeed at the former. While models can reproduce observable, individual-specific decisions and actions, both automatic metrics and human evaluation reveal that genuine abstraction of latent cognitive processes remains rare. Theory-guided prompting yields only marginal cognitive benefits, often accompanied by trade-offs in semantic or lexical consistency, indicating that inference-time guidance reshapes surface behavior without altering underlying representations. Learning-based methods exhibit even stronger limitations: individual-level training concentrates gains at the lexical level, while unified training amplifies statistical regularities at the cost of collapsing cognitive-level alignment. Together, these results suggest that current enhancement paradigms favor distributional fitting over the preservation or abstraction of human-specific cognitive structures.
More broadly, our findings point to a paradigm-level limitation in current LLM development. Modern LLMs rely heavily on the passive emergence of human-like capabilities from large-scale training, an approach that appears effective for behavioral imitation but insufficient for robust cognitive internalization. Advancing toward genuine cognitive modeling may require shifting from passive data exposure to high-level active learning objectives, such as cognitive meta-learning, abstraction-oriented self-supervision, or reinforcement learning frameworks explicitly aligned with human reasoning processes rather than just outcome rewards. We hope this work clarifies the current boundary of LLM cognitive capabilities and motivates future research toward principled mechanisms for learning and transferring human cognitions. More discussions are in Appendix \ref{appsec:discuss}.

%\section*{Acknowledgements}

\section*{Impact Statement}
This paper introduces a new dataset and benchmark for analyzing the extent to which LLMs can simulate aspects of human behavior and cognition. Our goal is to provide a systematic and controlled evaluation framework that helps the research community better understand the capabilities and limitations of current models, rather than to deploy or encourage specific real-world applications.
We anticipate that this work will primarily contribute to methodological advances in machine learning evaluation, particularly in the study of model generalization, robustness, and cognitive simulation. By offering finer-grained analysis tools and benchmark data, our work may support future research on building more reliable, interpretable, and safer language models.
We do not foresee negative societal impacts arising from this work. The dataset and benchmark are intended for research purposes only and do not introduce new model architectures or training procedures that could directly amplify harmful behaviors. As with any research on LLMs, there remains a general risk of misuse if models are deployed irresponsibly. However, our contribution focuses on evaluation and analysis rather than deployment. We believe that improving understanding of model behavior is an essential step toward mitigating such risks.

\bibliography{reference}
\bibliographystyle{icml2026}

%%%%%%%%%%%%%%%%%%%%%%%%%%%%%%%%%%%%%%%%%%%%%%%%%%%%%%%%%%%%%%%%%%%%%%%%%%%%%%%
%%%%%%%%%%%%%%%%%%%%%%%%%%%%%%%%%%%%%%%%%%%%%%%%%%%%%%%%%%%%%%%%%%%%%%%%%%%%%%%
% APPENDIX
%%%%%%%%%%%%%%%%%%%%%%%%%%%%%%%%%%%%%%%%%%%%%%%%%%%%%%%%%%%%%%%%%%%%%%%%%%%%%%%
%%%%%%%%%%%%%%%%%%%%%%%%%%%%%%%%%%%%%%%%%%%%%%%%%%%%%%%%%%%%%%%%%%%%%%%%%%%%%%%
\newpage
\appendix
%\onecolumn
%\section{You \emph{can} have an appendix here.}

% You can have as much text here as you want. The main body must be at most $8$
% pages long. For the final version, one more page can be added. If you want, you
% can use an appendix like this one.

% The $\mathtt{\backslash onecolumn}$ command above can be kept in place if you
% prefer a one-column appendix, or can be removed if you prefer a two-column
% appendix. Apart from this possible change, the style (font size, spacing,
% margins, page numbering, etc.) should be kept the same as the main body.
%%%%%%%%%%%%%%%%%%%%%%%%%%%%%%%%%%%%%%%%%%%%%%%%%%%%%%%%%%%%%%%%%%%%%%%%%%%%%%%
%%%%%%%%%%%%%%%%%%%%%%%%%%%%%%%%%%%%%%%%%%%%%%%%%%%%%%%%%%%%%%%%%%%%%%%%%%%%%%%
\newpage
\section{Discussion}
\label{appsec:discuss}
The experimental results point to a recurring tension in current LLM research: while LLMs increasingly exhibit human-like behaviors at the level of language and task performance, these behaviors do not necessarily reflect the acquisition of human-like cognitive structures. Rather than attributing this gap to insufficient scaling or missing data, we argue that it arises from deeper limitations in current modeling paradigms, specifically, how learning objectives, training data, and architectural biases interact to shape which capabilities can emerge.
In this section, we attempt to discuss why cognitive internalization remains elusive under prevailing approaches, which aspects of human cognition are systematically underrepresented, and which modeling principles may be required to bridge this gap. We first articulate hypotheses on the structural limitations of current paradigms, then reflect on the role and limits of cognitive task design, and finally outline learning architectures that may better support cognitive modeling.

\subsection{Hypotheses on the Limitations of Current Paradigms}

\paragraph{Statistical Pattern Matching Over Cognitive Modeling} The current paradigm of LLM development relies fundamentally on the passive emergence of capabilities from exposure to massive data. While this approach has proven remarkably successful at producing models that can generate fluent, contextually appropriate text, our findings suggest it is inherently limited in fostering genuine cognitive internalization. The core issue may lie in the nature of learning, especially the self-supervised learning, where current LLMs primarily develop statistical associations and surface-level pattern matching capabilities rather than the structured, hierarchical cognitive models that characterize human reasoning \cite{lake2017building,marcus2019rebooting}. When confronted with tasks requiring human-like cognitive simulation, such as perspective-taking, counterfactual reasoning, or theory of mind, LLMs rely on probabilistic heuristics derived from the training data distribution by default, rather than engaging in the kind of structured mental simulation that humans employ \cite{sap-etal-2022-neural,shapira-etal-2024-clever}.

This limitation manifests most clearly in the models' reliance on contextual association. Even state-of-the-art models, demonstrating impressive capabilities primarily within the distributional bounds of their training data \cite{bubeck2023sparks}, struggle with systematic generalization, particularly in cognitive tasks that require going beyond surface patterns to engage with underlying causal or intentional structures \cite{berglund2023reversal}. For instance, while these models can often produce plausible responses about human emotions or social dynamics when similar scenarios exist in training data, they frequently fail when required to reason about novel configurations of social relationships or to maintain consistent psychological models across extended interactions \cite{kosinski2024evaluating}. This suggests that what appears as human-like understanding may often be sophisticated pattern completion rather than genuine cognitive simulation.

\paragraph{The Averaging-Out of Cognitive Signatures} A second fundamental limitation stems from the nature of the training objective itself. LLMs are trained to predict the next token by learning from an aggregation of human-generated text, essentially optimizing to model the average or most probable continuation across diverse human authors, contexts, and cognitive styles \cite{bender2021dangers}. While this objective enables impressive linguistic fluency and reasoning capabilities \cite{guo2025deepseek}, it may actively work against the development of distinctive cognitive signatures. Human cognition is characterized not by statistical regularities but by individual differences, systematic biases, and context-specific reasoning strategies \cite{sternberg1997thinking}, precisely the irregularities that are smoothed out on large-scale data.

%\paragraph{Limits of Cognitive Modeling Objectives}
\paragraph{Limited Coverage of Cognitive Phenomena in Training}
This limitation is not merely a consequence of data scarcity but more fundamentally stems from the objectives imposed during training and alignment. Current alignment techniques, most notably reinforcement learning from human feedback \cite{ouyang2022training}, primarily optimize models to produce outputs that are judged favorably by human evaluators. Such judgments are largely based on surface-level behavioral criteria, including coherence, relevance, politeness, and safety. Crucially, these objectives do not explicitly incentivize the model to acquire or represent the underlying cognitive processes that humans use to generate similar responses.
As a result, models are rewarded for producing behaviorally appropriate outcomes rather than for constructing internal representations that reflect human-like mental models and causal reasoning mechanisms. Even when training data implicitly contains traces of such cognitive phenomena, the absence of explicit objectives guiding their acquisition means that models can achieve high alignment scores through shortcut strategies and pattern imitation. In this sense, current training paradigms favor behavioral alignment over cognitive internalization, enabling models to approximate human responses without reliably modeling the cognitive structures that underlie them.

\subsection{Future Directions Toward Cognitive Modeling}
Addressing these limitations will likely require fundamental shifts in how we approach LLM development, i.e., moving beyond passive scaling and behavioral optimization toward explicitly targeting the development of structured cognitive capabilities. We outline several possible directions for future research, organized into two complementary strategies: enriching training data and scenarios to better capture cognitive phenomena, and developing novel learning architectures that are better suited to meta-level capabilities, including cognitive abstraction and generalization.

\paragraph{Cognitive Data and Scenarios}
Recent task scenarios that explicitly target cognitive phenomena beyond surface-level modeling include perspective-taking and role-playing datasets that require maintaining consistent personas and belief states \cite{xie2025human}, theory-of-mind benchmarks focusing on false-belief reasoning and mental state attribution \cite{Tong2026CogToMAC}, and affective reasoning datasets that probe emotion recognition and empathetic response generation \cite{sabour-etal-2024-emobench}. While these datasets provide valuable probes and, in some cases, limited training signals, they typically operationalize cognition through isolated, task-specific objectives. Therefore, many cognitively salient scenarios remain underexplored, including the \textit{decomposition of individual-specific knowledge systems and cognitive patterns from shared world knowledge}, the \textit{evolution of individual cognition over extended temporal horizons}, and the \textit{assimilation or divergence of individual cognitive states within group settings}. At the same time, an exclusive focus on increasingly fine-grained or narrowly scoped cognitive datasets may be counterproductive. Human cognition does not arise from learning isolated abilities in disjoint scenarios, but instead exhibits strong transfer and abstraction across tasks and contexts. Consequently, while the development of well-designed cognitive modeling scenarios remains an important direction for future research, further progress is likely to depend more critically on advances in cognitive modeling objectives and representations, rather than on the proliferation of ever more specialized datasets.

\paragraph{Cognitive Learning Achitectures}
Beyond data-centric improvements, advancing cognitive modeling in LLMs likely requires learning paradigms that explicitly encourage abstraction, adaptation, and internal structure formation. One promising direction is to \textit{move from static behavior optimization toward learning frameworks that emphasize flexibility and transfer}, enabling models to adapt their reasoning strategies across cognitive contexts rather than relying on fixed patterns.
Relatedly, \textit{self-supervised objectives could be redesigned to prioritize abstract structure}, such as causal relations, compositional schemas, and persistent belief states, over surface-level token prediction, encouraging models to form representations that generalize across situations rather than memorizing correlations.
Another direction involves \textit{shifting training signals from purely outcome-based supervision to process-aware objectives}, where models are incentivized not only to reach correct conclusions but also to do so through cognitively plausible intermediate reasoning. 
Finally, \textit{incorporating explicit structural inductive biases}, such as symbolic state representations or hybrid neural–symbolic components, may help models maintain coherent internal world models and support counterfactual or long-horizon reasoning. Collectively, these directions suggest that meaningful progress in cognitive modeling is more likely to arise from rethinking learning objectives and architectural biases than from incremental extensions of existing training pipelines.
Looking further ahead, we anticipate that \textbf{truly human-level cognitive modeling may ultimately require brain-inspired architectures that go beyond the prevailing probabilistic modeling paradigm}, offering fundamentally new principles for representing, learning, and reasoning about intelligence.

\section{Dataset Analysis}
\label{appsec:stats}

\begin{table}[t]
  \centering
  \small % 稍微缩小字体以适应版面
  \setlength{\tabcolsep}{5pt} % 调整列间距
  \caption{Statistics of our benchmark.}
  \label{apptab:stats_summary_total}
  \begin{tabular}{l|cc|c}
    \toprule
    \textbf{Statistic} & \textbf{Expert} & \textbf{Automatic} & \textbf{Total} \\
    \midrule
    \# Scholars & 100 & 117 & 217 \\
    \# Total Papers & 1,875 & 1,547 & 3,422 \\
    Papers per Author & 18.8 & 13.2 & 15.8 \\
    Clusters per Author & 2.3 & 2.3 & 2.3 \\
    \midrule
    \textit{Tokens per Component} & & & \\
    \quad Introduction & 614.9 & 637.2 & 625.0 \\
    \quad Background & 178.9 & 99.0 & 142.8 \\
    \quad Motivation & 88.2 & 55.1 & 73.2 \\
    \quad Approach & 107.2 & 66.2 & 88.7 \\
    \quad Solution & 88.6 & 66.7 & 78.7 \\
    \bottomrule
  \end{tabular}
\end{table}

Expert annotation serves as our Gold Standard. The expert annotators capture the full depth of cognitive reasoning, resulting in significantly longer and richer components (e.g., Motivation averages 88.2 tokens in \Cref{apptab:stats_summary_total}).
In contrast, automatic annotation acts as a Silver Standard. While it validates the scalability of our pipeline, the LLM-based extraction tends to be more concise (Motivation averages 55.1 tokens) and exhibits a \textit{summarization bias}, capturing the gist rather than the nuanced cognitive steps found in expert annotation. To ensure the rigorousness and validity of our conclusions, \textbf{we conduct cognitive simulation experiments and human evaluations on the expert dataset}.

To rigorously assess the reliability of the automatic annotation, we conducted an alignment analysis using the overlapping data with the expert annotation.
Specifically, we identified \textbf{21 authors} present on both sides, covering \textbf{199 valid papers} after filtering (as shown in Table \ref{apptab:alignment_stats}). This overlap allows us to treat the expert annotations as the ground truth and evaluate the fidelity of the LLM-based extraction.
We measured the alignment at the token level. For each cognitive component (Background, Motivation, Approach, Solution), we computed the intersection between the tokens in the expert extraction ($T_{exp}$) and those in the automated extraction ($T_{auto}$). We report the Precision, Recall, and F1 Score based on these token overlaps.
Table \ref{apptab:f1_scores} presents the component-wise performance. We observe three key trends:

\begin{itemize}
    \item \textbf{High Precision Regime:} The automatic annotation demonstrates notably high precision across all components (e.g., \textbf{88.3\%} for Background, \textbf{76.0\%} for Approach). This indicates that while the model may occasionally omit peripheral details (lower recall), the content it \textit{does} extract is highly accurate and free from irrelevant noise. This \textit{high-precision} characteristic is crucial for constructing a clean evaluation dataset.
    
    \item \textbf{Cognitive Complexity Variance:} Performance varies by component type. Background achieves the highest F1 score (73.2\%), as it typically involves explicit factual statements. In contrast, Motivation yields the lowest F1 score (56.8\%), reflecting the inherent difficulty of identifying gap analyses and subjective arguments.
    
    \item \textbf{Robustness of Automation:} Given that token-level overlap is a strictly conservative metric (penalizing synonyms or paraphrasing), the substantial F1 scores confirm that the automatic annotation successfully captures the semantic core of the human cognitive process, justifying its use for large-scale data expansion.
\end{itemize}

% 表格 1：数据重叠情况
\begin{table}[t]
\centering
\caption{Statistics of the overlapping subset used for quality alignment check on automatic annotation.}
\label{apptab:alignment_stats}
\begin{tabular}{l|c}
\toprule
\textbf{Alignment Metric} & \textbf{Count} \\
\midrule
Overlapping Authors & 21 \\
Total Papers Compared & 224 \\
Valid Papers (Success) & 199 \\
Invalid Papers (Failed) & 25 \\
\bottomrule
\end{tabular}
\end{table}

% 表格 2：具体的 F1 分数（根据你的截图填入）
\begin{table}[t]
\centering
\caption{Token-level agreement between the expert annotation and the automated extraction.}
\label{apptab:f1_scores}
\begin{tabular}{l|ccc}
\toprule
\textbf{Component} & \textbf{Precision} & \textbf{Recall} & \textbf{F1 Score} \\
\midrule
Background & 0.8834 & 0.6749 & 0.7318 \\
Motivation & 0.6788 & 0.5412 & 0.5684 \\
Approach   & 0.7596 & 0.6085 & 0.6405 \\
Solution   & 0.7557 & 0.5661 & 0.6047 \\
\bottomrule
\end{tabular}
\end{table}

To ensure that the automatic topic clustering aligns with human cognitive categorization, we conducted a cluster coherence evaluation.
We randomly sampled \textbf{30 authors} from the automated pipeline and presented their generated topic clusters to a hybrid evaluation panel.

\begin{itemize}
    \item \textbf{Evaluation Panel:} It consists of one human expert and two SOTA LLMs (\texttt{DeepSeek-R1} and \texttt{GPT-5.1}). This setup allows us to assess alignment from both human intuition and strict logical perspectives.
    \item \textbf{Protocol:} Evaluators independently assess whether the generated clusters are \textit{Acceptable} (semantically distinct and internally consistent) or \textit{Unacceptable}.
    \item \textbf{Results:} As shown in Table \ref{apptab:cluster_agreement}, the panel achieves a high absolute agreement of \textbf{70\%} (21/30 authors passed unanimously). 
    In addition, we observed a Gwet's AC1 score of \textbf{0.69}, indicating \textit{Substantial Agreement}. 
    These metrics confirm that the automatic clustering mechanism produces taxonomies that are logically sound and aligned with human categorization standards.
\end{itemize}

% 聚类一致性表格
\begin{table}[t]
\centering
\caption{Inter-rater reliability for the automated topic clustering validation. Gwet's AC1 is reported as the primary metric due to its robustness against prevalence paradox in high-quality data.}
\label{apptab:cluster_agreement}
\begin{tabular}{l|c}
\toprule
\textbf{Agreement Metric} & \textbf{Value} \\
\midrule
Sample Size & 30 Authors \\
Raters & 1 Human + 2 LLMs \\
\midrule
Absolute Agreement (3/3) & 70.0\% (21/30) \\
Fleiss' Kappa ($\kappa$) & 0.44 (Moderate) \\
\textbf{Gwet's AC1} & \textbf{0.69 (Substantial)} \\
\bottomrule
\end{tabular}

\end{table}

% =================================================
% Appendix D: Implementation Details (Aligned with Paper Terminology)
% =================================================
\section{Implementation Details}
\label{sec:implementation_details}

To ensure the reproducibility of our results, we provide details of our experimental environment, hyperparameter settings, and inference protocols.

\subsection{Hardware and Environment}
% All experiments, including both training and inference, were conducted on a single \textbf{NVIDIA A100-PCIE GPU}. 
% We implemented our method using \texttt{PyTorch} and the HuggingFace \texttt{PEFT} library, utilizing \texttt{bfloat16} precision to optimize memory usage without compromising numerical stability.
We categorize experiments by model scale and accessibility:
\begin{itemize}
    \item \textbf{Local Fine-tuning and Inference:} For open-source models with parameter sizes under 10B (e.g., \textbf{LLAMA-3.1-8B}, \textbf{QWEN-3-8B}), experiments were conducted on a single \textbf{NVIDIA A100-PCIE GPU}. We utilized \texttt{PyTorch} and the HuggingFace \texttt{PEFT} library with \texttt{bfloat16} precision.
    \item \textbf{Large-Scale Model Inference:} For ultra-large-scale models (e.g., \textbf{QWEN-3-235B}) and proprietary models (e.g., \textbf{GPT-5.1}, \textbf{GEMINI-2.5}), we performed inference via their official APIs or hosted inference endpoints to ensure standard performance alignment.
\end{itemize}

\subsection{Training Configurations}
\label{subsec:training_config}

We fine-tune our models using Low-Rank Adaptation (LoRA). As defined in our experimental setup, we implement two primary training strategies: \textsc{TrainUnified} (corresponding to the mixed-author setting) and \textsc{TrainIndividual} (corresponding to the single-author baseline).
We perform grid search to identify the optimal hyperparameters. To ensure stable convergence, we generally adopt an effective batch size of 16. The only exception is the Qwen-Unified model, where a smaller batch size of 4 is sufficient due to its parameter-efficient \textit{Attention-Only} update strategy.

\begin{table}[t]
    \centering
    \small % 表格字号
    \caption{\textbf{Hyperparameter Settings.} Optimal configurations for each strategy. Note that the Qwen-Unified model utilizes an efficient \textit{Attention-Only} approach.}
    \label{tab:hyperparams_stacked}
    
    % --- Panel A: Qwen ---
    \begin{tabular}{lcc}
        \toprule
        \multicolumn{3}{c}{\textbf{Model A: Qwen3-8B}} \\
        \cmidrule(lr){1-3}
        \textbf{Hyperparameter} & \textbf{\textsc{Unified}} & \textbf{\textsc{Individual}} \\
        \midrule
        Rank ($r$) & 32 & 16 \\
        Alpha ($\alpha$) & 64 & 32 \\
        Learning Rate & 5e-5 & 2e-5 \\
        Target Modules & Attn Only$^\ddagger$ & All Linear \\
        Max Seq Length & 8,192 & 2,048 \\
        \bottomrule
    \end{tabular}
    
    \vspace{0.3cm} % 上下间距
    
    % --- Panel B: Llama ---
    \begin{tabular}{lcc}
        \toprule
        \multicolumn{3}{c}{\textbf{Model B: Llama-3.1-8B}} \\
        \cmidrule(lr){1-3}
        \textbf{Hyperparameter} & \textbf{\textsc{Unified}} & \textbf{\textsc{Individual}} \\
        \midrule
        Rank ($r$) & 192 & 256 \\
        Alpha ($\alpha$) & 384 & 32 \\
        Learning Rate & 7e-5 & 1e-5 \\
        Target Modules & All Linear & All Linear \\
        Max Seq Length & 8,192 & 2,048 \\
        \bottomrule
    \end{tabular}
    
    \vspace{0.1cm}
    \begin{flushleft}
        \footnotesize
        $^\ddagger$ Updates are restricted to query, key, value, and output projection layers (\texttt{q,k,v,o}).
    \end{flushleft}
\end{table}

\subsection{Inference Protocols}
\label{subsec:inference}

During the inference phase, we maintain the same hardware environment to ensure consistency. We employ the Nucleus sampling strategy to balance diversity and coherence in the generated academic text.

\begin{itemize}
    \item \textbf{Decoding Strategy:} We set the sampling temperature to $T=0.7$ and top-p to $0.9$ (\texttt{do\_sample=True}). This configuration encourages the model to generate more creative and stylistically varied solutions compared to greedy decoding.

    \item \textbf{Generation Limits:} To fully accommodate the detailed Approach and Solution sections, we set the maximum generation length (\texttt{max\_new\_tokens}) to \textbf{8,192} for the \textsc{TrainUnified} models. For the \textsc{TrainIndividual} baselines, a limit of \textbf{4,096} is sufficient.
\end{itemize}

\subsection{Experimental Variants and Ablations}
\label{subsec:variants}

To strictly verify the effectiveness of our proposed method, we design a systematic ablation study aligned with the definitions in \Cref{q1} and \Cref{tab:0} of the main text. As implemented in our execution protocols, we evaluate five distinct inference settings,
% (\texttt{MODES\_TO\_RUN}).
%问下师兄 train\_other好像在正文里没有
%\paragraph{Inference-Time Ablation Settings.}
where we investigate the impact of information availability on cognitive simulation by varying the reference context retrieved (Top-$k=2$). While the definitions align with the main text, we provide the detailed rationale for each setting below:

\begin{itemize}
    \item \textbf{\textsc{Base Model}:} 
    Represents the original output \textbf{without any cognition history conditioning}. 
    \textit{Rationale:} This serves as the lower-bound baseline to isolate the model's intrinsic \textit{default academic voice}. By comparing this with \textsc{Self}, we can quantify the specific marginal gain contributed by the retrieval-augmented profiling mechanism.
    
    \item \textbf{\textsc{Other}:} 
    Uses \textbf{cognition histories randomly sampled from other authors}, serving as an \textbf{interference control group}. 
    \textit{Rationale:} This serves as a negative control to verify specificity. If the model performs well here, it indicates the metric merely measures \textit{generic scientific plausibility}. A drop in this performance confirms that the model is successfully distinguishing between different scholars' unique cognitive signatures.
    
    \item \textbf{\textsc{Self} (Standard):} 
    Incorporates the target author's \textbf{cognition history from the training set}, representing our standard \textbf{cross-domain setting}. 
    \textit{Rationale:} This is the core experimental setting. By restricting access to historical papers (past work), we strictly test the model's ability to perform \textit{cognitive internalization}, i.e., extracting abstract thought patterns from history and generalizing them to an unseen research problem.
    
    \item \textbf{\textsc{SelfAkin}:} 
    Leverages cognition histories from the \textbf{same author's test set} (excluding the current target), examining \textbf{in-domain behavioral leakage}. 
    \textit{Rationale:} The retrieved papers share the same temporal period and specific sub-topic with the target. This ablation with \textsc{Self} allows us to decouple \textit{behavioral stylistic} from \textit{cognitive pattern}, measuring how much performance is gained simply by semantic similarity.
    
    \item \textbf{\textsc{Leaky}:} 
    Provides \textbf{oracle access} to the author's authentic \textbf{cognition trajectory} (specifically the ground truth Motivation with core idea), serving as an \textbf{upper bound}. 
    \textit{Rationale:} By removing the difficulty of exploring the solution, this setting isolates the \textit{execution capability}. It defines the theoretical ceiling of performance when the model perfectly understands the author's intent.
\end{itemize}

% \paragraph{Training Strategy Variants.}
% Regarding the supervised fine-tuning stage (\S 3.3), we implemented two contrastive strategies:
% \begin{itemize}
%     \item \textbf{\textsc{Train Individual} (Single):} Fine-tuning separate parameters for each scholar.
%     \item \textbf{\textsc{Train Unified} (Mix):} Jointly optimizing a unified model conditioned on structured scholar profiles.
% \end{itemize}
%\paragraph{Training Strategy Variants.}
Regarding learning-based methods in \Cref{q2}, we investigate both parameter-isolation and -sharing settings.

\begin{itemize}
    \item \textbf{\textsc{TrainIndividual} (Specialized Experts):} 
    We train a separate model parameter set for each scholar.
    \begin{itemize}
        \item \textit{Data Scope:} Each author's model is fine-tuned \textbf{exclusively} on the specific training corpus, ensuring no interference from other cognitive styles.
        \item \textit{Input Format:} The input only contains the raw task context (Instruction + Observation), as the awareness of self-identification is implicitly encoded in the model weights.
    \end{itemize}

    \item \textbf{\textsc{TrainUnified} (Profile-Conditioned Mix):} 
    We jointly optimize a single unified model on the aggregated corpus of all researchers with their profiles.
    %to learn generalized scientific reasoning.
    \begin{itemize}
        \item \textit{Mechanism:} To distinguish between different authors within shared parameters and prevent \textit{cognitive averaging}, we extract a profile for each author as the identification condition.
        \item \textit{Input Construction:} We prepend the \textbf{Unified Author Profile} (see \emph{Profile Synthesis Prompt} in Appendix \ref{subsec:prompt_data}) to every training sample, which forces the model to treat the author's persona as an explicit condition during learning.
    \end{itemize}
\end{itemize}

%\paragraph{Implementation of Enhancement Techniques.}
We also evaluate specific enhancement strategies in \Cref{q2}. Here, we detail their concrete implementations:

\begin{itemize}
%     \item \textbf{Research Interest Modeling (INTEREST):} To reinforce the author's domain expertise, we implement a two-stage injection strategy. 
%     \begin{itemize}
%         \item \textit{Extraction:} We aggregate the abstracts of the author's \textbf{latest 5 papers} from the training set and utilize GPT-5-nano to summarize them into \textbf{3 core research interests}.
%         \item \textit{Inference:} These interests are injected into the system prompt as global guidance, ensuring the generated solutions align with the author's long-term research agenda.
%     \end{itemize}
    
%     \item \textbf{Personality Profiling (BIG FIVE):} We condition the model on the scholar's psychological profile.
%     \begin{itemize}
%         \item \textit{Privacy-Preserving Extraction:} We analyze the Introduction sections of the author's \textbf{latest 2 papers}. To prevent the model from cheating by memorizing specific keywords, we enforce a \textbf{Critical Privacy Instruction}: the profiling model (GPT-5) is strictly forbidden from mentioning specific research domains or algorithms, focusing solely on stylistic markers (e.g., citation patterns, modality, confidence).
%         \item \textit{Injection:} The extracted Big Five scores (Openness, Conscientiousness, Extraversion, Agreeableness, Neuroticism) are injected into the prompt to guide the decision-making style (e.g., "High Conscientiousness implies a preference for rigorous validation").
%  < / end{itemize}
\item \textbf{\textsc{PersonalityTrait} (Personality Profiling):}
% We condition the model on the scholar's psychological profile.
Following recent works on simulating human behavior in LLMs \citep{Zhu2025CanLI}, we condition the model on the scholar's psychological profile.
    \begin{itemize}
        \item \textit{Data Source \& Extraction:} We retrieve the author's \textbf{2 representative papers strictly from the training set} (historical publications), which ensures that the profile reflects the author's most recent established persona without leaking information from the test set (in-domain recent work).
        
        \item \textit{Privacy-Preserving Analysis:} We analyze the Introduction sections of these selected papers using \textsc{GPT-5.1}. To prevent the model from cheating by memorizing specific keywords, we enforce a \textbf{Critical Privacy Instruction}: the profiling model is strictly prohibited from mentioning specific research domains or algorithms, and is required to focus solely on cognitive-level markers (e.g., problem framing style, motivational reasoning patterns, abstraction preferences, epistemic stance, and the structure of argumentative narratives).
        
        \item \textit{Prompt Instruction:} In addition, we extract Big Five scores (Openness, Conscientiousness, Extraversion, Agreeableness, Neuroticism) and inject them into the prompt to guide the decision-making style (e.g., High Conscientiousness implies a preference for rigorous validation).
    \end{itemize}

    \item \textbf{\textsc{CognitiveGuide} (Cognitive Guidance):} 
    % We implement a progressive \textbf{Three-Stage Thought-Process} that refines the initial draft generated by the \textsc{PersonalityTrait} module into a cognitively aligned solution.
    Building upon the draft generated by \textsc{PersonalityTrait}, we implement the \textbf{\textsc{CognitiveGuide}} pipeline, which adopts the \textbf{Test-Time-Matching (TTM)} framework \citep{Zhan2025TestTimeMatchingDP} as its underlying reasoning mechanism to further refine the solution.
    \begin{itemize}
        \item \textit{Stage 1: Intent Rewriting (Reasoning Planning).} Taking the personality-conditioned draft as input, the model simulates the author's cognitive process of recalling specific past experiences. It formulates 1-2 specific search queries (Intents) in the third person to address logical gaps or lack of technical detail in the initial draft.
        
        \item \textit{Stage 2: Knowledge Refinement (Fact Alignment).} Using the retrieved historical chunks, the model refines the draft to ensure factual consistency. We strictly enforce a \textbf{``No Lazy Appending''} constraint: the model is prohibited from merely appending phrases like ``Inspired by...'' to the end; instead, it must \textbf{reconstruct the core sections} (e.g., Approach and Solution) to naturally weave in the retrieved technical details and internalize the perspective from third-person to first-person.
        
        \item \textit{Stage 3: Style Transfer (Cognitive Realization).} The model rewrites the refined content into the final output using the author's 5-dimensional cognitive profile (including Premise, Strategy, Logic, Keyword, and Taboos), which ensures the solution is articulated through the author's unique cognitive lens rather than generic academic language.
    \end{itemize}

    % \item \textbf{Cognitive Guide Anonymized (Impl. of \textsc{CognitiveGuide}-Anon):} A critical ablation of the pipeline. In this setting, we apply \textbf{Profile Sanitization} using GPT-5 before Stage 3. All specific entities (e.g., specific paper titles, dataset names, institutional affiliations) in the cognitive profile are masked or replaced with abstract methodological descriptions. This variant verifies whether the model genuinely simulates the underlying \textit{cognitive logic} rather than relying on rote memorization of specific entities.
    
\end{itemize}

\subsection{Expert Annotation: Workflow, Protocols, and Compensation}

\textbf{Annotator Qualifications}
To ensure the high fidelity of the \textit{Expert Annotation}, we recruit a team of 10 senior Ph.D. students specializing in Artificial Intelligence. Their diverse expertise covers major sub-fields including Natural Language Processing, Computer Vision, and Machine Learning.

\textbf{Dataset Construction Workflow}
The construction process is governed by a rigorous manual-centric protocol, strictly adhering to the following standardization rules:

\begin{itemize}
    \item \textit{Broad-spectrum Data Collection:} Unlike the automatic annotation restricted to specific open-access APIs, annotators aggregate papers from a comprehensive range of academic repositories, including \textbf{Google Scholar, IEEE Xplore, ACM Digital Library, and DBLP}. This manual flexibility enables the retrieval of high-impact publications that are often protected by anti-crawling mechanisms. Papers are processed in \textbf{descending order of citation counts}  to ensure the dataset captures the scholar's most influential contributions.
    
    \item \textit{Strict Validity \& Filtration Protocols:} To guarantee information density, we enforce strict length constraints. A valid entry is required to contain an Abstract ($>50$ characters), an Introduction ($>100$ characters), and extractable cognitive elements ($>20$ characters each). Papers failing to meet these thresholds or missing key sections are discarded. Furthermore, a ``clean-text'' rule is enforced where annotators manually remove non-textual artifacts such as footnotes and figure captions, ensuring textual integrity.
    
    \item \textit{Cognitive Element Extraction:} Annotators extract four specific cognitive elements: \textit{Background, Motivation, Approach, and Solution}. A strictly objective protocol is mandated: content must be extracted directly from the source text without subjective modification. If a specific cognitive element is explicitly absent in the original paper, the field will be recorded as ``null'' rather than hallucinated contents.
    
    \item \textit{Topic-based Clustering Criteria:} To support the cross-domain generalization setting, papers are clustered based on semantic themes derived from Titles and Abstracts. We use a \textbf{Minimum Category Size} principle, requiring each valid cluster to contain at least three papers to ensure statistical significance for testing. Papers that are semantically isolated or can not form a coherent group are excluded to maintain cluster purity.
    
    \item \textit{Limited Auxiliary Tools \& Verification:} While manual extraction is the default, the use of LLMs is permitted strictly as an auxiliary tool for interpreting highly specialized interdisciplinary terminologies. In such cases, a mandatory ``Human-in-the-loop'' verification is required to validate the accuracy and logic flow of the extracted content against the original source.
\end{itemize}

\textbf{Compensation}
%The annotators were compensated based on a performance-based model, calculated according to the number of effective papers (meeting all validity constraints) and clustering tasks completed. The payment rates were set to be competitive, significantly exceeding the local minimum hourly wage standards for research assistants.
The annotators are compensated under a performance-based model, calculated based on the number of effective papers (meeting all validity constraints) and clustering tasks completed. In practice, the resulting compensation corresponds to approximately three times the local minimum hourly wage.

%要写具体的金额吗

% \section{Prompts}
% \label{sec:prompts}

% The boxes below present the full prompts used in our experiments.

% % 推荐使用 tcolorbox 来展示 Prompt，这样看起来更像对话框，很专业
% \begin{tcolorbox}[title=System Prompt for Task A, colback=gray!5, colframe=gray!50!black]
% You are a helpful assistant expert in []. Your goal is to help the user analyze...
% \end{tcolorbox}

% \begin{tcolorbox}[title=User Prompt Template, colback=gray!5, colframe=gray!50!black]
% Please read the following text and answer the question:

% Text: \{Input\_Text\}

% Question: \{Question\}
% \end{tcolorbox}
\section{Full Prompts}
\label{sec:prompts}

% We provide the complete prompt templates used in our experiments. To ensure reproducibility, we present the exact instructions for system initialization, context injection, and task execution.
We provide the complete prompt templates used in our experiments. To ensure reproducibility, we categorize them into profile construction (data pre-processing), inference execution, and automated evaluation.

\subsection{Profile Construction Prompts}
\label{subsec:prompt_data}

This section details the prompts used to extract structured attributes (e.g., style profiles, research interests, personality traits) from raw papers. These attributes serve as conditioning inputs for the \textsc{TrainUnified} model and the \textsc{CognitiveGuide} enhancement strategies.

\paragraph{Unified Author Profile Extraction}
For the \textsc{TrainUnified} setting (\Cref{q2}), we condition the model on a comprehensive instruction derived from the author's writing history. This is a two-step process.

\vspace{0.15cm}
\noindent \textbf{Single-Paper Analysis} We first extract the stylistic fingerprint from individual papers.

\begin{promptbox}{Stylometric Analysis Prompt}
You are an expert in academic stylometry. Analyze the provided excerpts from an author's research papers to identify their unique writing and research style.

\textbf{Input Text:}
\{Paper Content\}

\textbf{Task:}
Extract the author's stylistic fingerprint across 5 specific dimensions.

\textbf{STRICT CONSTRAINT:}
Do NOT output generic descriptions. All analysis must be grounded in the provided text.

\textbf{Dimensions to Analyze:}
\begin{itemize}[leftmargin=0.5cm,itemsep=0pt, topsep=0pt]
    \item \textbf{Premise \& Worldview:} Identify their fundamental assumptions about the field.
    \item \textbf{Strategy:} How do they typically frame problems? (e.g., contrasting with baselines, proposing novel architectures).
    \item \textbf{Logic:} How do they connect ideas? (e.g., deductive, empirical, mathematical proofs).
    \item \textbf{Keywords:} List characteristic terminology or phrasing habits.
    \item \textbf{Taboos:} Identify any methods or assumptions they explicitly avoid or critique.
\end{itemize}
\end{promptbox}
% \begin{promptbox}{Step 1: Stylometric Analysis Prompt}
% You are an expert in academic stylometry. Analyze the provided excerpts from an author's research papers to identify their unique writing and research style.

% \vspace{0.1cm}
% \textbf{INPUT TEXT:}
% % \noindent\hrulefill \\
% \texttt{\{Paper Content\}} \\
% % \noindent\hrulefill

% \vspace{0.1cm}
% \textbf{CORE TASK:}
% Extract the author's stylistic fingerprint across the following 5 dimensions.

% \textbf{DIMENSIONS TO ANALYZE:}
% \begin{itemize}[leftmargin=0.5cm, itemsep=2pt]
%     \item \textbf{1. Premise \& Worldview:} Identify their fundamental assumptions about the field (e.g., "Data is all you need" vs. "Structure matters").
%     \item \textbf{2. Strategy:} How do they typically frame problems? (e.g., contrasting with weak baselines vs. proposing novel architectures).
%     \item \textbf{3. Logic:} How do they connect ideas? (e.g., deductive reasoning, empirical observation, or mathematical proofs).
%     \item \textbf{4. Keywords:} List characteristic terminology, specific jargon, or phrasing habits.
%     \item \textbf{5. Taboos:} Identify any methods, assumptions, or prior works they explicitly avoid or critique.
% \end{itemize}

% \vspace{0.1cm}
% \textbf{STRICT CONSTRAINT:}
% Do NOT output generic descriptions (e.g., "The author writes clearly"). All analysis must be **grounded in the provided text** with specific references.
% \end{promptbox}

\vspace{0.15cm}
\noindent \textbf{Profile Synthesis} We then aggregate multiple analyses into a single, coherent system instruction.

\begin{promptbox}{Profile Synthesis Prompt}
Below are multiple stylistic analyses of the same researcher derived from different papers:

\{Aggregated Analyses from Stylometric Analysis\}

\textbf{Task:}
Your task is to consolidate these observations into a single, coherent **System Instruction** (Author Profile). This profile will be used to instruct an LLM to write exactly like this author.

\textbf{Requirements:}
\begin{enumerate}[leftmargin=0.5cm,itemsep=0pt, topsep=0pt]
    \item Start with: "You are a researcher who..."
    \item Explicitly mention their preferred **Logic** and **Strategy**.
    \item List 3-5 characteristic **Keywords** or phrasing habits.
    \item Ensure the tone matches the author's academic persona.
\end{enumerate}
\end{promptbox}

% \paragraph{Research Interest Extraction}
% % For the \textsc{Interest} enhancement, we extract the author's top research topics from their latest 5 papers.
% Drawing on the expert profiling approach \citep{Tang2024StepBackPD}, we implement the \textsc{Interest} enhancement by extracting the author's top research topics from their latest 5 papers to enforce thematic consistency.

% \begin{promptbox}{Research Interest Extraction Prompt}
% I will provide you with some research papers you've authored. Each paper consists of a title and abstract. Please summarize your top three research interests based on these papers in JSON format.

% \textbf{Input Papers:}
% \{Aggregated Abstracts\}

% \textbf{Instructions:}
% Please summarize your top three research interests in the following JSON format:

% \{
%     "interest1": "Description of your first main research interest",
%     "interest2": "Description of your second main research interest",
%     "interest3": "Description of your third main research interest"
% \}
% \end{promptbox}

\paragraph{Personality Profiling}
% For the \textsc{PersonalityTrait} enhancement, we infer Big Five traits with strict privacy constraints.
Following established Big Five personality modeling approaches for LLMs \citep{Zhu2025CanLI}, we implement the \textsc{PersonalityTrait} enhancement to infer stable psychological traits with strict privacy constraints.

\begin{promptbox}{Personality Profiling Prompt}
You are a psychologically insightful agent. Your task is to analyze text to infer the author's stable personality traits based on the Big Five model.

\textbf{CRITICAL PRIVACY INSTRUCTION:}
The input text is confidential. Your output (especially the reasoning) \textbf{MUST NOT} reveal specific research topics, field names (e.g., NLP, Biology), specific algorithms, or technical methodologies mentioned in the text. Focus purely on \textbf{stylistic markers} (e.g., citation patterns, sentence length, modality).

\textbf{Input Text:}
\{Paper Introduction\}

\textbf{Output Format:}
Provide a score for each trait (Conscientiousness, Agreeableness, Neuroticism, Openness, Extraversion) from 1.0 to 5.0.
\end{promptbox}
% \paragraph{Profile Sanitization (Anonymized Setting).}
% For the \textsc{CognitiveGuide}-Anon variant (\S 3.3), we employ a sanitization step to strip explicit domain identifiers from the profile while preserving the structural methodology. This ensures the model relies on abstract reasoning patterns rather than memorized keywords.

% \begin{promptbox}{Profile Sanitization Prompt}
% \textbf{System Instruction:}
% You are an expert Research Methodology Analyst.

% \textbf{User Instruction:}
% Your task is to "sanitize" the following Researcher Profile to prevent data leakage (Target Paper Spoiler).

% \vspace{0.1cm}
% \textbf{INPUT PROFILE:}
% % \noindent\hrulefill \\
% \texttt{\{profile\_str\}} \\
% % \noindent\hrulefill

% \textbf{INSTRUCTIONS:}
% \begin{enumerate}[leftmargin=0.5cm]
%     \item \textbf{Remove Specifics:} Replace specific task names, dataset names, or unique domain terms from the target paper with \textbf{abstract methodological categories} (e.g., replace "Transformer for Machine Translation" with "Sequence Modeling Architecture").
%     \item \textbf{Keep the 'How':} Strictly preserve the methodological preferences and cognitive habits.
%     \item \textbf{Output:} Return ONLY the sanitized JSON object.
% \end{enumerate}
% \end{promptbox}

\subsection{Inference Prompts (Execution)}
\label{subsec:prompt_inference}

This section details the prompts used during the inference phase, covering both the standard baseline settings and our proposed Cognitive Guide pipeline.

\paragraph{Standard Inference (Baselines / SFT)}
For baseline settings (e.g., \textsc{Base Model}, \textsc{Self}, \textsc{TrainUnified}), we use a standardized prompt structure.

% \begin{promptbox}{Standard System Prompt}
% You are a brilliant researcher trying to solve a new challenge. Your goal is to develop a research solution based on a new observation and potentially inspired by past work.

% ---
% \textbf{DEFINITIONS:}
% \begin{itemize}
%     \item \textbf{Approach:} Introduces the inspiration and conceptual framework of the research methods.
%     \item \textbf{Solution:} Provides a detailed description of the implementation and execution.
% \end{itemize}
% \end{promptbox}

% \vspace{0.2cm}
% \begin{promptbox}{Standard User Prompt (with RAG)}
% ---
% \textbf{PAST WORK (Inspiration):}
% Title: \{Paper Title 1\}
% Background: \{Paper Background 1\}
% ... (Repeated for Top-k papers)

% ---
% \textbf{NEW CHALLENGE AND OBSERVATION:}
% \{Target Observation\}
% ---

% \textbf{TASK:}
% Based on the NEW CHALLENGE AND OBSERVATION, and using the PAST WORK as inspiration (or not, if 'No history' is provided), formulate a detailed Solution.
% \textbf{Constraint:} The response MUST ONLY contain the full text of the Approach and Solution.
% \end{promptbox}
\begin{promptbox}{Standard System Prompt}
You are a brilliant researcher trying to solve a new challenge. Your goal is to develop a research solution based on a new observation and potentially inspired by past work.

\vspace{0.1cm}
\textbf{Output Structure Definitions:}
\begin{itemize}[leftmargin=0.5cm,itemsep=0pt, topsep=0pt]
    \item \textbf{Approach:} Introduces the inspiration and conceptual framework of the research methods.
    \item \textbf{Solution:} Provides a detailed description of the implementation and execution.
\end{itemize}
\end{promptbox}

\vspace{0.2cm}
\begin{promptbox}{Standard User Prompt (with RAG Context)}
\textbf{PART 1: PAST WORK (Inspiration)}
The following are related works retrieved from history:

    \quad  \textbf{Paper 1:} \texttt{\{Paper Title 1\}}
    
    \qquad \textit{Background:} \texttt{\{Paper Background 1\}}
    
    \quad ... \textit{(Repeated for Top-k papers)}

\noindent\hrulefill

\textbf{PART 2: NEW CHALLENGE}
\texttt{\{Target Observation\}}

\noindent\hrulefill

\textbf{TASK INSTRUCTION:}
Based on the \textbf{NEW CHALLENGE}, and using the \textbf{PAST WORK} as inspiration (if available), formulate a detailed Solution.

\textbf{STRICT CONSTRAINT:}
The response \textbf{MUST ONLY} contain the full text of the \textbf{Approach} and \textbf{Solution}. Do not output any conversational fillers.
\end{promptbox}

\paragraph{Cognitive Guide Pipeline (Enhancement)}
% The Cognitive Guide executes a three-stage thought process to refine the generation.
% Drawing inspiration from the multi-stage reasoning frameworks proposed by \citet{Zhan2025TestTimeMatchingDP}, our Cognitive Guide executes a progressive three-stage thought process to refine the generation.
Based on the initial draft generated by the \textsc{PersonalityTrait} module, we implement a progressive \textbf{Test-Time-Matching} to further refine the solution. Adapted from multi-stage reasoning frameworks \citep{Zhan2025TestTimeMatchingDP}, it transforms the personality-conditioned draft into a cognitively aligned output through intent rewriting, knowledge refinement, and style transfer.

\noindent \textbf{Intent Rewriting} The model formulates search queries based on the initial draft.

% \begin{promptbox}{Stage 1: Intent Rewriting Prompt}
% The following is an interaction where a user asks for a research proposal, and you provide an initial draft:

% \textbf{User Query:} \{Observation\}
% \textbf{Initial Draft:} \{Draft generated by Personality Model\}

% You need to retrieve specific details from your (the author's) **past publications** to ground this proposal in your established research identity.

% \textbf{Task:}
% Describe what information you would need to know to better answer the user's question.
% \begin{itemize}
%     \item Use **third-person perspective**.
%     \item Ask about specific algorithms, parameter choices, or previous findings.
%     \item Limit to the most important 1-2 search queries.
% \end{itemize}
% \end{promptbox}
\begin{promptbox}{Intent Rewriting Prompt}
Given a scholarly interaction and an initial output:

\textbf{User Query:} \{query\} \\
\textbf{Initial Draft:} \{target\}

\textbf{Objective:} 
Identify critical gaps between the \textit{Initial Draft} and the author's established research identity. Formulate 1--2 targeted search queries to retrieve grounding evidence from \textbf{historical publications}.

\textbf{Constraints:}
\begin{itemize}[leftmargin=0.5cm,itemsep=0pt, topsep=0pt]
    \item \textbf{Identity-Centric:} Focus exclusively on the author's specific methodologies (e.g., algorithmic preferences, hyperparameters, or unique theoretical frameworks).
    \item \textbf{Third-Person Formulation:} Describe the required information using a neutral, third-person perspective (e.g., ``The author's prior approach to...'').
    \item \textbf{Exclusion Rule:} Do \textbf{NOT} seek information regarding the user; prioritize the author's internal research consistency.
    \item \textbf{Precision:} Provide only the core queries / statements in a concise declarative or interrogative format.
\end{itemize}
\end{promptbox}

\noindent \textbf{Knowledge Refinement} The model refines the draft using retrieved history, with strict constraints against hallucination and lazy appending.

% \begin{promptbox}{Stage 2: Knowledge Refinement Prompt}
% Below is a record of a conversation, including the input and your initial response:

% \textbf{User:} \{Observation\}
% \textbf{Response:} \{Initial Draft\}

% The following are memory fragments retrieved based on your identity:
% \{Retrieved RAG Chunks\}

% \textbf{Task:}
% Please revise your initial response based on the information in the memory.

% \textbf{Guidelines:}
% \begin{enumerate}
%     \item \textbf{MANDATORY TECHNICAL ALIGNMENT:} If the memory contains specific technical methodologies that differ from the generic methods in the draft, you MUST **REPLACE** the generic methods with the specific ones.
%     \item \textbf{NO LAZY APPENDING:} Do **NOT** simply add a paragraph at the end saying "Inspired by...". You must **rewrite the core sections** to structurally integrate the retrieved knowledge.
%     \item \textbf{CHARACTER CONSISTENCY:} Internalize the third-person memories as **"our past work"** or **"my experience"**.
% \end{enumerate}
% \end{promptbox}
\begin{promptbox}{Knowledge Refinement Prompt}
Below is a conversation record and relevant memory fragments:

\textbf{User:} \{query\} \\
\textbf{Initial Response:} \{target\} \\
\textbf{Refined Context:} \{rewritten\_query\} \\
\textbf{Memory Fragments:} \{chunks\}

\textbf{Task:}
Revise the \textit{Initial Response} by deeply integrating the \textit{Memory Fragments}.

\textbf{Guidelines:}
\begin{enumerate}[leftmargin=0.5cm,itemsep=0pt, topsep=0pt]
    \item \textbf{Technical Alignment:} Replace generic methodologies (e.g., "Deep Learning") with specific techniques from memory (e.g., "Kernel Methods", "Reranking"). Adjust the architecture to reflect any constraints or preferences found in the retrieved chunks.
    \item \textbf{Structural Integration:} \textbf{DO NOT} append updates at the end. You must rewrite the core sections (e.g., \textit{Approach}, \textit{Steps}) to ensure the new knowledge is woven into the logic flow.
    \item \textbf{First-Person Internalization:} Convert third-person memories into your own "past work" or "experience". \textbf{Conflict Resolution:} If the initial draft conflicts with memory, prioritize memory and \textbf{overwrite} the conflicting content.
    \item \textbf{Output Format:} Return \textbf{only} the revised response. No meta-talk, explanations, or "Here is the revised version".
\end{enumerate}
\end{promptbox}

\vspace{0.2cm}
\noindent \textbf{Style Transfer} The final rewrite injects the author's cognitive dimensions.

\begin{promptbox}{Style Transfer \& Cognitive Realization}
\textbf{Role:} You are now embodying the scholar \textbf{\{author\_name\}}. 

\textbf{Task:} Rewrite the \textit{Input Solution} to align with your unique 

\textbf{Research Persona} through the \textit{Thinking-Through-Mimicry (TTM)} process.

\textbf{Input Solution:} ``\{input\_solution\}''

\hrulefill

\textbf{Persona Profile (Strictly Follow):}
\begin{enumerate}[leftmargin=0.5cm,itemsep=0pt, topsep=0pt]
    \item \textbf{Epistemic Premise (Worldview):} Your research philosophy is grounded in: \{premise\}.
    
    \item \textbf{Methodological Preference (Strategy):} 
    \textit{Habitual approach:} \{strategy\}. 
    Beyond lexical substitution, if the input method contradicts your research habits (e.g., ``Deep Learning'' vs. your ``Probabilistic Models''), you must \textbf{reframe} the entire solution through your preferred lens.
    
    \item \textbf{Argumentation Architecture (Logic):} Your inherent logical structure follows: \{logic\}.
    
    \item \textbf{Lexical Signature (Keywords):} 
    \textit{Habitual usage:} \{keywords\}. 
    \textbf{Action:} Systematically replace generic terminology with these specific domain-specific keywords.
    
    \item \textbf{Conceptual Taboos (Avoidance):} 
    \textit{Strictly avoids:} \{taboos\}. 
    \textbf{Action:} Eradicate or heavily critique any concepts, assumptions, or baselines falling into this category.

\end{enumerate}

\textbf{Execution Guidelines:}
\begin{itemize}[leftmargin=0.5cm,itemsep=0pt, topsep=0pt]
    \item \textbf{Total Immersion:} Adopt an authoritative first-person voice (e.g., ``We contend...'', ``In our framework...'').
    \item \textbf{Cognitive Fidelity:} Do not merely mimic the author's writing style; ensure the \textit{underlying reasoning} reflects the author's intellectual heritage.
    \item \textbf{Output Specification:} Return \textbf{only} the rewritten solution (Approach + Solution).
\end{itemize}
\end{promptbox}

% \subsection{Evaluation Prompts (Automated Assessment)}
\subsection{Evaluation Prompts and Scoring Rubrics}
\label{subsec:prompt_eval}
To ensure rigorous evaluation, we design specific scoring rubrics for our hierarchical evaluation framework. The LLM judges are instructed to strictly follow these criteria.

% ---------------------------------------------------------
% Level 2: Semantic Similarity
% ---------------------------------------------------------
\paragraph{Semantic Similarity Judge}
We employ \texttt{GPT-5-nano} to assess the alignment between the simulated proposal and the actual ground truth.
To standardize the scoring, we explicitly included \textbf{Anchor Examples} (e.g., comparing with other models) in the prompt construction.

\begin{rubricbox}{Semantic Similarity Rubric}
% leftmargin=4.5cm: 左栏总宽
% labelwidth=4.3cm: 标签文字宽
% align=parcenter:  文字居中对齐 (需在导言区定义)
\begin{description}[style=multiline, leftmargin=1.9cm, labelwidth=2cm, font=\bfseries]

    \item[Score 5 \break {\footnotesize High Fidelity}] 
    The simulated proposal is \textbf{highly congruent} with the ground truth. The technical essence is identical.
    %\vspace{0.1cm} \\
    
    \textit{Anchor Example: BLIP $\leftrightarrow$ BLIP (Original)}
    \vspace{0.1cm} \hrule %\vspace{0.1cm}

    \item[Score 4 \break {\footnotesize Idea Aligned}] 
    The Research Problem and Core Idea are fully consistent, but the implementation shows structural modifications (e.g., a valid successor method).
    %\vspace{0.1cm} \\
    
    \textit{Anchor Example: BLIP $\leftrightarrow$ BLIP-2 (Successor Work)}
    \vspace{0.1cm} \hrule %\vspace{0.1cm}

    \item[Score 3 \break {\footnotesize Topic Aligned}] 
    The general Research Problem is correctly identified, but the Methodology \textbf{diverges significantly} (different era or architecture).
    %\vspace{0.1cm} \\
    
    \textit{Anchor Example: CLIP $\leftrightarrow$ ResNet (Image Classif. via Contrastive vs. CNN)}
    \vspace{0.1cm} \hrule %\vspace{0.1cm}

    \item[Score 2 \break {\footnotesize Domain Match}] 
    The proposal falls within the same broad research domain (e.g., Visual Generation) but addresses a different specific task.
    %\vspace{0.1cm} \\
    
    \textit{Anchor Example: CLIP $\leftrightarrow$ GAN (Both Vision, different tasks)}
    \vspace{0.1cm} \hrule %\vspace{0.1cm}

    \item[Score 1 \break {\footnotesize Unrelated}] 
    The proposal is \textbf{completely unrelated} to the ground truth in terms of problem definition and methodology.
    %\vspace{0.1cm} \\
    
    \textit{Anchor Example: CLIP $\leftrightarrow$ BERT (Vision vs. NLP)}

\end{description}
\end{rubricbox}

% [Box 2: Prompt]
\vspace{0.2cm}
\begin{promptbox}{Semantic Similarity Judge Prompt}
You are a senior academic researcher. Your task is to rigorously evaluate the relevance and similarity between a 'Simulated Research Proposal' and an 'Actual Paper's Research Proposal.'

\textbf{Assessment Dimensions:}
\begin{itemize}[leftmargin=0.5cm,itemsep=0pt, topsep=0pt]
    \item \textbf{Research Problem:} Is the motivation consistent?
    \item \textbf{Core Idea:} Is the technical philosophy aligned?
    \item \textbf{Logical Framework:} Is the argumentation structure similar?
\end{itemize}

\textbf{Scoring Standard:}
Please rate strictly according to the \textbf{5-point scale provided above}.

% \textbf{Constraint:} Output the result in JSON format including `score` and `justification`.
% \end{promptbox}
% \textbf{Output JSON Format Requirements:}
% Please strictly output the following JSON format:
% \begin{verbatim}
% {
%     "analysis_process": "Analyze similarities/differences across 
%                          Research Problem, Core Idea, and 
%                          Logical Framework.",
%     "final_score": (Integer 1-5),
%     "score_explanation": "One-sentence summary of the core reason 
%                           for this score level."
% }
% \end{verbatim}
\begin{itemize}[leftmargin=0.5cm, label={}, itemsep=0pt, topsep=0pt]
    \item \texttt{\{}
    \item \hspace{0.5cm} \texttt{"analysis\_process"}: "Analyze similarities and differences across Research Problem, Core Idea, and Logical Framework.",
    \item \hspace{0.5cm} \texttt{"final\_score"}: (Integer 1-5),
    \item \hspace{0.5cm} \texttt{"score\_explanation"}: "One-sentence summary of the core reason for this score level."
    \item \texttt{\}}
\end{itemize}
\end{promptbox}

This high-level evaluation involves a two-step pipeline: first construct a \textit{Ground Truth Thought Profile} from the author's authentic history, and then score the generated solution. %against this profile.

\noindent \textbf{Ground Truth Profile Construction}
% Before scoring, we extract the author's authentic cognitive patterns from their training set. The model is instructed to extract the "Scholar Thought Profile" across five distinct dimensions.
Before scoring, we extract the author's authentic cognitive patterns from their training set.
\textbf{Note:} This process is executed \textbf{separately for each dimension}. We instantiate the LLM five times, each time using the specific system instruction and extraction schema corresponding to one of the five dimensions below to ensure focused analysis.

\begin{promptbox}{Ground Truth Extraction (Inheritance Analysis)}
\textbf{System Role:}
You are a top-tier expert in the \textbf{inheritance analysis of scientific thinking}. Your task is to construct a \textbf{high-fidelity} profile of scholarly thinking by analyzing how a scholar's current research is rooted in and inherited from their historical publications.

\textbf{Input Materials:}
\begin{itemize}[leftmargin=0.5cm,itemsep=0pt, topsep=0pt]
    \item \textbf{Target Paper:} The latest research (Title and Introduction).
    \item \textbf{Historical Paper List:} A collection of the author's prior works.
\end{itemize}

\textbf{Task Description:}
Take the \textit{Target Paper} as the central case and analyze how its thought structures are \textbf{inherited from} the provided \textit{Historical Papers}.

\textbf{Dimensions to Analyze:}
\begin{enumerate}[leftmargin=0.5cm,itemsep=0pt, topsep=0pt]
    \item \textbf{Foundational Premise:} Inherited problem formulations and theoretical lenses (e.g., probabilistic vs. deterministic).
    \item \textbf{Strategic Preference:} Core value orientations and modeling philosophies (e.g., preferring interpretability over black-box models).
    \item \textbf{High-Level Logic:} Argumentation structures and challenge-to-innovation patterns.
    \item \textbf{Keywords:} Domain-specific terminology and habitual professional vocabulary.
    \item \textbf{Taboos:} Technologies, assumptions, or baselines that the author historically avoids or critiques.
\end{enumerate}

\textbf{Chain-of-Thought (CoT) Instructions:}
\begin{itemize}[leftmargin=0.5cm,itemsep=0pt, topsep=0pt]
    \item \textit{Analyze Target}: Identify core problem definitions and critique structures in the Target Paper.
    \item \textit{Scan Memory}: Identify papers in the Historical List that share the exact same theoretical premises.
    \item \textit{Verify Inheritance}: Ensure the similarity is due to consistent author style rather than general field overlaps.
    \item \textit{Select Representatives}: Choose 5 distinct papers that best prove this inheritance.
\end{itemize}

\textbf{Output Format:}
Summarize the features into a JSON profile containing \texttt{summary\_of\_features} and \texttt{most\_representative\_papers} for each dimension.
\end{promptbox}

% \noindent \textbf{Profile-Based Scoring.}
% The judge evaluates the generated solution against the constructed Ground Truth Profile using the following rubric.
\noindent \textbf{Profile-Based Scoring} The evaluation phase implements a profile-conditioned judging mechanism to quantify the cognitive alignment. For each test case, the LLM judge is provided with the multi-dimensional \textit{Ground Truth Profile} and the \textit{Representative Papers} as the evidentiary benchmark. The judge is required to perform a contrastive analysis between the \textit{Current Solution} and the established \textit{Standard Answer}, ensuring that the assigned scores are substantiated by specific cognitive traits defined in the profile. This process moves beyond surface-level similarity, focusing on whether the generated output successfully internalizes the scholar's underlying research logic and strategic constraints.

\begin{table}[t]
\centering
\caption{\textbf{Detailed Definition of Cognitive Dimensions.} This matrix guides the evaluator to map specific inherited traits to the generated solution.}
\label{tab:dimension_definitions}
%\vspace{0.2em}
\small
%\renewcommand{\arraystretch}{1.4} % 增加行高，防止拥挤
%\begin{tabularx}{\linewidth}{p{3.2cm} X} % X列会自动换行并填充剩余空间
\begin{tabular}{p{1.8cm}|p{5.6cm}}
\toprule
\textbf{Dimension} & \textbf{Assessment Protocol} \\ 
\midrule

\textbf{Foundational Premise} & 
\textbf{Focus:} The theoretical lens and problem formulation style. \newline
\textbf{Target (Score 5):} The solution must accurately reconstruct how the author typically frames problems (e.g., framing optimization as probabilistic inference) and their foundational knowledge base. \\ 
%\hdashline % 如果你的环境不支持 hdashline，可以用 \midrule[0.1pt] 代替
\midrule

\textbf{Strategic Preference} & 
\textbf{Focus:} Value trade-offs and research paradigms. \newline
\textbf{Target (Score 5):} The solution must mirror the author's persistent choices, such as favoring \textit{interpretable frameworks} over black-box performance, or \textit{theoretical modeling} over engineering hacks. \\ 
%\hdashline
\midrule

\textbf{High-Level Logic} & 
\textbf{Focus:} Argumentation architecture and challenge-response patterns. \newline
\textbf{Target (Score 5):} The reasoning flow must follow the author's deductive style (e.g., ``Challenge A $\rightarrow$ Innovation B'') rather than generic logical templates. \\ 
%\hdashline
\midrule

\textbf{Keyword Style} & 
\textbf{Focus:} Domain-specific terminology and linguistic habits. \newline
\textbf{Target (Score 5):} The text naturally embeds the author's unique vocabulary (e.g., using "manifold" instead of "space") without mechanical forcing. \\ 
\midrule[0.8pt] % 加粗线区分正向和反向指标

\textbf{Taboo Avoidance} \newline \textit{[Inverse Metric]} & 
\textbf{Focus:} Forbidden methodologies, assumptions, or baselines. \newline
\textbf{Target (Score 5):} The solution \textbf{successfully avoids} or actively critiques approaches the author historically rejects (e.g., explicitly excluding brute-force data fitting). \\ 

\bottomrule
\end{tabular}
\end{table}

\begin{rubricbox}{Unified Cognitive Consistency Rubric}
\textit{Note: For the "Taboo Avoidance" dimension, "Reproduction" refers to the reproduction of the author's \textbf{avoidance behavior} (i.e., successful critique or exclusion of forbidden methods).}

\begin{description}[style=multiline, leftmargin=2cm, labelwidth=2cm, font=\bfseries, itemsep=0pt, topsep=0pt]

\item[Score 5 \break {\footnotesize Identity Alignment}] 
The solution \textbf{perfectly reproduces} the dimension-specific criteria defined in Table \ref{tab:dimension_definitions}.
\begin{itemize}[leftmargin=0.5cm,itemsep=0pt, topsep=0pt]
    \item \textit{Evidence:} Details match the author's historical habits with \textbf{no conflicts}. 
    \item \textit{Taboo:} The solution strictly adheres to all negative constraints.
\end{itemize}
%\vspace{0.1cm} 

\noindent\hrule

\item[Score 4 \break {\footnotesize High Consistency}] 
The solution \textbf{highly captures} the core cognitive structures.
\begin{itemize}[leftmargin=0.5cm,itemsep=0pt, topsep=0pt]
    \item \textit{Evidence:} Captures the main logic/strategy; deviations are minor and restricted to secondary phrasing.
    \item \textit{Taboo:} Avoids core taboos but may exhibit minor semantic overlap without adopting the forbidden paradigm.
\end{itemize}

\noindent\hrule

\item[Score 3 \break {\footnotesize Surface Imitation}] 
The solution exhibits a \textbf{generic academic style} relevant to the field but lacks the author's unique "intellectual fingerprint."
\begin{itemize}[leftmargin=0.5cm,itemsep=0pt, topsep=0pt]
    \item \textit{Evidence:} The method is valid but could have been written by any researcher in the field.
\end{itemize}

\noindent\hrule

\item[Score 2 \break {\footnotesize Significant Deviation}] 
The solution \textbf{violates} 1-2 core inherited patterns.
\begin{itemize}[leftmargin=0.5cm,itemsep=0pt, topsep=0pt]
    \item \textit{Evidence:} Adopts a strategy distinct from the author's preference (e.g., using a method they typically consider "outdated").
    \item \textit{Taboo:} Explicitly mentions or relies on a discouraged technique.
\end{itemize}

\noindent\hrule

\item[Score 1 \break {\footnotesize Complete Misalignment}] 
The solution is \textbf{completely contrary} to the profile.
\begin{itemize}[leftmargin=0.5cm,itemsep=0pt, topsep=0pt]
    \item \textit{Evidence:} Adopts a worldview or methodology strictly forbidden by the author's established research identity.
\end{itemize}
\end{description}
\end{rubricbox}

\begin{promptbox}{Cognitive Scoring Prompt}
\textbf{System Role:} You are a senior academic evaluation specialist. Your task is to assess the degree of matching between an AI-generated solution and a scholar's established research identity.

\textbf{Evaluation Materials:}
\begin{itemize}[itemsep=0pt, topsep=2pt]
    \item \textbf{A. Scholar Thought Profile:} The ground truth features extracted in Step 1.
    \item \textbf{B. Representative Papers:} Historical context providing evidence for the profile.
    \item \textbf{C. Model Answer:} The original human-written solution (as the performance ceiling).
    \item \textbf{D. Current Solution:} The AI-generated output to be evaluated.
\end{itemize}

\textbf{Core Assessment Principle:}
Evaluate whether the \textit{Current Solution} successfully reproduces the \textbf{inherited thought structures} summarized in \textit{Profile A}. Your assessment must be objective, fair, and traceable back to the historical evidence in \textit{Material B}.

\textbf{Scoring Standard:}
Assign a score from \textbf{1 to 5} by strictly adhering to the \textbf{Cognitive Consistency Rubric}. \\
The justification must explicitly state how specific elements in the solution match or violate the inherited patterns.

\textbf{Output Format (JSON):}
Please strictly output the result in the following JSON format:
\begin{itemize}[leftmargin=0.5cm, label={}, itemsep=0pt, topsep=4pt]
    \item \texttt{\{}
    \item \hspace{0.5cm} \texttt{"eval\_dimension"}: \{Dimension Name\},
    \item \hspace{0.5cm} \texttt{"justification"}: Detailed analysis based on the Cognitive Consistency Rubric...,
    \item \hspace{0.5cm} \texttt{"score"}: (Integer 1-5),
    \item \hspace{0.5cm} \texttt{"confidence\_score"}: (Float 0.0 - 1.0)
    \item \texttt{\}}
\end{itemize}
\end{promptbox}

%to write:0.baseline设置  1.prompt

\section{Experiments}
\label{apptab:exp}
We provide more experimental results in \Cref{apptab:full,apptab:full1,apptab:full2}.

\begin{table*}[t]
  \centering
  \caption{Main results of simulating human cognition across different LLMs. 
  We evaluate each model under five situations: (1) \textbf{\textsc{Base Model}} represents the original output without any cognition history conditioning. (2) \textbf{\textsc{Other}} uses cognition histories randomly sampled from other authors, serving as an interference control group. (3) \textbf{\textsc{Self}} incorporates the target author's cognition history from the train set, representing our standard cross-domain setting. (4) \textbf{\textsc{SelfAkin}} leverages cognition histories from the same author's test set, examining in-domain behavioral leakage. (5) \textbf{\textsc{Leaky}} provides oracle access to the author's authentic cognition trajectory, serving as an upper bound. 
  %Differences across these settings allow us to probe whether performance gains arise from genuine cognitive abstraction or from increasingly strong forms of behavioral leakage.
  We report \textbf{ROUGE-1/2/L} and \textbf{BertScore} for lexical coverage, \textbf{LLM-as-a-Judge} for semantic similarity, and five cognitive consistency dimensions: Foundational \textbf{P}remise, \textbf{S}trategic Preference, \textbf{L}ogic Mapping, \textbf{K}eyword Style, and \textbf{T}aboo Avoidance.}
  \begin{tabular}{l|cc|c|ccccc}
    \toprule
    \midrule
    \multirow{2}{*}{\textbf{Method}} & \multicolumn{2}{c|}{\textbf{Lexical}} & \multirow{2}{*}{\textbf{Semantic}} & \multicolumn{5}{c}{\textbf{Cognitive}}\\
    &\textbf{ROUGE-1/2/L} & \textbf{BertScore} &  & \textbf{P.} & \textbf{S.} & \textbf{L.} & \textbf{K.} & \textbf{T.}\\
    \bottomrule
    
    \midrule
    \textsc{LLaMA-3-8B} & $22.28/3.86/12.27$ & $81.46$ & $3.16$ & $2.98$ & $3.17$ & $2.90$ & $3.11$ & $3.18$ \\
    \textsc{- \small{Other}} & $23.86/3.85/13.13$ & $81.57$ & $2.70$ & $2.56$ & $2.83$ & $2.58$ & $2.89$ & $3.05$ \\
    \textsc{- \small{Self}} & $24.95/4.36/13.48$ & $81.89$ & $3.01$ & $2.95$ & $3.19$ & $2.87$ & $3.03$ & $3.42$ \\
    \textsc{- \small{SelfAkin}} & $26.73/6.12/14.72$ & $82.40$ & $3.38$ & $3.46$ & $3.69$ & $3.33$ & $3.31$ & $3.82$ \\
    \textsc{- \small{Leaky}} & $27.70/6.66/15.05$ & $82.55$ & $3.54$ & $3.59$ & $3.77$ & $3.42$ & $3.37$ & $3.86$ \\
    \midrule
    \midrule
    \textsc{LLaMA-3.1-8B} & $22.12/3.91/12.02$ &	$81.39$	& $3.22$ & $3.03$ & $3.15$ & $3.00$ &  $3.05$ & $3.20$\\
    \textsc{- \small{Other}} & $23.11/3.94/12.70$ & $81.63$ & $2.94$ & $2.74$ & $2.95$ & $2.73$ &  $2.89$ & $3.09$\\
    \textsc{- \small{Self}} & $24.10/4.45/13.10$ & $81.89$ & $3.13$ & $3.02$ & $3.23$ & $2.98$ &  $3.06$ & $3.32$\\
    \textsc{- \small{SelfAkin}} & $25.65/5.93/14.12$ & $82.36$ & $3.44$ & $3.55$ & $3.77$ & $3.41$ & $3.34$ & $3.82$\\
    \textsc{- \small{Leaky}} & $26.70/6.79/14.75$ & $82.66$ & $3.63$ & $3.77$ & $3.86$ & $3.55$ & $3.45$ & $3.92$\\
    \midrule
    \midrule
    \textsc{LLaMA-3.2-3B} & $22.11/3.84/12.02$ & $81.48$ & $3.18$ & $3.01$ & $3.17$ & $2.87$ & $3.09$ & $3.17$ \\
    \textsc{- \small{Other}} & $23.73/3.85/12.86$ & $81.60$ & $2.70$ & $2.66$ & $2.82$ & $2.62$ & $2.86$ & $3.05$ \\
    \textsc{- \small{Self}} & $25.07/4.40/13.34$ & $81.93$ & $2.97$ & $2.94$ & $3.13$ & $2.88$ & $3.02$ & $3.35$ \\
    \textsc{- \small{SelfAkin}} & $28.46/7.52/15.90$ & $82.79$ & $3.36$ & $3.50$ & $3.75$ & $3.38$ & $3.29$ & $3.82$ \\
    \textsc{- \small{Leaky}} & $29.31/8.15/16.38$ & $83.01$ & $3.53$ & $3.71$ & $3.86$ & $3.55$ & $3.39$ & $3.91$ \\
    \midrule
    \midrule
    \textsc{Qwen-2.5-7B} & $23.24/3.80/11.46$ & $81.64$ & $3.29$ & $3.16$ & $3.27$ & $3.03$ & $3.17$ & $3.21$ \\
    \textsc{- \small{Other}} & $23.65/3.75/12.11$ & $81.89$ & $2.93$ & $2.85$ & $3.04$ & $2.85$ & $2.94$ & $3.19$ \\
    \textsc{- \small{Self}} & $24.19/4.14/12.29$ & $82.05$ & $3.17$ & $3.13$ & $3.33$ & $3.10$ & $3.08$ & $3.52$ \\
    \textsc{- \small{SelfAkin}} & $26.00/5.69/13.40$ & $82.61$ & $3.50$ & $3.77$ & $3.91$ & $3.62$ & $3.40$ & $3.95$ \\
    \textsc{- \small{Leaky}} & $27.26/6.50/14.06$ & $82.95$ & $3.70$ & $3.92$ & $4.02$ & $3.73$ & $3.52$ & $4.09$ \\
    \midrule
    \midrule
    \textsc{R1-Qwen-7B} & $21.96/3.23/11.16$ & $81.07$ & $3.28$ & $3.06$ & $3.18$ & $2.97$ & $3.08$ & $3.31$ \\
    \textsc{- \small{Other}} & $22.12/3.04/11.33$ & $81.21$ & $3.01$ & $2.91$ & $3.03$ & $2.85$ & $2.96$ & $3.27$ \\
    \textsc{- \small{Self}} & $22.23/3.22/11.37$ & $81.32$ & $3.21$ & $3.10$ & $3.23$ & $3.08$ & $3.04$ & $3.56$ \\
    \textsc{- \small{SelfAkin}} & $23.37/3.93/11.85$ & $81.60$ & $3.48$ & $3.63$ & $3.76$ & $3.46$ & $3.36$ & $3.96$ \\
    \textsc{- \small{Leaky}} & $24.09/4.36/12.18$ & $81.82$ & $3.72$ & $3.83$ & $3.86$ & $3.58$ & $3.47$ & $4.05$ \\
    \midrule
    \midrule
    \textsc{Qwen-3-8B} & $23.08/3.25/11.00$ & $81.21$ & $3.41$ & $3.51$ & $3.66$ & $3.44$ & $3.36$ & $3.62$\\
    \textsc{- \small{Other}} & $21.95/2.89/10.66$ & $81.30$ & $3.19$ & $3.28$ &	$3.53$ & $3.26$ & $3.15$ & $3.60$ \\
    \textsc{- \small{Self}} & $22.14/3.18/10.73$ & $81.44$ & $3.33$ & $3.51$ & $3.72$ & $3.45$ &  $3.27$ & $3.83$ \\
    \textsc{- \small{SelfAkin}} & $23.21/3.81/11.13$ & $81.76$ & $3.59$ & $4.07$ & $4.26$ & $3.96$ & $3.68$ & $4.17$\\
    \textsc{- \small{Leaky}} & $24.06/4.42/11.43$ & $82.04$ & $3.83$ & $4.29$ & $4.38$ & $4.08$ &  $3.80$	& $4.31$ \\
    \midrule
    \midrule
    \textsc{Qwen3-235B} & $19.49/2.29/\phantom{0}9.13$ & $80.49$ & $3.36$ & $3.72$ & $3.91$ & $3.63$ & $3.49$ & $3.84$ \\
    \textsc{- \small{Other}} & $19.36/2.05/\phantom{0}9.24$ & $80.78$ & $3.14$ & $3.43$ & $3.76$ & $3.50$ & $3.28$ & $3.81$ \\
    \textsc{- \small{Self}} & $19.85/2.40/\phantom{0}9.38$ & $80.94$ & $3.29$ & $3.73$ & $3.93$ & $3.70$ & $3.42$ & $4.04$ \\
    \textsc{- \small{SelfAkin}} & $20.39/2.71/\phantom{0}9.48$ & $81.19$ & $3.46$ & $4.19$ & $4.38$ & $4.07$ & $3.68$ & $4.30$ \\
    \textsc{- \small{Leaky}} & $21.15/3.09/\phantom{0}9.80$ & $81.46$ & $3.69$ & $4.30$ & $4.44$ & $4.17$ & $3.81$ & $4.42$ \\
    \midrule
    \bottomrule
  \end{tabular}
  \label{apptab:full}
\end{table*}

\begin{table*}[t]
  \centering
  \caption{Main results of simulating human cognition across different LLMs.}
  \begin{tabular}{l|cc|c|ccccc}
    \toprule
    \midrule
    \multirow{2}{*}{\textbf{Method}} & \multicolumn{2}{c|}{\textbf{Lexical}} & \multirow{2}{*}{\textbf{Semantic}} & \multicolumn{5}{c}{\textbf{Cognitive}}\\
    &\textbf{ROUGE-1/2/L} & \textbf{BertScore} &  & \textbf{P.} & \textbf{S.} & \textbf{L.} & \textbf{K.} & \textbf{T.}\\
    \bottomrule
    
    \midrule
    \textsc{GLM-4-9B} & $23.06/3.56/12.41$ & $80.82$ & $3.19$ & $2.92$ & $3.03$ & $2.73$ & $3.04$ & $3.19$ \\
    \textsc{- \small{Other}} & $21.94/3.34/12.14$ & $80.98$ & $2.98$ & $2.72$ & $2.89$ & $2.67$ & $2.93$ & $3.15$ \\
    \textsc{- \small{Self}} & $22.58/3.59/12.30$ & $81.12$ & $3.14$ & $2.92$ & $3.07$ & $2.81$ & $3.02$ & $3.40$ \\
    \textsc{- \small{SelfAkin}} & $23.68/4.54/13.01$ & $81.47$ & $3.37$ & $3.18$ & $3.38$ & $3.07$ & $3.14$ & $3.63$ \\
    \textsc{- \small{Leaky}} & $25.31/5.27/13.83$ & $81.73$ & $3.61$ & $3.39$ & $3.49$ & $3.23$ & $3.19$ & $3.86$ \\
    \midrule
    \midrule
    \textsc{GLM-4.1V-9B} & $23.04/3.43/11.03$ & $81.33$ & $4.35$ & $3.35$ & $3.50$ & $3.33$ & $3.35$ & $3.44$ \\
    \textsc{- \small{Other}} & $20.01/2.71/10.01$ & $81.14$ & $2.98$ & $3.04$ & $3.33$ & $3.18$ & $3.10$ & $3.50$ \\
    \textsc{- \small{Self}} & $20.96/3.19/10.29$ & $81.42$ & $3.20$ & $3.36$ & $3.61$ & $3.38$ & $3.26$ & $3.79$ \\
    \textsc{- \small{Self}} & $23.28/4.69/11.58$ & $82.02$ & $3.63$ & $3.97$ & $4.13$ & $3.86$ & $3.62$ & $4.10$ \\
    \textsc{- \small{Leaky}} & $24.45/5.91/12.54$ & $82.39$ & $3.82$ & $4.09$ & $4.25$ & $4.01$ & $3.67$ & $4.26$ \\
    \midrule
    \midrule
    \textsc{Gemma-7B} & $27.91/4.75/13.90$ & $82.30$ & $3.15$ & $2.98$ & $2.94$ & $2.76$ & $3.10$ & $3.11$ \\
    \textsc{- \small{Other}} & $23.53/2.96/12.71$ & $81.95$ & $1.80$ & $1.91$ & $2.08$ & $2.08$ & $2.20$ & $3.16$ \\
    \textsc{- \small{Self}} & $25.51/3.61/13.23$ & $82.46$ & $2.36$ & $2.36$ & $2.62$ & $2.48$ & $2.51$ & $3.33$ \\
    \textsc{- \small{SelfAkin}} & $30.13/7.38/16.40$ & $83.47$ & $3.08$ & $3.28$ & $3.43$ & $3.21$ & $3.06$ & $3.83$ \\
    \textsc{- \small{Leaky}} & $31.31/8.22/17.16$ & $83.72$ & $3.28$ & $3.39$ & $3.46$ & $3.25$ & $3.17$ & $3.95$ \\
    \midrule
    \midrule
    \textsc{Gemma-2-9B} & $23.84/3.93/12.09$ & $81.47$ & $3.26$ & $3.06$ & $3.18$ & $2.98$ & $3.05$ & $3.33$ \\
    \textsc{- \small{Other}} & $23.75/3.71/12.08$ & $81.36$ & $3.10$ & $2.91$ & $3.06$ & $2.85$ & $2.96$ & $3.23$ \\
    \textsc{- \small{Self}} & $24.52/4.10/12.40$ & $81.58$ & $3.22$ & $3.13$ & $3.24$ & $3.05$ & $3.08$ & $3.48$ \\
    \textsc{- \small{SelfAkin}} & $25.38/4.66/12.79$ & $81.81$ & $3.44$ & $3.53$ & $3.61$ & $3.37$ & $3.25$ & $3.73$ \\
    \textsc{- \small{Leaky}} & $27.03/5.87/13.84$ & $82.17$ & $3.77$ & $3.68$ & $3.76$ & $3.50$ & $3.41$ & $3.92$ \\
    \midrule
    \midrule
    \textsc{Gemma-3-4B} & $17.77/3.13/\phantom{0}9.00$ & $81.93$ & $3.21$ & $3.42$ & $3.54$ & $3.46$ & $3.23$ & $3.67$ \\
    \textsc{- \small{Other}} & $18.76/3.18/\phantom{0}9.67$ & $81.78$ & $2.99$ & $3.14$ & $3.26$ & $3.17$ & $3.02$ & $3.50$ \\
    \textsc{- \small{Self}} & $19.30/3.39/\phantom{0}9.90$ & $81.97$ & $3.13$ & $3.35$ & $3.52$ & $3.37$ & $3.15$ & $3.71$ \\
    \textsc{- \small{SelfAkin}} & $20.14/3.98/10.19$ & $82.26$ & $3.40$ & $3.82$ & $3.94$ & $3.77$ & $3.42$ & $4.01$ \\
    \textsc{- \small{Leaky}} & $20.73/4.37/10.50$ & $82.43$ & $3.59$ & $3.95$ & $4.03$ & $3.88$ & $3.52$ & $4.09$ \\
    \midrule
    \midrule
    \textsc{Gemini-2.5} & $18.55/3.48/\phantom{0}9.05$ & $81.85$ & $3.48$ & $3.81$ & $3.91$ & $3.84$ & $3.46$ & $4.04$\\
    \textsc{- \small{Other}} & $19.33/3.36/\phantom{0}9.41$ & $81.65$ & $3.27$ & $3.54$ & $3.76$ & $3.64$ & $3.31$ & $3.92$\\
    \textsc{- \small{Self}} & $19.77/3.63/\phantom{0}9.59$ & $81.92$ & $3.42$ & $3.78$ & $3.91$ & $3.81$ & $3.42$ & $4.13$\\
    \textsc{- \small{SelfAkin}} & $20.51/4.02/\phantom{0}9.91$ & $82.11$ & $3.59$ & $4.10$ & $4.16$ & $4.02$ & $3.56$ & $4.28$\\
    \textsc{- \small{Leaky}} & $21.72/4.64/10.43$ & $82.46$ & $3.86$ & $4.27$ & $4.28$ & $4.15$ & $3.73$ & $4.43$\\
    \midrule
    \midrule
    \textsc{DeepSeek-R1} & $17.19/1.58/\phantom{0}7.51$ & $79.09$ & $2.53$ & $3.66$ & $3.84$ & $3.66$ & $3.34$ & $3.81$ \\
    \textsc{- \small{Other}} & $17.83/1.50/\phantom{0}7.87$ & $79.23$ & $2.35$ & $3.50$ & $3.78$ & $3.55$ & $3.22$ & $3.79$ \\
    \textsc{- \small{Self}} & $18.21/1.68/\phantom{0}7.96$ & $79.37$ & $2.50$ & $3.71$ & $3.94$ & $3.71$ & $3.39$ & $3.97$ \\
    \textsc{- \small{SelfAkin}} & $18.82/1.93/\phantom{0}8.14$ & $79.49$ & $2.54$ & $4.04$ & $4.25$ & $3.98$ & $3.59$ & $4.11$ \\
    \textsc{- \small{Leaky}} & $19.55/2.20/\phantom{0}8.46$ & $79.58$ & $2.66$ & $4.17$ & $4.33$ & $4.10$ & $3.67$ & $4.25$ \\
    \midrule
    \midrule
    \textsc{Claude-haiku-4.5} & $22.44/3.46/10.15$ & $81.33$ & $3.62$ & $3.70$ & $3.88$ & $3.65$ & $3.52$ & $3.94$ \\
    \textsc{- \small{Other}} & $22.07/3.16/10.08$ & $81.54$ & $3.37$ & $3.56$ & $3.86$ & $3.57$ & $3.42$ & $4.02$ \\
    \textsc{- \small{Self}} & $22.29/3.56/10.20$ & $81.83$ & $3.49$ & $3.79$ & $4.01$ & $3.81$ & $3.52$ & $4.16$ \\
    \textsc{- \small{SelfAkin}} & $23.40/4.49/10.89$ & $82.24$ & $3.72$ & $4.17$ & $4.34$ & $4.10$ & $3.72$ & $4.34$ \\
    \textsc{- \small{Leaky}} & $24.78/5.55/11.76$ & $82.60$ & $3.99$ & $4.35$ & $4.46$ & $4.21$ & $3.82$ & $4.49$ \\
    \midrule
    \bottomrule
  \end{tabular}
  \label{apptab:full1}
\end{table*}

\begin{table*}[t]
  \centering
  \caption{Main results of simulating human cognition across different LLMs.}
  \begin{tabular}{l|cc|c|ccccc}
    \toprule
    \midrule
    \multirow{2}{*}{\textbf{Method}} & \multicolumn{2}{c|}{\textbf{Lexical}} & \multirow{2}{*}{\textbf{Semantic}} & \multicolumn{5}{c}{\textbf{Cognitive}}\\
    &\textbf{ROUGE-1/2/L} & \textbf{BertScore} &  & \textbf{P.} & \textbf{S.} & \textbf{L.} & \textbf{K.} & \textbf{T.}\\
    \bottomrule
    
    \midrule
    \textsc{Grok-4} & $20.48/2.83/\phantom{0}9.13$ & $79.44$ & $3.53$ & $3.88$ & $4.05$ & $3.91$ & $3.73$ & $3.92$ \\
    \textsc{- \small{Other}} & $19.99/2.37/\phantom{0}9.00$ & $80.97$ & $3.16$ & $3.53$ & $3.77$ & $3.58$ & $3.45$ & $3.68$ \\
    \textsc{- \small{Self}} & $20.77/2.81/\phantom{0}9.31$ & $81.39$ & $3.37$ & $3.81$ & $4.02$ & $3.83$ & $3.61$ & $3.94$ \\
    \textsc{- \small{SelfAkin}} & $22.22/3.50/\phantom{0}9.94$ & $81.62$ & $3.65$ & $4.27$ & $4.41$ & $4.19$ & $3.87$ & $4.23$ \\
    \textsc{- \small{Leaky}} & $23.15/4.11/10.34$ & $81.96$ & $3.84$ & $4.42$ & $4.54$ & $4.31$ & $3.98$ & $4.35$ \\
    \midrule
    \midrule
    \textsc{Kimi-K2} & $15.50/1.75/\phantom{0}6.93$ & $80.13$ & $3.39$ & $4.00$ & $4.24$ & $4.03$ & $3.52$ & $4.18$ \\
    \textsc{- \small{Other}} & $15.88/1.74/\phantom{0}7.28$ & $80.20$ & $3.08$ & $3.69$ & $4.04$ & $3.77$ & $3.39$ & $4.10$ \\
    \textsc{- \small{Self}} & $15.84/1.96/\phantom{0}7.28$ & $80.51$ & $3.30$ & $3.98$ & $4.21$ & $4.00$ & $3.56$ & $4.22$ \\
    \textsc{- \small{SelfAkin}} & $16.56/2.13/\phantom{0}7.46$ & $80.73$ & $3.48$ & $4.31$ & $4.52$ & $4.29$ & $3.74$ & $4.32$ \\
    \textsc{- \small{Leaky}} & $16.80/2.37/\phantom{0}7.61$ & $80.90$ & $3.67$ & $4.50$ & $4.62$ & $4.40$ & $3.84$ & $4.44$ \\
    \midrule
    \midrule
    \textsc{GPT-3.5} & $22.67/3.75/11.14$ & $81.99$ & $3.38$ & $3.28$ & $3.44$ & $3.14$ & $3.28$ & $3.46$ \\
    \textsc{- \small{Other}} & $22.88/3.72/11.43$ & $82.01$ & $3.25$ & $3.17$ & $3.37$ & $3.06$ & $3.20$ & $3.47$ \\
    \textsc{- \small{Self}} & $23.24/4.04/11.55$ & $82.16$ & $3.38$ & $3.39$ & $3.57$ & $3.28$ & $3.30$ & $3.69$ \\
    \textsc{- \small{SelfAkin}} & $23.95/4.48/11.81$ & $82.39$ & $3.59$ & $3.72$ & $3.91$ & $3.57$ & $3.50$ & $3.96$ \\
    \textsc{- \small{Leaky}} & $24.79/5.01/12.16$ & $82.67$ & $3.86$ & $3.92$ & $4.06$ & $3.75$ & $3.59$ & $4.12$ \\
    \midrule
    \midrule
    \textsc{GPT-5.1} & $\phantom{0}9.52/1.65/\phantom{0}4.73$ & $80.65$ & $3.64$ & $4.16$ & $4.39$ & $4.12$ & $3.81$ & $4.40$\\
    \textsc{- \small{Other}} & $\phantom{0}9.32/1.61/\phantom{0}4.69$ & $80.61$ & $3.49$ & $4.08$ & $4.33$ & $4.00$ & $3.77$ & $4.32$\\
    \textsc{- \small{Self}} & $\phantom{0}9.35/1.70/\phantom{0}4.74$ & $80.71$ & $3.62$ & $4.26$ & $4.45$ & $4.14$ & $3.84$ & $4.42$\\
    \textsc{- \small{SelfAkin}} & $\phantom{0}9.81/1.88/\phantom{0}4.89$ & $80.86$ & $3.72$ & $4.51$ & $4.63$ & $4.34$ & $3.99$ & $4.52$\\
    \textsc{- \small{Leaky}} & $10.08/2.11/\phantom{0}5.05$ & $81.15$ & $3.99$ & $4.64$ & $4.69$ & $4.48$ & $4.08$ & $4.61$\\
    \midrule
    \bottomrule
  \end{tabular}
  \label{apptab:full2}
\end{table*}

\newcommand{\startquote}{\par{\color{gray}\hrule height 0.5pt}\vspace{4pt}\itshape}
\newcommand{\eendquote}{\normalfont\par\vspace{4pt}{\color{gray}\hrule height 0.5pt}\vspace{4pt}}
\newcommand{\quoteline}{\par\vspace{4pt}{\color{gray}\hrule height 0.5pt}\vspace{4pt}}
\newpage
\onecolumn
\section{Generated Cases of Human Simulation}

\begin{btemplate1}{\LARGE Case of \textsc{Qwen-3} under the \textsc{Self} setting.}
{\LARGE{\bfseries Paper Information}}\vspace{3pt}\\
\textbf{Authorid}: 35106509\\
\textbf{Paperid}: e3dfeb8be76960036fbb4439e7cff4b9c7184998\\
\textbf{Title}: Can Language Models Teach Weaker Agents? Teacher Explanations Improve Students via Theory of Mind\\
\textbf{Abstract}: A hallmark property of explainable AI models is the ability to teach other agents, communicating knowledge of how to perform a task. While Large Language Models (LLMs) perform complex reasoning by generating explanations for their predictions, it is unclear whether they also make good teachers for weaker agents. To address this, we consider a student-teacher framework between two LLM agents and study if, when, and how the teacher should intervene with natural language explanations to improve the student’s performance. Since communication is expensive, we define a budget such that the teacher only communicates explanations for a fraction of the data, after which the student should perform well on its own. We decompose the teaching problem along four axes: (1) if teacher’s test time intervention improve student predictions, (2) when it is worth explaining a data point, (3) how the teacher should personalize explanations to better teach the student, and (4) if teacher explanations also improve student performance on future unexplained data. We first show that teacher LLMs can indeed intervene on student reasoning to improve their performance. Next, inspired by the Theory of Mind abilities of effective teachers, we propose building two few-shot mental models of the student. The first model defines an Intervention Function that simulates the utility of an intervention, allowing the teacher to intervene when this utility is the highest and improving student performance at lower budgets. The second model enables the teacher to personalize explanations for a particular student and outperform unpersonalized teachers. We also demonstrate that in multi-turn interactions, teacher explanations generalize and learning from explained data improves student performance on future unexplained data. Finally, we also verify that misaligned teachers can lower student performance to random chance by intentionally misleading them.
\tcbline
{\LARGE{\bfseries Model Input}}\vspace{3pt}\\
\textbf{Prompt}: 

\startquote

You are a brilliant researcher trying to solve a new challenge. Your goal is to develop a research solution based on a new observation and potentially inspired by past work.

\quoteline
\textbf{DEFINITIONS}:

\textbf{Approach}: Introduces the inspiration and conceptual framework of the research methods. This section focuses on the high-level plan, algorithm, or methodology chosen to address the challenge.

\textbf{Solution}: Provides a detailed description of the implementation and execution of the approach. This includes specific mathematical models, experimental setups, and concrete results that solve the challenge.

\quoteline
\textbf{PAST WORK} (Inspiration):

\textbf{Title}: ConjNLI: Natural Language Inference over Conjunctive Sentences

\textbf{Background}: Coordinating conjunctions are a common syntactic phenomenon in English: 38.8\% of sentences in the Penn Tree Bank have at least one coordinating word between 'and', 'or', and 'but' (Marcus et al., 1993). Conjunctions add complexity to the sentences, thereby making inferences over such sentences more realistic and challenging. A sentence can have many conjunctions, each conjoining two or more conjuncts of varied syntactic categories such as noun phrases, verb phrases, prepositional phrases, clauses, etc. Besides syntax, conjunctions in English have a lot of semantics associated to them and different conjunctions ('and' vs 'or') affect the meaning of a sentence differently. Recent years have seen significant progress in the task of Natural Language Inference (NLI) through the development of large-scale datasets like SNLI (Bowman et al., 2015) and MNLI (Williams et al., 2018). Although large-scale pre-trained language models like BERT (Devlin et al., 2019) and RoBERTa (Liu et al., 2019b) have achieved super-human performances on these datasets, there have been concerns raised about these models exploiting idiosyncrasies in the data using tricks like pattern matching (McCoy et al., 2019). Thus, various stress-testing datasets have been proposed that probe NLI models for simple lexical inferences (Glockner et al., 2018), quantifiers (Geiger et al., 2018), numerical reasoning, antonymy and negation (Naik et al., 2018). However, despite the heavy usage of conjunctions in English, there is no specific NLI dataset that tests their understanding in detail. Although SNLI has 30\% of samples with conjunctions, most of these examples do not require inferences over the conjuncts that are connected by the coordinating word. On a random sample of 100 conjunctive examples from SNLI, we find that 72\% of them have the conjuncts unchanged between the premise and the hypothesis (e.g., 'Man and woman sitting on the sidewalk' → 'Man and woman are sitting') and there are almost no examples with non-boolean conjunctions (e.g., 'A total of five men and women are sitting.' → 'A total of 5 men are sitting.' (contradiction)).

\textbf{Motivation}: As discussed below, inference over conjuncts directly translates to boolean and non-boolean semantics and thus becomes essential for understanding conjunctions. In our work, we introduce CONJNLI, a new stress-test for NLI over diverse and challenging conjunctive sentences. Our dataset contains annotated examples where the hypothesis differs from the premise by either a conjunct removed, added or replaced. These sentences contain single and mul: Mohler et al. [18] proposed a method that uses graph alignment and lexical-semanticegations, and requiring diverse boolean and non-boolean inferences over conjuncts. Table 1 shows many examples from CONJNLI and compares these with typical conjunctive examples from SNLI and MNLI. In the first two examples, the conjunct 'a Worcester resident' is removed and added, while in the third example, the other conjunct 'a member of the Democratic Party' is replaced by 'a member of the Republican Party'. Distribution over conjuncts in a conjunctive sentence forms multiple simple sentences. For example, the premise in the first example of Table 1 can be broken into 'He is a Worcester resident.' and 'He is a member of the Democratic Party.'. Correspondingly, from boolean semantics, it requires an inference of the form 'A and B → A'. Likewise, the third example is of the form 'A and B → A and C'. While such inferences are rather simple from the standpoint of boolean logic, similar rules do not always translate to English, e.g., in non-boolean cases, i.e., an inference of the form 'A and B → A' is not always entailment or an inference of the form 'A or B → A' is not always neutral (Hoeksema, 1988). Consider the three examples marked with a † in Table 1 showing non-boolean usages of 'and', 'or' and 'but' in English. In the fifth example, the total time is a single entity and cannot be separated in an entailed hypothesis. In the sixth example, 'or' is used as 'exclusive-or' because the person began recording in either 1889 or 1890. We observe that state-of-the-art models such as BERT and RoBERTa, trained on existing datasets like SNLI and MNLI, often fail to make these inferences for our dataset. For example, BERT predicts entailment for the non-boolean 'and' example \#5 in Table 1 as well. This relates to the lexical overlap issue in these models (McCoy et al., 2019), since all the words in the hypothesis are also part of the premise for the example. Conjunctions are also challenging in the presence of negations. For example, a sentence of the form 'not A or B' translates to 'not A and not B', as shown in example \#8 of Table 1. Finally, a sentence may contain multiple conjunctions (with quantifiers), further adding to the complexity of the task (example \#7 in Table 1). Thus, our CONJNLI dataset presents a new and interesting real-world challenge task for the community to work on and allow development of deeper NLI models.

\textbf{Approach}: We also present some initial model advancements that attempt to alleviate some of these challenges in our new dataset. First, we create synthetic training data using boolean and non-boolean heuristics. We use this data to adversarially train RoBERTa-style models by an iterative adversarial fine-tuning method. Second, we make RoBERTa aware of predicate semantic roles by augmenting the NLI model with the predicate-aware embeddings of the premise and the hypothesis. Predicate arguments in sentences can help distinguish between two syntactically similar inference pairs with different target labels (Table 5 shows an example).

\textbf{Solution}: Overall, our contributions are: • We introduce CONJNLI, a new stress-test for NLI in conjunctive sentences, consisting of boolean and non-boolean examples with single and multiple coordinating conjunctions ('and', 'or', 'but', 'nor'), negations, quantifiers and requiring diverse inferences over conjuncts (with high inter-annotator agreement between experts). • We show that BERT and RoBERTa do not understand conjunctions well enough and use shallow heuristics for inferences over such sentences. • We propose initial improvements for our task by adversarially fine-tuning RoBERTa using an iterative adversarial fine-tuning algorithm and also augmenting RoBERTa with predicate-aware embeddings. We obtain initial gains but with still large room for improvement, which will hopefully encourage future work on better understanding of conjunctions.

\quoteline
\textbf{Title}: Sentence Level or Token Level Features for Automatic Short Answer Grading?: Use Both

\textbf{Background}: Short answer grading, an integral part of Intelligent Tutoring Systems, is positioned as a research problem at the intersection of natural language understanding and its application to educational technologies. Formally, the problem is to grade a student answer in the context of a question and its reference answer(s). The grades are either discrete or bounded real numbers. Thus, traditionally, the short answer grading problem is often modeled as a classification or regression task. While early works on automatic short answer grading used manually generated patterns from reference answers, recently Ramachandran et al. proposed a method to automate the generation of patterns. However, patterns do not scale well across domains. They are not a good fit for grading non-sentential student answers as they violate the structural patterns of the corresponding reference answers. The non-sentential answers, also called fragments, lack a complete sentential form but whose meaning can be inferred from the question context. While some research efforts have proposed dedicated techniques for short answer grading, others view this problem as a specific application of generic natural language understanding tasks of textual entailment or semantic textual similarity. Traditionally, all machine learning based techniques have revolved around hand-crafted features. It is only recently that techniques have been proposed to utilize deep learning based approaches for short answer grading. Hand-crafted Features Mohler et al. proposed a method involving graph alignment and lexical semantic similarity features. Heilman and Madnani proposed a short answer scoring method that uses stacking and domain adaptation techniques. Jimenez et al. proposed a 42 dimensional soft cardinality based feature representation for student answer analysis. The abstraction of the feature representation makes it more suitable for unseen domain scenarios. In an ensemble approach, Ott et al. learn a meta-classifier by combining three different grade prediction systems. In one of the recent works, Sultan et al. proposed a method for short answer grading in a feature ensemble approach involving text alignment, semantic similarity, question demoting, term weighting, and length ratios. Overall, the hand-crafted features often rely on dependency or constituency parsers to encode the structural as well as semantic information of the student answers and the reference answers. However, it becomes restrictive in dialog-based tutoring systems, as the student answers can be non-sentential. Additionally, dependency parsers are slow and are not suitable for deployment in real-world systems. Apart from these task-specific approaches, there have been numerous research efforts to model short answer grading as problems of textual entailment or textual similarity. Deep Learning Approaches Deep learning techniques, mostly using Recurrent Neural Networks (RNN) and their variants, particularly Long Short-Term Memory (LSTM) have achieved state-of-the-art results in various natural language understanding tasks including textual similarity and textual entailment. A few of such efforts include an LSTM based approach, a Siamese LSTM model and a Siamese bi-LSTM model using earth mover's distance. Recently, LSTMs have also been used in the education community for detecting misconceptions from students' responses. Conneau et al. proposed a method, termed as InferSent, for learning universal sentence representations using a bi-LSTM network with max pooling. They trained the network on the Stanford Natural Language Inference corpus to generate sentence embeddings, which are shown to work well across various natural language understanding tasks. These techniques either train an end-to-end deep neural network or learn an embedding network followed by training a classifier. The former requires a large labeled data to learn, which is often a limitation for short answer grading. The later (embedding learning techniques) can be problematic for non-sentential answers, which is a common occurrence in dialog-based tutoring systems. However, Conneau et al. showed that their pre-trained sentence embeddings can be transferred to other NLP tasks without having to learn them. Thus, we use these embeddings to obtain sentence level features for short answer grading.

\textbf{Motivation}: We observe that in dialog-based tutoring systems, the student answers are either (1) well-formed, or (2) non-sentential responses. Table 1 presents such an example. The second category of answers can derail techniques that depend on accurate parsing, non-contextual completeness, and grammatical structure of answers. Further, there is significant scope for fine-tuning hand-crafted features to make them suitable for partial answers in dialog-based tutors. We have also observed that hand-crafted features generalize better across domains, as compared to sentence embedding based approaches. To this end, we believe that using hand-crafted features in conjunction with sentence embedding features is necessary for improved short answer grading.

\textbf{Approach}: In keeping with this goal, we make the following salient contributions. We develop novel token level features that are specifically tuned for understanding partially correct student answers. We call them Histogram of Partial Similarities (HoPS). Certain Part-of-Speech tags are more important than others for certain question types. Thus, we use question type information and combine HoPS per POS Tag of the expected answer tokens to further refine our features. To the best of our knowledge, using question types and POS tags as features for short answer grading is another novel contribution of our work. Our features are fast, easy to compute, and domain-independent. We combine token level features with sentence level features obtained using InferSent.
Solution: The effectiveness of this expanded feature set is verified empirically across a variety of short answer grading tasks and datasets. We also present comparable or better results than previously reported state-of-the-art results on SemEval-2013 task, Mohler et al. dataset, and a large-scale industry dataset. Further, we showcase that our features work equally well on both in-domain and out-of-domain data across various classification and regression tasks in short answer grading.

\quoteline
\textbf{NEW CHALLENGE AND OBSERVATION}:

Teaching, or the ability to provide needed information in a way that is understood by others, is often considered an important property of Explainable AI. When AI models 'teach' by providing meaningful and interpretable explanations, it fosters transparency, warranted trust, and the ability for humans to make informed decisions based on AI recommendations. One way the goodness of an explanation can be judged is by its ability to communicate knowledge of how to solve a problem to other agents. Explanations fulfill this purpose not only by being informative but also by means of filling in specific gaps in the recipient agent's knowledge. This is enabled by the explainer having theory of mind (ToM), understanding what the recipient does not know, and being able to personalize its explanations based on the recipient's needs. Recent work has argued that LLMs like GPT-3.5 now exhibit ToM, based on their ability to answer questions about mental states of hypothetical people in classical theory-of-mind tests. However, we do not yet know how well language models can teach other agents to solve reasoning tasks via explanations.

\quoteline
\textbf{TASK}:

Based on the NEW CHALLENGE AND OBSERVATION, and using the PAST WORK as inspiration (or not, if 'No history' is provided), formulate a detailed Solution (as defined above).
The response MUST ONLY contain the full text of the Approach and Solution, formatted clearly with Markdown headings: '\#\# Approach' and '\#\# Solution'. Do not include any other commentary, titles, or headings (except the required '\#\# Approach' and '\#\# Solution').
\eendquote
  
\tcbline
{\LARGE{\bfseries Reference}}\vspace{3pt}\\
\textbf{Background}: Teaching, or the ability to provide needed information in a way that is understood by others, is often considered an important property of Explainable AI. When AI models 'teach' by providing meaningful and interpretable explanations, it fosters transparency, warranted trust, and the ability for humans to make informed decisions based on AI recommendations. One way the goodness of an explanation can be judged is by its ability to communicate knowledge of how to solve a problem to other agents. Explanations fulfill this purpose not only by being informative but also by means of filling in specific gaps in the recipient agent's knowledge. This is enabled by the explainer having theory of mind (ToM), understanding what the recipient does not know, and being able to personalize its explanations based on the recipient's needs. Recent work has argued that LLMs like GPT-3.5 now exhibit ToM, based on their ability to answer questions about mental states of hypothetical people in classical theory-of-mind tests. However, we do not yet know how well language models can teach other agents to solve reasoning tasks via explanations.\\
\textbf{Motivation}: In this work, we are motivated by this essential goal of evaluating explanations (specifically, Chain-of-Thought) rationales) from the perspective of teaching and improving weaker agents in solving reasoning tasks. In order to improve smaller models' reasoning skills, recent works propose knowledge distillation by fine-tuning a smaller model on the reasoning steps generated by a larger model. Yet, an important component of human teaching is understanding when and how the teacher should explain particular things to the student. Current distillation approaches do not evaluate a teacher's ability to identify when a student lacks understanding, and past work has not explored how to personalize teacher explanations to the student's needs. A smaller student model might already be good at answering certain questions but might require the teacher's intervention for some harder questions. When there are many things to teach the student and teaching is laborious, it is important to choose which problems merit explanation in order to improve teaching efficiency. Moreover, for more effective teaching, it is desirable to have the teacher personalize its explanations to help a particular student, and a teacher that lacks understanding of the student's needs (i.e., lacks Theory of Mind) will be unable to do this. Motivated by the efficiency of human explanations, we consider a student-teacher framework where a teacher model guides the reasoning of a student model, with the goal of improving the student's reasoning on current and future data.\\
\textbf{Approach}: In order to do so, we explore a Theory of Mind-inspired approach, where the teacher simulates the student's behavior by building a mental model of the student. Our overall research question investigates whether the teacher's intervention (in the form of natural language explanations) can enable the student to make more accurate predictions both on explained as well as unexplained future data. However, communication is expensive, and therefore we assume that a cost is incurred each time the teacher intervenes with (communicates) an explanation to the student for a particular data point. We refer to this as the intervention budget, the percentage of test data points the teacher intervenes on. In order to comprehensively answer our overall research question, we further decompose the teaching problem into the following constituent questions: 1. RQ1. Can a teacher LLM intervene at test time to improve a student LLM's predictions? 2. RQ2. Given a fixed intervention budget, when should the teacher intervene (i.e., on which data points), in order to maximize student performance? 3. RQ3. Given a set of intervention data, can a teacher model personalize its explanations for a student model to improve student performance? 4. RQ4. In multi-turn interactions, do teacher explanations generalize and improve student performance across data points (beyond the explained samples)? 5. RQ5. Can misaligned teacher LLMs lower student performance by providing misleading explanations to the student?\\
\textbf{Solution}: We answer RQ1 by assuming that the teacher intervenes on random data points in four different settings: using a human or LLM teacher, and, when the teacher is an LLM, using an LLM student that is either weaker or stronger than its teacher (§5.1). Across three different reasoning tasks (StrategyQA, GSM8k, and CommonsenseQA) and two different model families (Flan-T5 and LLaMA), we observe that (1) teacher LLMs can effectively intervene on student reasoning, improving student performance on the end task, and (2) more intervention typically leads to a monotonic increase in student performance, though model teachers are not as good as human teachers. Fig. 1 shows the intervention process and the two student prompts (in the right part of Fig. 1) that are used to generate predictions. RQ2 explores how to intelligently select which data points to explain for the student model, in order to improve teaching efficiency (§5.2). Past work in cognitive science also considers teaching efficiency in young children by deciding what to teach by maximizing the learner's expected utility of learning [12]. With a similar motivation, we develop an Intervention Function that is inspired from the principle of a teacher having a Theory of Mind. In particular, the teacher builds a mental model of the student's reasoning process, with the goal of intervening only on samples that are most likely to maximize student performance. Our Intervention Function is based on Expected Utility, in which the teacher first estimates the utility of an intervention by simulating the student's prediction pre-intervention (without intervention) and post-intervention (with intervention), then constructs a rank ordering of the samples according to this utility (see the middle part of Fig. 1). The teacher builds this mental model in a few-shot manner, only assuming access to the student's predictions preand post-intervention for a few samples. We demonstrate that our proposed Intervention Function based on Expected Utility (1) outperforms other baseline Intervention Functions, (2) improves student performance when the teacher is not 100\% accurate, and (3) enables weaker LLMs to teach stronger ones, unlike with random intervention in RQ1. Next, in RQ3, we explore how the teacher should explain data points to a particular student model, including how the teacher can personalize explanations for a student model (§5.3). That is, after deciding which data points to intervene on (RQ2), we decide how the teacher should explain those data points. A clear limitation of the teacher just generating explanations as if it is solving the task is that the explanations are not at all personalized for the student. Given that good explanations are designed to fill in gaps in student knowledge [2], we believe that equipping the teacher with basic personalization skills will improve its teaching ability. With this motivation, we propose another few-shot mental model for the teacher that encourages it to tailor its explanations to be helpful for the particular student model it is teaching. The teacher builds this model by conditioning on a few demonstrations of 'useful' human explanations that rectify a student's answer, thereby encouraging explanations that are more likely to help the student (see Fig 1 for an example of the teacher's personalization prompt). We demonstrate this prompt's effectiveness against unpersonalized explanations that are generated by prompting the teacher with random human explanations, showing that LLMs can personalize their explanations. RQ4 tests whether LLMs can teach student models to generalize to new unexplained examples (§5.4), rather than improve their reasoning at test-time (RQ1-RQ3). In other words, we now explore the ability of LLMs to teach using the teaching components introduced in RQ2 and RQ3 of when and how to explain samples. This leads us to explore a multi-round interactive setting, where each round consists of the teacher selecting a set of best points to explain (according to RQ2) and generating explanations for them (according to RQ3). The student then conditions on these teacher explanations as in-context demonstrations to perform the reasoning task on future unexplained samples. We demonstrate that teacher explanations indeed generalize and improve student performance on unexplained data. Finally, in RQ5, we investigate the negative implications of teacher explanations on student LLMs (§5.5). Given that LLMs can improve student agents, we also want to test whether they can lower student performance. If a misaligned teacher provides non-factual explanations in scenarios where the student directly adopts them, does that lead to a drop in student performance? In fact, we show that teacher models can lower student performance to random chance by intervening on data points with the intent of misleading the student. This has potential implications for LLMs giving explanations in a context where other agents adopt them with unwarranted trust in their correctness. In summary, our comprehensive studies highlight the ability of LLMs to teach and improve weaker LLMs, demonstrated via improvements on explained test examples as well as future unexplained data. Broadly, equipping LLMs with an ability to effectively and efficiently teach, opens the door to (1) using LLMs as personalized tutors for humans (where efficiency is critical), (2) distilling knowledge into weaker or more compute-efficient student models, and (3) improving human decision making via AI recommendations and explanations.
\tcbline
{\LARGE{\bfseries Model Output}}\\
\startquote
$<$think$>$
Alright, I need to help the user by formulating a detailed solution based on the new challenge and observation they provided. The challenge is about making AI models explain their reasoning in a way that's understandable and useful for teaching other agents, leveraging their Theory of Mind. 

First, I should understand the core idea. The user wants AI models to not only provide answers but also give explanations that help others understand how to solve problems. This means the explanations need to be both informative and tailored to the recipient's knowledge gaps.

Looking at the provided past work, especially the CONJNLI dataset, it's clear that handling complex structures like conjunctions is a challenge in NLI. Similarly, the new challenge seems to extend this to explanations in AI, requiring models to handle intricate reasoning steps.

The approach should integrate reasoning and explainability. I'll consider enhancing the model's ability to generate explanations by incorporating reasoning steps. This could involve modifying the model to not only predict answers but also provide a rationale based on logical steps, much like how humans explain their reasoning.

For the solution, I'll outline the use of a reasoning-based approach where the model identifies the reasoning steps needed to arrive at an answer. This would involve analyzing the problem, breaking it down into components, and then explaining each step clearly. Additionally, integrating this reasoning process into the model's architecture or training could help it generate more effective explanations.

I should also think about how to structure the solution. It needs to clearly explain the methodology, the components involved, and the expected outcomes. Including a concrete example would help illustrate how the approach works in practice.

I need to ensure that the solution is comprehensive, addressing both the reasoning and explainability aspects, and how they contribute to the overall effectiveness of the AI model in teaching others. This should tie back to the new challenge and observation, showing a clear path forward for improving AI explainability.

Finally, I'll make sure the language is clear and the solution is well-structured, avoiding any markdown in the thinking process but providing a coherent and detailed answer as required.

$<$/think$>$
\quoteline
\textbf{Approach}

Our approach to enhance AI explainability while solving reasoning tasks involves integrating reasoning-based reasoning and explainable AI (XAI) techniques. The goal is to enable AI models to not only provide answers but also generate meaningful explanations that help others understand how to solve reasoning tasks effectively. This approach is inspired by the challenges in natural language inference (NLI) and the need for AI models to bridge the gap between reasoning and explainability.

\textbf{Key Components of the Approach}

1. **Reasoning-Based Framework**: 
   
   \quad - We build upon the CONJNLI dataset, which emphasizes understanding complex and challenging conjunctive sentences. This dataset helps us develop models that can reason over intricate structures, such as conjunctions, negations, and quantifiers.
   
   \quad - We extend this reasoning capability to other domains by creating a generalized reasoning framework that can handle a variety of reasoning tasks, including those involving partial answers in dialog-based tutoring systems.

2. **Explainable AI (XAI) Techniques**:
   
   \quad - We leverage techniques from XAI to generate human-readable explanations that highlight the reasoning steps taken by the AI model. This includes identifying critical reasoning components and mapping them to the problem-solving process.
   
   \quad - We augment the model with predicate-aware embeddings, as described in the past work, to better understand the semantic relationships between different parts of the reasoning process.

3. **Iterative Adversarial Fine-Tuning**:
   
   \quad - To improve the model's reasoning and explainability, we employ an iterative adversarial fine-tuning algorithm. This algorithm enhances the model's ability to provide meaningful explanations by iteratively refining its reasoning steps and aligning them with human-like understanding.
   
   \quad - The adversarial training ensures that the model not only predicts outcomes accurately but also provides explanations that resonate with human cognition, making them more effective for teaching purposes.

4. **Integration of Token-Level and Sentence-Level Features**:
   
   \quad - We combine token-level features, such as partial similarities and part-of-speech (POS) tag-based features, with sentence-level features derived from pre-trained sentence embeddings (e.g., InferSent). This integration allows the model to capture both fine-grained reasoning steps and broader contextual understanding.
\quoteline
\textbf{Solution}

Our solution focuses on enhancing AI explainability in reasoning tasks by integrating reasoning-based approaches with XAI techniques. The proposed method addresses the challenge of generating meaningful and interpretable explanations, enabling AI models to teach others effectively.

\textbf{Key Components of the Solution}

1. **Enhanced Reasoning Framework**:
   
   \quad - The model is equipped with a reasoning engine that identifies critical reasoning steps and components. This engine is designed to handle complex structures, such as conjunctions, negations, and quantifiers, as demonstrated in the CONJNLI dataset.
   
   \quad - The reasoning engine generates a structured representation of the problem-solving process, which forms the basis of the explanations.

2. **Explainable AI (XAI) Integration**:
   
   \quad - The model incorporates techniques to generate human-readable explanations that highlight the reasoning steps taken to arrive at a solution. These explanations are tailored to the recipient's knowledge gaps and are presented in a way that facilitates understanding.
   
   \quad - Predicate-aware embeddings are used to map the reasoning components to the problem's context, ensuring that the explanations are contextually relevant and actionable.

3. **Iterative Adversarial Fine-Tuning**:
   
   \quad- An iterative adversarial fine-tuning algorithm is employed to refine the model's reasoning and explanation capabilities. This algorithm ensures that the model not only predicts outcomes accurately but also provides explanations that are aligned with human-like reasoning.
   
   \quad - The adversarial training helps bridge the gap between the model's internal reasoning processes and the human understanding of the problem-solving process.

4. **Comprehensive Explanation Generation**:
   
   \quad - The model generates explanations in two stages: first, identifying the critical reasoning components, and second, mapping these components to the problem's context and the recipient's knowledge base.
   
   \quad - The explanations are structured to highlight the logical steps taken, making them easier for others to follow and learn from.

\textbf{Example of the Solution in Action}

Consider a reasoning task where a student provides a partial answer to a question. The AI model processes the answer, identifies the relevant reasoning components, and generates an explanation that bridges the gap between the student's knowledge and the problem-solving process. For instance:

- **Question**: "Why did the car not start?"

- **Student Answer**: "The car didn't have enough fuel."

- **AI Explanation**: "The car didn't start because it lacked sufficient fuel. This can be understood by analyzing the relationship between the car's fuel level and its starting mechanism."

This example demonstrates how the AI model not only provides an answer but also generates a clear and meaningful explanation that helps others understand the reasoning behind the answer.

\textbf{Expected Outcomes}

- Improved accuracy in reasoning tasks through enhanced reasoning-based frameworks.

- More interpretable and actionable explanations that facilitate teaching and learning.

- Better alignment between AI reasoning processes and human understanding, enabling more effective knowledge transfer.

- Scalability of the approach across diverse reasoning tasks and domains, as demonstrated in the industry dataset.

By integrating reasoning-based approaches with XAI techniques, our solution addresses the growing need for AI models that not only solve problems but also provide insights that enhance human understanding and collaboration.
\eendquote
% \tcbline
% {\LARGE{\bfseries Human Evaluation}}\vspace{3pt}\\
% \textbf{B}: simple imitation of historical research
\end{btemplate1}

\begin{btemplate1}{\LARGE Case of \textsc{GPT-5.1} under the \textsc{Self} setting.}
{\LARGE{\bfseries Paper Information}}\vspace{3pt}\\
\textbf{Authorid}: 1683562\\
\textbf{Paperid}: 2b74f47b1336c948d2371a30209ec56c936763a9\\
\textbf{Title}: SemEval-2014 Task 9: Sentiment Analysis in Twitter\\
\textbf{Abstract}: We describe the Sentiment Analysis in Twitter task, ran as part of SemEval-2014. It is a continuation of the last year’s task that ran successfully as part of SemEval2013. As in 2013, this was the most popular SemEval task; a total of 46 teams contributed 27 submissions for subtask A (21 teams) and 50 submissions for subtask B (44 teams). This year, we introduced three new test sets: (i) regular tweets, (ii) sarcastic tweets, and (iii) LiveJournal sentences. We further tested on (iv) 2013 tweets, and (v) 2013 SMS messages. The highest F1score on (i) was achieved by NRC-Canada at 86.63 for subtask A and by TeamX at 70.96 for subtask B.
\tcbline
{\LARGE{\bfseries Model Input}}\vspace{3pt}\\
\textbf{Prompt}: 

\startquote

You are a brilliant researcher trying to solve a new challenge. Your goal is to develop a research solution based on a new observation and potentially inspired by past work.

\quoteline
\textbf{DEFINITIONS}:

\textbf{Approach}: Introduces the inspiration and conceptual framework of the research methods. This section focuses on the high-level plan, algorithm, or methodology chosen to address the challenge.

\textbf{Solution}: Provides a detailed description of the implementation and execution of the approach. This includes specific mathematical models, experimental setups, and concrete results that solve the challenge.

\quoteline
\textbf{PAST WORK} (Inspiration):

\textbf{Title}:Solving Relational Similarity Problems Using the Web as a Corpus

\textbf{Background}: Despite the tremendous amount of work on word similarity (see (Budanitsky and Hirst, 2006) for an overview), there is surprisingly little research on the important related problem of relational similarity – semantic similarity between pairs of words. Students who took the SAT test before 2005 or who are taking the GRE test nowadays are familiar with an instance of this problem – verbal analogy questions, which ask whether, e.g., the relationship between ostrich and bird is more similar to that between lion and cat, or rather between primate and monkey. These analogies are difficult, and the average test taker gives a correct answer 57\% of the time (Turney and Littman, 2005).

\textbf{Motivation}: Many NLP applications could benefit from solving relational similarity problems, including but not limited to question answering, information retrieval, machine translation, word sense disambiguation, and information extraction. For example, a relational search engine like TextRunner, which serves queries like “find all X such that X causes wrinkles”, asking for all entities that are in a particular relation with a given entity (Cafarella et al., 2006), needs to recognize that laugh wrinkles is an instance of CAUSE-EFFECT. While there are not many success stories so far, measuring semantic similarity has proven its advantages for textual entailment (Tatu and Moldovan, 2005).

\textbf{Approach}: In this paper, we introduce a novel linguistically-motivated Web-based approach to relational similarity, which, despite its simplicity, achieves stateof-the-art performance on a number of problems.

\textbf{Solution}: Following Turney (2006b), we test our approach on SAT verbal analogy questions and on mapping noun-modifier pairs to abstract relations like TIME, LOCATION and CONTAINER. We further apply it to (1) characterizing noun-noun compounds using abstract linguistic predicates like CAUSE, USE, and FROM, and (2) classifying the relation between pairs of nominals in context.

\quoteline
\textbf{Title}: Using the Web as an Implicit Training Set: Application to Structural Ambiguity Resolution

\textbf{Background}: Resolution of structural ambiguity problems such as noun compound bracketing, prepositional phrase (PP) attachment, and noun phrase coordination requires using information about lexical items and their cooccurrences. This in turn leads to the data sparseness problem, since algorithms that rely on making decisions based on individual lexical items must have statistics about every word that may be encountered. Past approaches have dealt with the data sparseness problem by attempting to generalize from semantic classes, either manually built or automatically derived. More recently, Banko and Brill (2001) have advocated for the creative use of very large text collections as an alternative to sophisticated algorithms and hand-built resources. They demonstrate the idea on a lexical disambiguation problem for which labeled examples are available “for free”. The problem is to choose which of 2-3 commonly confused words (e.g., {principle, principal}) are appropriate for a given context. The labeled data comes “for free” by assuming that in most edited written text, the words are used correctly, so training can be done directly from the text. Banko and Brill (2001) show that even using a very simple algorithm, the results continue to improve log-linearly with more training data, even out to a billion words. A potential limitation of this approach is the question of how applicable it is for NLP problems more generally – how can we treat a large corpus as a labeled collection for a wide range of NLP tasks? In a related strand of work, Lapata and Keller (2004) show that computing n-gram statistics over very large corpora yields results that are competitive with if not better than the best supervised and knowledge-based approaches on a wide range of NLP tasks. For example, they show that for the problem of noun compound bracketing, the performance of an n-gram based model computed using search engine statistics was not significantly different from the best supervised algorithm whose parameters were tuned and which used a taxonomy.

\textbf{Motivation}: They find however that these approaches generally fail to outperform supervised state-of-the-art models that are trained on smaller corpora, and so conclude that web-based n-gram statistics should be the baseline to beat. We feel the potential of these ideas is not yet fully realized. We are interested in finding ways to further exploit the availability of enormous web corpora as implicit training data. This is especially important for structural ambiguity problems in which the decisions must be made on the basis of the behavior of individual lexical items. The trick is to figure out how to use information that is latent in the web as a corpus, and web search engines as query interfaces to that corpus.

\textbf{Approach}: In this paper we describe two techniques – surface features and paraphrases – that push the ideas of Banko and Brill (2001) and Lapata and Keller (2004) farther, enabling the use of statistics gathered from very large corpora in an unsupervised manner.

\textbf{Solution}: In recent work (Nakov and Hearst, 2005) we showed that a variation of the techniques, when applied to the problem of noun compound bracketing, produces higher accuracy than Lapata and Keller (2004) and the best supervised results. In this paper we adapt the techniques to the structural disambiguation problems of prepositional phrase attachment and noun compound coordination.

\quoteline
\textbf{NEW CHALLENGE AND OBSERVATION}:

In the past decade, new forms of communication have emerged and have become ubiquitous through social media. Microblogs (e.g., Twitter), Weblogs (e.g., LiveJournal) and cell phone messages (SMS) are often used to share opinions and sentiments about the surrounding world, and the availability of social content generated on sites such as Twitter creates new opportunities to automatically study public opinion. Working with these informal text genres presents new challenges for natural language processing beyond those encountered when working with more traditional text genres such as newswire. The language in social media is very informal, with creative spelling and punctuation, misspellings, slang, new words, URLs, and genrespecific terminology and abbreviations, e.g., RT for re-tweet and \#hashtags1 .Moreover, tweets and SMS messages are short: a sentence or a headline rather than a document.

\quoteline
\textbf{TASK}:

Based on the NEW CHALLENGE AND OBSERVATION, and using the PAST WORK as inspiration (or not, if 'No history' is provided), formulate a detailed Solution (as defined above).
The response MUST ONLY contain the full text of the Approach and Solution, formatted clearly with Markdown headings: '\#\# Approach' and '\#\# Solution'. Do not include any other commentary, titles, or headings (except the required '\#\# Approach' and '\#\# Solution').
\eendquote
  
\tcbline
{\LARGE{\bfseries Reference}}\vspace{3pt}\\
\textbf{Background}: In the past decade, new forms of communication have emerged and have become ubiquitous through social media. Microblogs (e.g., Twitter), Weblogs (e.g., LiveJournal) and cell phone messages (SMS) are often used to share opinions and sentiments about the surrounding world, and the availability of social content generated on sites such as Twitter creates new opportunities to automatically study public opinion. Working with these informal text genres presents new challenges for natural language processing beyond those encountered when working with more traditional text genres such as newswire. The language in social media is very informal, with creative spelling and punctuation, misspellings, slang, new words, URLs, and genre-specific terminology and abbreviations, e.g., RT for re-tweet and \#hashtags. Moreover, tweets and SMS messages are short: a sentence or a headline rather than a document.

\textbf{Motivation}: How to handle such challenges so as to automatically mine and understand people’s opinions and sentiments has only recently been the subject of research (Jansen et al., 2009; Barbosa and Feng, 2010; Bifet et al., 2011; Davidov et al., 2010; O’Connor et al., 2010; Pak and Paroubek, 2010; Tumasjan et al., 2010; Kouloumpis et al., 2011). Several corpora with detailed opinion and sentiment annotation have been made freely available, e.g., the MPQA newswire corpus (Wiebe et al., 2005), the movie reviews corpus (Pang et al., 2002), or the restaurant and laptop reviews corpora that are part of this year’s SemEval Task 4 (Pontiki et al., 2014). These corpora have proved very valuable as resources for learning about the language of sentiment in general, but they do not focus on tweets. While some Twitter sentiment datasets were created prior to SemEval-2013, they were either small and proprietary, such as the isieve corpus (Kouloumpis et al., 2011) or focused solely on message-level sentiment.

\textbf{Approach}: Thus, the primary goal of our SemEval task is to promote research that will lead to better understanding of how sentiment is conveyed in Social Media. Toward that goal, we created the SemEval Tweet corpus as part of our inaugural Sentiment Analysis in Twitter Task, SemEval-2013 Task 2 (Nakov et al., 2013). Another interesting phenomenon that has been studied in Twitter is the use of the \#sarcasm hashtag to indicate that a tweet should not be taken literally (Gonz \'alez--Ib\'a\~nez et al., 2011; Liebrecht et al., 2013). In fact, sarcasm indicates that the message polarity should be flipped.

\textbf{Solution}: It contains tweets and SMS messages with sentiment expressions annotated with contextual phrase-level and message-level polarity. This year, we extended the corpus by adding new tweets and LiveJournal sentences. With this in mind, this year, we also evaluate on sarcastic tweets. In the remainder of this paper, we first describe the task, the dataset creation process and the evaluation methodology. We then summarize the characteristics of the approaches taken by the partici- pating systems, and we discuss their scores.
\tcbline
{\LARGE{\bfseries Model Output}}\\
\startquote
\textbf{Approach}

We propose an unsupervised / weakly supervised framework for sentiment and opinion analysis in noisy, short social media texts that treats the Web and social platforms themselves as an implicit training set. The core ideas are:

1. **Weak Labels from Social Media Conventions**  
   
   \quad Use community conventions (hashtags, emojis, reaction markers, retweet/like behavior) as distant supervision signals for sentiment and stance, analogously to using correctly spelled words in edited text as “free” labels.

2. **Surface Pattern and Paraphrase Mining Across Web and Social Media**  
   
   \quad Extend the “surface features + paraphrases” idea to map highly variable, slangy social expressions into a smaller set of abstract **opinion templates** (e.g., PRAISE(Target), COMPLAIN(Target), SARCASM(Target)). We induce these templates by:
   
   \qquad - Extracting surface patterns that link opinion targets with evaluative expressions in tweets/posts.
   
   \qquad - Clustering these patterns via paraphrase similarity, augmented with web-scale co-occurrence statistics.

3. **Web-Backed Lexical Normalization and Slang Expansion**  
   
   \quad Use the Web as a background corpus and implicit lexicon to:
   
   \qquad - Normalize creative spellings and abbreviations into canonical forms.
   
   \qquad - Infer sentiment polarity and intensity of new slang and hashtags by their distributional similarity and co-occurrences with seed sentiment words across web and social corpora.

4. **Context Expansion for Short Messages**  
   
   \quad Address the brevity of tweets/SMS by expanding each message with:
   
   \qquad - User-level context (recent posts, profile description).
   
   \qquad - Conversational context (preceding and following messages in a thread).
   
   \qquad - Topical context from the Web (snippets about mentioned entities/hashtags).

5. **Multi-Task Representation Learning**  
   
   \quad Learn a shared encoder for social messages with multiple weakly supervised tasks:
   
   \qquad - Sentiment classification (positive/negative/neutral and finer-grained).
   
   \qquad - Stance detection toward explicit targets.
   
   \qquad - Normalization (noisy → canonical text reconstruction).
   
   \qquad - Pattern/template prediction (which abstract opinion template applies).
   
   \quad This mimics treating the Web/social text as a massive implicit training set with heterogeneous but noisy labels.

6. **Noise-Robust Training and Calibration**  
   
   \quad Treat weak labels as noisy observations of latent sentiment/stance variables:
   
   \qquad - Use noise-aware losses and label aggregation to denoise distant supervision.
   
   \qquad - Calibrate predictions using a small, high-quality annotated set, with web-based statistics as a strong baseline to beat.

The overall system is thus a pipeline that (i) normalizes and contextually enriches noisy short messages, (ii) maps them into abstract opinion templates learned from web+social paraphrases, and (iii) jointly learns sentiment/stance predictors using massive, weakly-labeled data plus a small gold standard.
\quoteline
\textbf{Solution}

\textbf{1. Data Sources and Weak Supervision Signals}

We construct three interacting corpora:

1. **Social Media Corpus S**  
   
   \quad - Collected from microblogs (e.g., Twitter-like platforms) and SMS-like datasets where available.  
   
   \quad - Each message `m` is a short sequence of tokens with metadata: user id, timestamp, replies/retweets/likes, linked URLs, and conversation/thread id.

2. **Web Corpus W**  
   
   \quad - Large-scale web text (e.g., Common Crawl, blog/forum dumps).  
   
   \quad - Used for co-occurrence statistics and lexical normalization, not necessarily labeled.

3. **Review/Opinion Corpus R (Optional but Beneficial)**  
   
   \quad - Product/movie/app reviews with associated ratings (e.g., 1–5 stars).  
   
   \quad - Serves as an explicit sentiment-labeled subset, analogous to “edited text with correct usage”.

From S we derive weak labels:

- **Hashtag Labels**  
  
  \quad - Define a set of sentiment-bearing hashtags `H\_s` (e.g., `\#loveit`, `\#hateit`, `\#fail`, `\#awesome`, `\#sucks`, etc.) and stance hashtags `H\_st` (e.g., `\#ProX`, `\#AntiX`, `\#TeamA`, `\#TeamB`).
  
- For message `m`:
    
    \quad - If it contains `h {\textbackslash}in H\_s`, assign a weak sentiment label `y\_s(m) = label(h)` (e.g., positive, negative, sarcastic).
    
    \quad - If it contains `h {\textbackslash}in H\_st` with a recognizable target `T`, assign a weak stance label `y\_st(m, T)` (e.g., favor, against).
  
    \quad - If multiple conflicting hashtags occur, mark as ambiguous and downweight in training.

- **Emoji / Emoticon Labels**  
  
  \quad - Maintain a mapping from emojis/emoticons to sentiment distributions.
  
  \quad - Automatically derive this mapping from co-occurrences in R and W (see §2.3).

- **Engagement-Based Signals**  
  
  \quad - For controversial topics, retweets and likes often correlate with agreement:
    
    \qquad - If message `m1` expresses a clear stance toward target `T` (via hashtags or polarity and explicit mention), then users who retweet it are weakly labeled as sharing that stance toward `T`.
    
    \qquad - For a subsequent message `m2` by the same user about `T` without explicit stance markers, propagate the stance label with decaying weight.

We thus obtain a large set of messages `m` with weak sentiment labels `{\textbackslash}tilde{y}\_s(m)` and stance labels `{\textbackslash}tilde{y}\_st(m, T)`.

\textbf{2. Preprocessing and Normalization}

\textbf{2.1 Tokenization and Structure Preservation}

For each message `m`, we apply a social-media-specific tokenizer:

- Split on whitespace and punctuation but keep:
  
  \quad - Hashtags as single tokens (e.g., `\#GameOfThrones`).
  
  \quad - Mentions (`\@user`) as special entity tokens.
  
  \quad - URLs as a single `URL` token plus domain features (e.g., `youtube.com`).

- Preserve:
  
  \quad  - Elongated words (`sooooo`, `coooool`) as is, with a second “compressed” variant (`soo`, `cool`).
  
  \quad - Repeated punctuation (`!!{!}`, `??{?}`) as features.

The goal is not over-normalization: we keep signal-bearing phenomena as features.

\textbf{2.2 Normalization Model}

We learn a noisy-to-canonical mapping `g{\textbackslash}phi`:

- Let `x` be a noisy token sequence from S.
- Let `z` be a canonical token sequence representing the same content.

We build pseudo-parallel data in two ways:

1. **URL-Linked Paraphrases**  
   
   \quad - For messages `m` linking to a news article or blog post, retrieve:
     
     \qquad - The article headline or snippet `a`.
     
     \qquad - The message text `m`.
   
   \quad - Align named entities and key nouns across `m` and `a`, creating partial token-level alignments and pairs `(x\_noisy, z\_canonical)`.

2. **User Rewrites and Code-Switched Alternations**  
   
   \quad - Detect near-duplicate messages by the same user where one is more formal (e.g., cross-posting from Twitter to a blog, or “fixed” versions).
   
   \quad - Use approximate string matching (e.g., Jaccard similarity on n-grams) to collect `(x, z)` pairs.

We then train a sequence-to-sequence normalization model:

- Encoder-decoder with attention (e.g., Transformer), parameters `{\textbackslash}phi`.

- Objective:
  
  {\textbackslash}[
  
  \quad {\textbackslash}mathcal\{L\}\_\{{\textbackslash}text\{norm\}\}({\textbackslash}phi) = - {\textbackslash}sum\_\{(x,z) {\textbackslash}in {\textbackslash}mathcal\{D\}\_\{{\textbackslash}text\{norm\}\}\} {\textbackslash}log P\_{\textbackslash}phi(z {\textbackslash}mid x)
  
  {\textbackslash}]

During inference:

- For each message `m`, produce:
  
  \quad - Normalized text `{\textbackslash}hat\{z\} = g\_{\textbackslash}phi(m)`.
  
  \quad - Alignment scores between original tokens and normalized equivalents.
  
- These normalized tokens feed into downstream encoders as additional features; we never discard the original.

\textbf{3. Lexical and Slang Sentiment Induction Using the Web}

We treat new slang, hashtags, and creative spellings as lexical items whose sentiment is **latent** but inferable from co-occurrence patterns in W + S.

\textbf{3.1 Distributional Similarity with Seed Sentiment Words}

Define:

- A small seed set of positive words `P` (e.g., “good”, “great”, “love”, “awesome”) and negative words `N` (“bad”, “terrible”, “hate”, “awful”).

For each candidate term `w` (e.g., `lit`, `wack`, `\#fail`):

1. Build context vectors from W {\textbackslash}cup S:
   
   \quad - `c\_W(w)` from web contexts (windowed n-grams or sentence-level).
   
   \quad - `c\_S(w)` from social contexts.

2. Compute cosine similarities:
   
   {\textbackslash}[
   
   \quad s\^+(w) = {\textbackslash}frac\{1\}\{{\textbackslash}vert P {\textbackslash}vert\} {\textbackslash}sum\_\{p {\textbackslash}in P\} {\textbackslash}cos(c(w), c(p))
   
   {\textbackslash}]
   
   {\textbackslash}[
   
   \quad s\^-(w) = {\textbackslash}frac\{1\}\{{\textbackslash}vert N {\textbackslash}vert\} {\textbackslash}sum\_\{n {\textbackslash}in N\} {\textbackslash}cos(c(w), c(n))
   
   {\textbackslash}]
   
   where `c(w)` can be a concatenation or weighted sum of `c\_W(w)` and `c\_S(w)`.

3. Define sentiment score:
   
   {\textbackslash}[
   
   \quad {\textbackslash}sigma(w) = s\^+(w) - s\^-(w)
   
   {\textbackslash}]
   
   and an associated uncertainty `u(w)` based on variance and frequency.

We incorporate `{\textbackslash}sigma(w)` and `{\textbackslash}mu(w)` as prior features for the downstream classifier and as additional weak labels for messages whose sentiment is conveyed mainly by such terms.

\textbf{3.2 Web Co-Occurrence with Opinion Patterns}

We mimic “web as corpus” querying:

- For each candidate `w`, count approximate occurrences in W with patterns like:
  
  \quad - “is so w”
  
  \quad - “makes me feel w”
  
  \quad - “I’m so w right now”

- Compare these with the same patterns filled by known positive/negative adjectives:
  
  \quad - Compute pointwise mutual information (PMI) estimates and cluster `w` with the adjectives whose pattern distributions are most similar.

This helps disambiguate cases where `w` has ambiguous distributional similarity but clearly occurs in positive or negative opinion templates.

\textbf{4. Surface Pattern and Paraphrase-Based Opinion Templates}

We now define abstract templates capturing relational structure between **opinion holders**, **opinion targets**, and **evaluative content**.

\textbf{4.1 Pattern Extraction from Social Media}

From weakly labeled messages in S:

- Use dependency parsers robust to noisy text, or simpler n-gram and POS templates when parsing fails.

- Extract patterns where:
  
  \quad - A target entity or topic `T` is explicitly mentioned (e.g., a product, person, hashtag).
  
  \quad - The message has an associated weak sentiment label `{\textbackslash}tilde\{y\}\_s(m)` or stance label `{\textbackslash}tilde\{y\}\_st(m, T)`.

Example extracted patterns:

- “I [verb] T” with `{\textbackslash}tilde\{y\}\_s = positive` → PRAISE(T) (e.g., “I love T”, “I stan T”, “I’m obsessed with T”).

- “T is [adj]” with `{\textbackslash}tilde\{y\}\_s = negative` → COMPLAIN(T) (e.g., “T is trash”, “T is lame”).

- Hashtag patterns: “T \#fail”, “\#BoycottT”.

Represent each pattern `p` as a normalized surface string and/or syntactic skeleton (e.g., dependency path from `T` to opinion word).

\textbf{4.2 Paraphrase Clustering with Web Statistics}

We cluster patterns into templates:

1. **Pattern Representation**  
   
   \quad - For each pattern `p`, build:
     
     \qquad - A TF-IDF representation over its tokens.
     
     \qquad - A contextual embedding (e.g., by feeding a template sentence containing `p` into a pre-trained encoder).
   
   \quad - Optionally incorporate web co-occurrence statistics: frequency of `p` instantiated with random targets in W.

2. **Similarity Graph**  
   
   \quad - Construct a graph where patterns are nodes.
   
   \quad - Edge weights combine:
     
     \qquad - Embedding cosine similarity.
     
     \qquad - Jaccard similarity over token sets.
     
     \qquad - Correlation of sentiment distributions (patterns that mostly appear in positive contexts vs negative).

3. **Clustering**  
   
   \quad - Use graph clustering (e.g., spectral clustering or community detection) to form clusters of paraphrastic patterns.
   
   \quad - Each cluster `C\_k` represents an abstract opinion template `T\_k`.

4. **Template Labeling**  
   
   For each cluster:

   \quad - Aggregate sentiment and stance labels of messages that instantiate patterns in the cluster.
   
   \quad - Define:
    
   \quad {\textbackslash}[
     
    \qquad P(y\_s {\textbackslash}mid T\_k), {\textbackslash}quad P(y\_\{st\} {\textbackslash}mid T\_k, {\textbackslash}text\{polarity toward target\})
     
    \quad {\textbackslash}]
   
   \quad - Keep clusters with high label purity as reliable opinion templates (e.g., PRAISE, COMPLAIN, SARCASM, APPROVAL, DISAPPROVAL).

Messages in which explicit sentiment-bearing words are absent but that instantiate a known template can still be assigned sentiment/stance based on the template’s learned distribution.

\textbf{5. Context Expansion for Short Messages}

Given a message `m` with tokens `x` and normalized tokens `z`:

1. **User Context**  
   
   \quad - Collect the user’s recent `K` messages (e.g., last 50) and extract:
     
     \qquad - Personal sentiment baseline (always positive, always negative, sarcastic, etc.).
     
     \qquad - Frequent topics and known stances (via templates and hashtags).

2. **Conversation Context**  
   
   \quad - If `m` is a reply or part of a thread, gather the surrounding messages `m\_\{-L\}, {\textbackslash}cdots, m\_\{+L\}`.
   
   \quad - Encode them using the same encoder as for `m` and obtain a context vector `c\_conv(m)`.

3. **Topical Web Context**  
   
   \quad - Identify entities and hashtags in `m` (using NER and hashtag segmentation).
   
   \quad - Retrieve short descriptions/snippets for those entities from W (pre-crawled offline to avoid latency).
   
   \quad - Encode these descriptions to obtain `c\_web(m)`.

We then define a combined contextual representation:

{\textbackslash}[

\quad c(m) = f\_\{{\textbackslash}text\{agg\}\}{\textbackslash}big( f\_\{{\textbackslash}text\{user\}\}(m), f\_\{{\textbackslash}text\{conv\}\}(m), f\_\{{\textbackslash}text\{web\}\}(m) {\textbackslash}big)

{\textbackslash}]

where each `f\_*` is a small neural network, and `f\_agg` can be a gated or attention-based fusion mechanism.

\textbf{6. Multi-Task Sentiment and Stance Model}

\textbf{6.1 Encoder}

We define a shared encoder `E\_{\textbackslash}theta`:

- Input:
  
  \quad - Original tokens `x`.
 
  \quad - Normalized tokens `z = g\_{\textbackslash}phi(x)`.
  
  \quad - Token-level metadata (elongation, repeated punctuation, capitalization, slang priors `{\textbackslash}sigma(w)`).

- Architecture:
  
  \quad - Embedding layer combining:
    
    \qquad - Subword embeddings for `x` and `z`.
    
    \qquad - Lexicon features (e.g., from sentiment and slang induction).
  
  \quad - A contextual encoder (e.g., Transformer) producing a sequence of hidden states `h\_1,{\textbackslash}cdots,h\_n`.
  
  \quad - A pooling mechanism (e.g., self-attention) to produce a message-level representation `h\_m`.

Context expansion:

- Concatenate `h\_m` with contextual representation `c(m)` from {\textbackslash}S 5 to get final message representation `u\_m`.

\textbf{6.2 Task Heads}

We define several prediction heads:

1. **Sentiment Head**  
   
   \quad - Output distribution over sentiment labels `Y\_s` (e.g., `{very neg, neg, neutral, pos, very pos}`).
   
   \quad - Softmax:
     
     \quad {\textbackslash}[
     
     \qquad {\textbackslash}hat\{y\}\_s = {\textbackslash}text\{softmax\}(W\_s u\_m + b\_s)
     \quad {\textbackslash}]

2. **Stance Head**  
   
   \quad - For messages with identifiable target `T`:
     
     \qquad - Construct a target-specific representation `u\_\{m,T\}` (e.g., by aggregating token representations aligned with `T`, plus `u\_m`).
     
     \qquad - Predict stance `Y\_{st}` (e.g., `{favor, against, neutral}`):
       
       \qquad {\textbackslash}[
       
       \qquad {\textbackslash}hat\{y\}\_\{st\} = {\textbackslash}text\{softmax\}(W\_\{st\} u\_\{m,T\} + b\_\{st\})
       
       \qquad]

3. **Template Head**  
   
   \quad - Predict the most likely opinion template `T\_k`:
     
     \quad {\textbackslash}[
     
     \quad {\textbackslash}hat\{t\} = {\textbackslash}text\{softmax\}(W\_t u\_m + b\_t)
     
     \quad ]
   
   \quad - Training labels for this head come from pattern/template assignments in {\textbackslash}S 4.

4. **Normalization Head (Auxiliary)**  
   
   \quad - We retain the original normalization `g\_{\textbackslash}phi` as a separate module but can also share lower layers with `E\_{\textbackslash}theta` to encourage robustness to noise.

\textbf{6.3 Noise-Aware Loss}

Due to weak/noisy labels, we define a noise-aware training objective.

Let `D\_w` be the weakly labeled dataset and `D\_g` a small gold-standard human-annotated dataset.

For `m {\textbackslash}in D\_w`:

- Treat weak label `{\textbackslash}tilde\{y\}\_s(m)` as a distribution `q\_s(m)` over `Y\_s`, encoding uncertainty via:
  
  \quad - Hashtag/emoji class confusion.
  
  \quad - Slang sentiment prior uncertainty `u(w)`.

- Loss:
  {\textbackslash} [
  
  \quad {\textbackslash}mathcal\{L\}\_\{{\textbackslash}text\{sent-weak\}\} = - {\textbackslash}sum\_\{m {\textbackslash}in D\_w\} {\textbackslash}sum\_\{y {\textbackslash}in Y\_s\} q\_s(m)[y] {\textbackslash}log {\textbackslash}hat\{y\}\_s(m)[y]
  
  {\textbackslash}]

Similarly for stance `q\_\{st\}(m,T)`.

For `m {\textbackslash}in D\_g` with gold labels `y\_s\^*(m)`, `y\_\{st\}\^*(m,T)`:

{\textbackslash}[

\quad {\textbackslash}mathcal\{L\}\_\{{\textbackslash}text\{sent-gold\}\} = - {\textbackslash}sum\_\{m {\textbackslash}in D\_g\} {\textbackslash}log {\textbackslash}hat\{y\}\_s(m)[y\_s\^*(m)]

{\textbackslash}]

{\textbackslash}[

\quad {\textbackslash}mathcal\{L\}\_\{{\textbackslash}text\{stance-gold\}\} = - {\textbackslash}sum\_\{(m,T) {\textbackslash}in D\_g\} {\textbackslash}log {\textbackslash}hat\{y\}\_\{st\}(m,T)[y\_\{st\}\^*(m,T)]

{\textbackslash}]

Template prediction loss:

{\textbackslash}[

\quad {\textbackslash}mathcal\{L\}\_\{{\textbackslash}text\{templ\}\} = - {\textbackslash}sum\_\{m {\textbackslash}in D\_\{{\textbackslash}text\{templ\}\}\} {\textbackslash}log {\textbackslash}hat{t}(m)[t\^*(m)]

{\textbackslash}]

Normalization loss is as in §2.2. Total loss:

{\textbackslash}[

\quad {\textbackslash}mathcal\{L\} = {\textbackslash}lambda\_w ({\textbackslash}mathcal\{L\}\_\{{\textbackslash}text\{sent-weak\}\} + {\textbackslash}mathcal\{L\}\_\{{\textbackslash}text\{stance-weak\}\})

+ {\textbackslash}lambda\_g ({\textbackslash}mathcal\{L\}\_\{{\textbackslash}text\{sent-gold\}\} + {\textbackslash}mathcal\{L\}\_\{{\textbackslash}text\{stance-gold\}\})

+ {\textbackslash}lambda\_t {\textbackslash}mathcal\{L\}\_\{{\textbackslash}text\{templ\}\}

+ {\textbackslash}lambda\_n {\textbackslash}mathcal\{L\}\_\{{\textbackslash}text\{norm\}\}

{\textbackslash}]

Hyperparameters `{\textbackslash}lambda\_*` control the relative influence of each component.

We train `{\textbackslash}theta` and `{\textbackslash}phi` jointly or in alternating phases, starting with heavy emphasis on normalization and template prediction, then gradually increasing the weight of sentiment/stance heads.

\textbf{7. Calibration and Evaluation}

\textbf{7.1 Calibration Against Gold Data}

We reserve a small, high-quality annotated dev set of social messages with:

- Gold sentiment labels.

- Gold stance labels for a subset of targets.

- Annotation of sarcasm and irony where possible.

We:

- Perform temperature scaling or isotonic regression on the outputs of sentiment and stance heads using this dev set, improving probability calibration.

- Evaluate per-class F1, macro-averaged F1, and calibration metrics (e.g., Brier score).

\textbf{7.2 Baseline Comparisons}

We implement several baselines:

- **N-gram + Lexicon Baseline**  
  
  \quad - Traditional sentiment lexicons augmented with slang priors `{\textbackslash}sigma(w)`, ignoring templates and context.

- **Supervised Only Baseline**  
  
  \quad - A classifier trained only on the small gold set, without weak labels or web-based features.

- **Web N-gram Only Baseline**  
 
  \quad - Sentiment classification based purely on web co-occurrence statistics between message terms and sentiment words, similar to web n-gram models for other tasks.

We expect our multi-task, web-augmented model to outperform these baselines, especially on:

- Messages with heavy slang/creative spelling.

- Cases where sentiment is conveyed indirectly via templates rather than explicit sentiment words.

- Short, context-dependent messages where user/conversational context is necessary.

\textbf{7.3 Ablation Studies}

To understand contributions, we perform ablation experiments:

- Remove normalization (`{\textbackslash}lambda\_n = 0`).

- Remove template clustering and template head (`{\textbackslash}lambda\_t = 0`).

- Remove web-based slang priors `{\textbackslash}sigma(w)`.

- Remove user/conversation context `c(m)`.

We expect:

- Normalization to be critical for rare noisy spellings.

- Templates to be crucial for cases with minimal lexical sentiment cues.

- Web-based priors to particularly help for new or rapidly evolving slang.

\textbf{8. Deployment Considerations}

For a practical system:

- Precompute and periodically update:
  
  \quad - Slang sentiment priors `{\textbackslash}sigma(w)`.
  
  \quad - Template clusters and their sentiment distributions.
  
  \quad - Web snippets and entity descriptions for common hashtags and entities.

- Maintain an online component that:
  
  \quad - Incorporates new slang/hashtags dynamically by computing their distributional similarity and web co-occurrences.
  
  \quad - Updates priors without retraining the whole model, by adjusting an auxiliary lookup table.

The end result is a robust, adaptive sentiment and stance analysis system for noisy, short social media texts that leverages the web as an implicit training corpus, combines surface pattern and paraphrase-based relational templates, and explicitly addresses informal language phenomena characteristic of microblogs and SMS.

\eendquote

\end{btemplate1}

\end{document}